\documentclass[twoside]{article}
\usepackage[accepted]{aistats2024}
\usepackage{natbib}
\usepackage[utf8]{inputenc} 
\usepackage[T1]{fontenc}    
\usepackage{url}            
\usepackage{subfigure}
\usepackage{booktabs}       
\usepackage{amsfonts}       
\usepackage{nicefrac}       
\usepackage{microtype}      
\usepackage{xcolor}         
\usepackage{amsmath, amssymb, bbm,xcolor, amsthm, enumitem}
\usepackage{algorithm}
\usepackage{graphicx}
\usepackage{algpseudocode}
\usepackage{hyperref}
\hypersetup{colorlinks,allcolors=blue,linktocpage=true}
\usepackage{thm-restate, thmtools}

\newtheorem{remark}{Remark}

\newtheorem{assumption}{Assumption}

\DeclareMathOperator*{\argmax}{arg\,max}

\newcommand{\esmaeil}[1]{
   {\color{brown} [Esmaeil: {#1}]}
  }

\newcommand\norm[1]{\lVert#1\rVert}

\begin{document}
\twocolumn[
\aistatstitle{Contextual Bandits with Budgeted Information Reveal}
\aistatsauthor{Kyra Gan \And Esmaeil Keyvanshokooh \And  Xueqing Liu \And Susan Murphy }
\aistatsaddress{ Cornell Tech \And  Texas A\&M University \And Duke-NUS Medical School \And Harvard University } ]
\begin{abstract}
Contextual bandit algorithms are commonly used in digital health to recommend personalized treatments. However, to ensure the effectiveness of the treatments, patients are often requested to take actions that have no immediate benefit to them, which
we refer to as  \emph{pro-treatment} actions.
In practice, clinicians have a limited budget to encourage patients to take these actions and collect additional information. 
We introduce a novel optimization and learning algorithm to address this problem.
This algorithm 
effectively combines
the strengths of two algorithmic approaches in a seamless manner, including
1) an online primal-dual algorithm for deciding the optimal timing to reach out to patients, and 2) a contextual bandit learning algorithm to deliver  personalized treatment to the patient.
We prove that this algorithm admits a sub-linear regret bound.
We illustrate the usefulness of this algorithm on both synthetic and real-world data.
\end{abstract}
\section{INTRODUCTION}\label{sec:intro}
In digital health, 
to ensure the effectiveness of treatments, 
patients are often requested to take actions that have no immediate benefit to them. 
We refer to these actions as \emph{pro-treatment actions}.
For instance, 
in personalized addiction treatment,
the effectiveness of the treatment may be compromised if patients fail to complete self-reports
\citep{carpenter2020developments}. 
Another 
scenario arises when utilizing commercial sensors or when there is a need to aggregate data across multiple patients.
In this case, data collected from the sensor
can exclusively be accessed via cloud servers, implying that the data-collecting device, such as wearables and electronic toothbrush, may only be able to communicate with the intervention-delivery device, such as smartphones, through the cloud. 
To ensure the proper delivery of personalized treatments, patients may need to open a dedicated app on their smartphones, enabling the app to retrieve the latest treatment recommendations from the cloud~\citep{trella2022designing}.
When patients neglect pro-treatment actions, clinicians may resort to \emph{limited}, \emph{costly} nudges, such as clinician follow-ups to encourage compliance.

We are interested in answering the following question: \emph{given a limited budget for expensive nudges  for use when patients fail to take pro-treatment actions, when should these nudges be used?}   
To answer this question, we \emph{reformulate} the problem
by introducing \emph{two agents}.
The first agent functions as a \textbf{recommender}, specifically,
a learning agent that
uses 
all the \emph{revealed} patient information 
up to the current time 
to recommend the treatment action for the subsequent time step.
The second agent, a \textbf{revealer}, possesses \emph{current} (or some surrogate of current) and past patient data, often collected by sensors, and determines whether to
reveal this information to the
 recommender, enabling the learning of personalized treatment. 

\begin{figure}[t]
    \centering
\includegraphics[width=\linewidth]{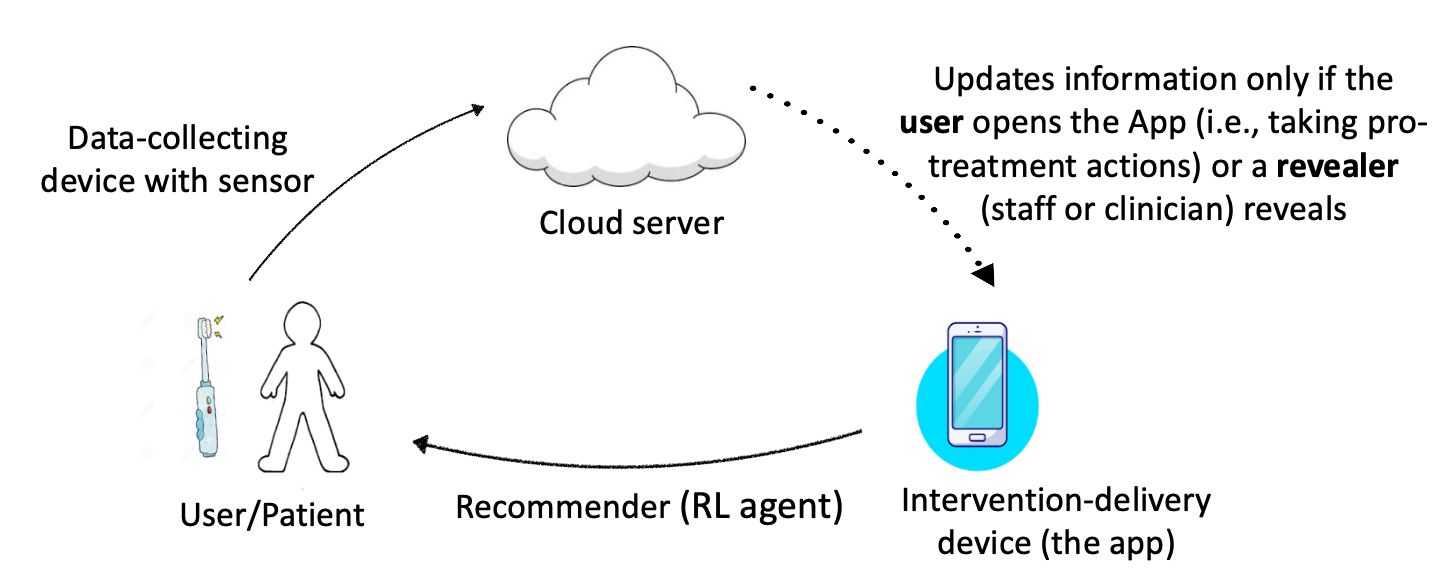}
    \caption{A depiction of the problem breakdown in the electronic toothbrush application, which we extensively investigate in our experiments. In our problem reformulation, it is equally valid to view the recommender as residing on the intervention-delivery device.}
    \label{fig:setup}
\end{figure}

In the commercial sensor example, 
the recommendations (smart phone notifications) are often made by 
a 
cloud-based
\emph{reinforcement learning} (RL) algorithm. 
Although the RL algorithm observes all the information up to the current time, these up-to-date recommendations remain \emph{hidden} to the patient unless they open a dedicated app on their phone (i.e., taking a pro-treatment action).
The clinician serves as the revealer, prompting the patient to open the app as a means of revealing information (as depicted in Figure~\ref{fig:setup}, assuming the patient never opens the app on their own).
At decision points, the clinician assesses all sensor data and decides whether to contact the patient.
If the clinician contacts the patient and the patient opens the mobile application, they receives the most recent recommended notification. Without such actions, the patient receives an ``outdated'' recommendation based on the information up to the \emph{last reveal}.
%
Therefore, \emph{equivalently}, 
in this problem, the recommender
only has access to the history up to the last reveal. However, 
upon opening the app, the recommender gains access to the entire sensor data history.


\paragraph{Our Contributions}
In this work, we provide an algorithm
for deciding the ``optimal'' timing for the  revealer to take action when the number of actions that it can take is limited.
%
We focus on the special case where the recommender acts 1) as a \emph{linear contextual bandit} algorithm when the revealer opts to reveal information and 2) as a \emph{multi-armed bandit} (MAB) algorithm when no additional information is revealed to the  recommender (due to the missing context). 
We show that our problem can be decomposed into two parts: 1) an online primal-dual optimization algorithm addressing the decision of the revealer, and 2) a contextual bandit learning algorithm with delayed feedback modeling the decision of the recommender.
%

In the online primal-dual algorithm, we introduce a novel learning constraint and prove that the value of the objective function of our proposed algorithm, Algorithm~\ref{alg:primal-dual-step-constraint}, is at least $\pi_{\min}(1-1/c)$ times that of an offline clairvoyant benchmark, where $\pi_{\min}$ is a problem dependent constant and $1-1/c$ is budget dependent 
constant that 
approaches $1-1/e$ as the budget grows. 
Furthermore, by introducing the novel learning constraint in the primal-dual algorithm, we are able to separate out the effect of delayed feedback in the bandit learning loss (defined in \S~\ref{sec:setup}), removing the dependency of the delayed feedback effect on the context dimension.
%
%
By combining these two parts,  we provide the first UCB-based algorithm
(Algorithm~\ref{alg:learning-reward}) that 
achieves a sublinear regret under suitable choice of parameters.

\paragraph{Related Work}
Our work is related to three streams of literature: (i) online optimization algorithms, (ii) contextual bandits under resource constraints, and (iii) contextual bandits with delayed feedback.
%
Studies in (i)
typically focus on two arrival settings: stochastic and adversarial. 
In the stochastic setting, online algorithms either rely on 
the forecasted
arrival pattern using historical data or
assume
a stochastic 
arrival pattern 
\citep{goel2008online, karande2011online, mahdian2011online, zhalechian2023data, feldman2009online, jaillet2014online, devanur2019near}, while in the adversarial setting, 
algorithms
are
robust to possible changes in the arrival pattern 
\citep{mehta2007adwords, buchbinder2007online, aggarwal2011online, 
keyvanshokooh2021online, zhalechian2022online,
devanur2012online, liu2024online}. 
The online primal-dual mechanism is one class of algorithms that leverages
the dual program
to guide
the decisions
of
the online algorithm \citep{buchbinder2009design, keyvanshokooh2021online}.
In our work, we introduce a new class of online primal-dual mechanisms  
with a learning component and
incorporate it 
as an online allocation sub-routine in our proposed framework. 
We evaluate its performance
using a
{\em competitive ratio}, 
which compares 
its performance 
to that of
a clairvoyant policy 
on
the worst-case input instance. 

Studies in (ii) usually assume that
each action consumes a certain amount of resources. Under such resource constraints, 
previous works have
proposed online algorithms for standard MAB \citep{agrawal2014bandits,badanidiyuru2018bandits,ferreira2018online}, contextual bandit, and  other RL methods~\citep{badanidiyuru2014resourceful,agrawal2016efficient,agrawal2016linear, wu2015algorithms,pacchiano2021stochastic,cao2023safe}.
In contrast, in our algorithm, the recommender is required to take an action at each time step regardless of whether it observes the current context.
A few works formulate contextual bandit with resource constraints by integrating Thompson sampling (bandit) algorithms with online optimization algorithms \citep{cheung2022inventory,zhalechian2022online}. 
In comparison, our work provide the first analysis for UCB-based bandit algorithms, and we
incorporate a learning component into the online allocation mechanisms while learning
only
happens in the bandit part of their algorithms. This additional level of learning
helps our algorithms 
achieve
a better performance. 

Studies in (iii) 
 include both
 delayed feedback in MAB 
\citep{joulani2013online, bistritz2019online, pike2018bandits} and contextual bandit
\citep{zhou2019learning,vernade2020linear, keyvanshokooh2023contextual}, where the delay patterns mostly fall into \emph{bounded} delays or \emph{stochastic} delays. However, in our problem, the delayed patterns are structured: unless the revealer takes an action, the recommender received no additional information. 
While applying algorithms from (iii) 
in our problem setting is feasible, our algorithm offers a significantly stronger theoretical guarantee.
The order of our regret bound is tight (optimal) up to a logarithmic factor
\citep{chu2011contextual}. Also, the delayed feedback only impacts our regret bound by an \emph{additive} factor of $\sum_{t=1}^T \beta_t$, which is of order $\mathcal{O}(\sqrt{T})$. Therefore, unlike \cite{zhou2019learning} and \cite{vernade2020linear}, our theoretical result removes the dependency of the 
delayed feedback effect on the context dimension $d$.

\section{PROBLEM FORMULATION}\label{sec:setup}
We start with the worst-case 
setting where  the  recommender \emph{never} observes  any additional information \emph{unless} the  revealer takes an action at each time step. 
When patients
occasionally take
pro-treatment actions on their own, 
we expect the relative performance of our algorithm (with respect to the benchmark algorithms in Section \ref{sec:experiments}) to stay the same.
In this section, we first introduce the contextual bandit problem and then discuss the setup of each agent. Lastly, we provide an overview of our proposed framework.

\paragraph{Linear Contextual Bandit} 
Let $\mathcal{S}=\{1, ..., K\}$ denote the set of contexts.
Given time horizon $T$,
at each time $t \in [T]$, a context $S_t$ arrives.
We assume that the contexts are drawn i.i.d. from a known distribution 
$\mathbf{p}^*$, where $\mathbf{p}^*_k := \mathbb{P}(S_t = k)$. (See Section \ref{sec:experiments}, and Appendix~\ref{subsec:learn_context}
for the situation where $\mathbf{p}^*$ is unknown.)  
{However, 
the ordering of the realized contexts can be \emph{adversarially} chosen, that is,
the adversary can choose the ordering in which the contexts appear.}\footnote{As seen in Section \ref{sec:auxiliary}, the performance of our proposed online algorithm depends on the context arrival \emph{sequence}.  
}

Let $\mathcal{A}$ denote the set of discrete actions that can be taken by the recommender. The reward $X_t$
under context $S_t$ and action $A_t\in \cal{A}$ is generated according to
$X_t = \langle \theta_*, \phi(S_t, A_t)\rangle + \eta_t$, where $\theta_*\in \mathbb{R}^d$ is an \emph{unknown} true reward parameter,  $\phi: \mathcal{S}\times \mathcal{A}\mapsto \mathbb{R}^d$ is a \emph{known} feature mapping, and the noise $\eta_t$ is conditional mean-zero $1$-sub-Gaussian. 

\paragraph{Recommender} 
For a given patient,
when the recommender has access to the  context $S_t$, it takes action according to a contextual bandit algorithm.
When the recommender does \emph{not} observe the current context $S_t$, it takes actions by treating the bandit problem as a 
MAB,
where the expected reward of each action is now weighted by the context distribution. 
We elaborate on this  structure in Section \ref{sec:auxiliary}, 
and describe our UCB-based bandit algorithms in Section \ref{sec:bandit} (Algorithm~\ref{alg:learning-reward}).
We note that this problem structure does \emph{not} affect the reward generating process, but rather it affects whether the expected reward averages out over context or not.

\paragraph{Revealer}
The revealer is given an expected budget of $B$ information reveals to the recommender throughout the horizon $T$. 
We assume that $B>2|\mathcal{A}|$ for technical ease.
At each time, 
the  binary decision variable for the revealer is
$O_t \in\{0,1\}$. 
Consider the general case where the revealer only observes part of the context. Namely, we can partition each state into two components: $S_t = [S_t^1, S_t^2]$. Let  $S_t^1$ be the part of the state that is \emph{always} observed by the revealer at each time $t$, and let $S_t^2$ be the part of the state that can \emph{only} be observed when the revealer takes the action $O_t=1$. 
Let $\ell(t)$ be the time of the last reveal. At each decision time $t$, the revealer observes the history $\mathcal{H}_t^{\mathrm{rev}} = \{ A_1,...,A_{t-1}, X_1, ...X_{t-1}, O_1, ...,O_{t-1},$ $S_1, ...,  S_{\ell(t)}, S_{\ell(t)+1}^1,  ..., S_t^1\}$,
and \emph{decides} the revealing  probability $o_t$; then $O_t \sim \mathrm{Bernoulli}(o_t)$.
If $O_t=1$, 
the revealer additionally observes $\{S_{\ell(t)+1}^2,  ..., S_t^2\}$ and the
recommender observes
$\mathcal{H}_t^{\mathrm{rec}} 
= \{ A_1, ..., A_{t-1},X_1, ..., $ $X_{t-1},  O_1, ..., O_{t-1}, S_1, ...,  S_t\}$.  Otherwise, the recommender observes the history up to time $\ell(t)$, $\mathcal{H}_{\ell(t)}^{\mathrm{rec}}$. 
We highlight the key information asymmetry here lies in the fact that the revealer can always observe $S_t^1$, but the recommender cannot.
The budget constraint requires $\sum_{t=1}^T o_t \leq B$.
%
%
%
For ease of exposition, we will focus on the special case where the revealer observes the entire context at time $t$, i.e., $S_t^1 = S_t$, from now on. 
Further discussion on how our theoretical guarantees apply to the above mentioned general setting is included in Appendix~\ref{append:partial_context}.


\paragraph{Framework Overview and Regret Decomposition} Given that the number of actions that the revealer can take is limited,
our objective is to
develop a data-driven optimization and learning framework that can 
1) decide the
optimal timing for the revealer to
reveal, 
and 2)
learn the optimal
treatment for the  recommender. We achieve the former by designing an  online primal-dual algorithm with a novel learning constraint (Algorithm~\ref{alg:primal-dual-step-constraint}) and achieve the latter by applying the UCB algorithm (Algorithm~\ref{alg:learning-reward}) 
which uses the online primal-dual algorithm (Algorithm~\ref{alg:primal-dual-step-constraint}) as a subroutine.

There are two main sources of uncertainty
in this problem:
(1) the unknown reward
parameter,
$\theta_*$, and (2) the  \emph{sequence of future context arrivals}, $\{s_t, ..., s_T\}$. 
We discuss 
unknown
context distribution $\mathbf{p}^*$ 
in Appendix \ref{subsec:learn_context}.
%
%
%
%
We evaluate the performance of our algorithm 
using an \emph{offline clairvoyant} benchmark, where 
both the revealer and the recommender know the reward distribution,  $\theta_*$ 
and additionally, the revealer knows the entire  context arrival sequence $\{s_1, ..., s_T\}$.
Note that \emph{no} algorithm can ever achieve this performance in practice since
future contexts are inherently unknown.

We introduce a novel 
regret analysis to assess the theoretical performance of our algorithm in relation to the clairvoyant problem. Our analysis 
 seamlessly combines
 a \emph{competitive ratio} bound for bounding the sub-optimality gap of the revealer with  a \emph{regret} bound of the recommender.
 This integration necessitates 1) defining an \emph{auxiliary problem} and 2) using a \emph{bridging argument}. In the auxiliary problem, we assume knowledge of
 the unknown 
 model parameter ($\theta_*$), 
but the \emph{future} context arrival sequence is \emph{unknown}. Specifically, the auxiliary problem simulates the \emph{online} version of the clairvoyant problem where the contexts
arrive sequentially.
We note that this terminology also has been used in the existing literature (see 
\cite{cheung2022inventory}).

Let $V^{\text{AUX}} \text{ and } V^{\text{ALG}}$ be the respective value functions of Algorithm~\ref{alg:learning-reward} when $\theta_*$ is known and unknown,
and let $V^{\text{CLV}}$ be the value function of the clairvoyant problem; see  Section \ref{sec:auxiliary} and
Section \ref{sec:bandit} for formal definitions.
%
We decompose the regret 
by establishing the following bridging 
argument:
\begin{align*}
\mathrm{Regret}_T   
&\leq
\mathbb{E}\left[ V^{\text{CLV}} \right] - \mathbb{E}\left[ V^{\text{ALG}}
\right]
\\& 
=  \underbrace{ \mathbb{E}\left[ V^{\text{CLV}} - V^{\text{AUX}} \right] }_\text{Information Reveal Loss} +
\underbrace{ \mathbb{E}\left[ V^{\text{AUX}} - V^{\text{ALG}} \right] }_\text{Bandit Learning Loss},
\end{align*}
In the above decomposition, the first expression represents the loss due to the optimality gap of the information revealing mechanism for solving the auxiliary problem. The expectation is taken over the stochasticity of the algorithm, as  the reward parameter is known in both problems.
The second represents 
the loss due to contextual bandit learning, i.e., learning the unknown reward. The expectation is taken over the stochasticity in both the algorithm and the environment.

\section{BOUNDING INFORMATION REVEAL LOSS}
\label{sec:auxiliary}
In this section, we first formally introduce the clairvoyant
problem. 
Then, by
developing an online primal-dual approach
for solving the clairvoyant problem in an online fashion, we
provide a feasible solution to the auxiliary problem.
Finally, we provide an upper bound on the information reveal loss. 

\textbf{Clairvoyant Problem}$\;\;$
Recall that in the clairvoyant problem, both the revealer and recommender know $\theta_*$, 
and the revealer additionally knows
the entire context \emph{realized} sequence $\{s_1, ..., s_T\}$.
The optimal strategy for the recommender is to take the optimal action corresponding to context $s_t$ when $s_t$ is observed, and to take the action with the highest expected (where the expectation is taken over the context distribution) reward when $s_t$ is \emph{not} observed. 
The former happens when the revealer takes action
$O_t=1$ at time $t$, corresponding to
\emph{revealing} the history $\mathcal{H}_t^{\mathrm{rev}}$ to the recommender, and 
the latter happens when the revealer decides \emph{not to} take action
at time $t$. 
Let 
$u_{s_t}^* =\max_{a\in\mathcal{A}} \left\langle\theta_*, \phi(s_t, a)\right\rangle$, and
$v^* = \max_{a\in\mathcal{A}} \langle\theta_*, \bar\phi(a) \rangle$,
where $\bar\phi(a)$ is the weighted feature mapping, i.e., $\bar\phi(a) = \sum_{k=1}^K \phi(k,a)\mathbf{p}^*_k$.
%

Using this optimal strategy for the recommender,
a natural objective of the revealer is to maximize the expected reward collected throughout the horizon: $\max_{o_t}\sum_{t=1}^T o_t\cdot u_{s_t}^* + (1-o_t)\cdot v^* $.
By removing the constant $v^*$, we obtain the following formulation for the clairvoyant problem (CLV):
\begin{align}
\label{pblm: clairvoyant}
\Bigg\{ \max_{o_t}  \sum_{t=1}^T o_t\cdot (u_{s_t}^* - v^*) :
&\sum_{t=1}^T o_t \leq B,o_t \in [0,1],\notag\\  
 &  \forall t \in [T]. \Big\}\tag{\textrm{\emph{CLV}}}
\end{align}
%
%
%
The optimal policy of the revealer in \eqref{pblm: clairvoyant} is first to select the contexts that yield more reward than $v^*$, i.e., with positive $u_{s_t}^* - v^*$, and second set $o_t = 1$ for the top $B$ contexts that have the highest expected reward, $u_{s_t}^*$'s.
Thus, $V^{\text{CLV}}$ is the sum of the expected optimal rewards where the expectation is taken over the decision variable $O_t$, i.e., 
$\mathbb{E}[V^{\text{CLV}}] = \max_{a\in\mathcal{A}}\sum_{t=1}^T (o_t^\text{CLV} \left \langle 
 \theta_*,
    \phi(s_t, a) \right\rangle + (1-o_t^\text{CLV})\left\langle \theta_*,
\bar\phi(a) \right\rangle ), 
$
where the sequence of $\{o_t^\text{CLV}\}_{t\in[T]}$ is the solution to \eqref{pblm: clairvoyant}. 
Note that
\emph{without} knowledge of 
current context $S_t$, each decision point becomes identical to the revealer, resulting in a
trivial optimal solution for the revealer 
-- randomly selecting $B$ times to reveal $\mathcal{H}_t^{\mathrm{rev}}$, highlighting the importance of information asymmetry between revealer and recommender.

\begin{remark}[Objective function of \ref{pblm: clairvoyant}]\label{remark:objective}
We note that if there are not enough contexts $s_t$'s with a positive $u_{s_t}^*-v^*$, then we do not use the entire budget $B$. We note that our objective function is suitable for the low-budget regime, i.e., when $B$ is less than or equal to the number of contexts with positive $u_{s_t}^*-v^*$. For larger $B$'s, one could remove $-o_t v^*$ from the objective function, and the rest of the result still holds.

An alternative objective function in this problem is $\max_{o_t} \sum_{t=1}^T o_t u_{s_t}^*$. Indeed, when the budget is low, these two objectives yield the same optimal solution to the clairvoyant problem.
However, as we will see in our online primal-dual algorithm (Algorithm~\ref{alg:primal-dual-step-constraint}), when we do not have access to the \emph{future} context arrival sequence, $v^*$ serves as a regularization term for spending the budget $B$ (by ignoring the contexts that yield negative $u_{s_t}^*-v^*$), yielding better algorithmic performance.
\end{remark}

As a direct consequence of Remark~\ref{remark:objective}, we assume that the budget is not too large when compared with the horizon length $T$:
\begin{assumption}\label{assum:budget}
We assume that $B=\mathcal{O}(\sqrt{T})$.
\end{assumption}
%

%
In \eqref{pblm: clairvoyant}, the optimal strategy of the revealer 
depends on the entire \emph{realized} context arrival sequence, including the future arrivals $\{s_{t+1}, ..., s_T\}$. While no algorithm in practice can achieve this performance, 
\eqref{pblm: clairvoyant}
has two critical advantages:
1) it provides an \emph{upper bound} for the optimal solution to the oracle problem: the oracle problem  can be viewed as  \eqref{pblm: clairvoyant} with the additional constraint that the context sequence is observed up to time $t$, $\{s_1, ..., s_t\}$; 2) it naturally provides insight into how to incorporate online primal-dual mechanisms. 

\textbf{Auxiliary Problem}$\;\;$ The \emph{auxiliary problem} is an \emph{online} version of \eqref{pblm: clairvoyant}, where the contexts arrive sequentially, and the decision of the revealer should be made to hedge against the adversarial context arrival sequence in the future. In the auxiliary problem, both agents know 
$\theta_*$, 
and neither has access to the future context arrival sequence 
(which might as well be \emph{adversarial}).
We develop an \emph{online primal-dual}  algorithm (Algorithm~\ref{alg:primal-dual-step-constraint}) 
to provide a feasible solution for the revealer in
the auxiliary problem. We
rigorously analyze it
using the \emph{competitive ratio analysis}.

\paragraph{Modified Clairvoyant Problem} Intuitively, the revealing probability $o_t$ at each time step should depend on both 1) the budget that we have spent so far and 2) the rate at which we learn the reward and context distributions.
However, as it currently stands, the dual of \eqref{pblm: clairvoyant} lacks a mechanism to connect
the quality of the estimates that the \emph{recommender} has at time $t$ to
the revealing decision $o_t$. Ideally, we would like $o_t$ to increase as the time since the last reveal increases.

To solve this technical challenge,
we next incorporate a novel \textbf{learning constraint}, Constraint~\eqref{eq:learning-constraint}.
In Appendix~\ref{append:subsec_without_learning}, we provide a road map for deriving this 
constraint.
We first
provide 
an algorithm that only takes the budget into account (Algorithm~\ref{alg:primal-dual}),
 %
 and then provide its theoretical guarantee  (Proposition~\ref{prop:primal-dual}).

The online primal-dual algorithm that we will develop in this section serves
as a \emph{subroutine} in our bandit learning algorithm.  The bandit algorithm provides  estimates of $u^*_{s_t}$ and $v^*$ to the online primal-dual algorithm. 
%
We make the following critical observation: 
at each time $t$,
the revealer 
has access to both $\mathcal{H}_t^{\mathrm{rev}}$ and $\mathcal{H}_t^{\mathrm{rec}}$,
the revealer can calculate both 
the recommender's optimal action, $\hat a_t$,  if the revealer were \emph{not} to reveal $\mathcal{H}_t^{\mathrm{rev}}$ ($O_t=0$), 
and  the optimal action,  $\tilde a_t$,  when   $\mathcal{H}_t^{\mathrm{rev}}$   is revealed ($O_t=1$). We describe the calculation of the above  in detail in Section~\ref{sec:bandit}.
Next, we introduce a constraint to force
the revealing probability to increase when the
estimated optimal treatment 
differs 
between the two agents, i.e., $\hat a_t\neq \tilde a_t$, \emph{and} the distance between the weighted feature mappings, $\|\bar\phi(\tilde a_t) -\bar\phi(\hat a_t)\|_2$, (recall $\bar\phi(a) = \sum_{k=1}^K \phi(k,a)\mathbf{p}^*_k$)
is large:
%
%
\begin{align}
\label{eq:learning-constraint}
    \left\|\bar\phi(\tilde a_t) -\bar\phi(\hat a_t)\right\|_2\mathbbm{1}\left(\hat a_t\neq \tilde a_t\right)(1-o_t) \nonumber \\
    \leq \beta_t(\mathcal{H}_t^{\mathrm{rev}}, \mathcal{H}_{t}^\mathrm{rec}), \;\forall t\in [T],
\end{align}
where $\{\beta_t(\mathcal{H}_t^\mathrm{rev}, \mathcal{H}_{t}^\mathrm{rec})\}_{t=1}^T$
is a sequence of positive constants that can be  \emph{initialized adaptively} by the expert and \emph{auto-adjusted} by our algorithm using the histories, to guarantee the feasibility of
\eqref{pblm: clairvoyant}
with the above constraint.
{To ease notation, we abbreviate the $\beta$'s using $\beta_1, ..., \beta_T$ from now on.}

Appendix~\ref{app:learning_primal} includes the
updated primal problem.
Let $y$, $z_t$'s, and $e_t$'s be the dual variables. We have the resulting
dual of the updated primal problem: 
\begin{align}
\label{Eq: dual-clairvoyant-Learning}
\min_{y, x_t,z_t} \quad & By+ \sum_{t=1}^T z_t  \tag{\textrm{\emph{Modified CLV Dual}}}  \\
& + \sum_{t=1}^T \left(\beta_t - \left\|\bar\phi(\tilde a_t) -\bar\phi(\hat a_t)\right\|_2\mathbbm{1}(\hat a_t\neq \tilde a_t)\right) e_t \notag\\
s.t. \quad &  y + z_t - \left\|\bar\phi(\tilde a_t) -\bar\phi(\hat a_t)\right\|_2\mathbbm{1}(\hat a_t\neq \tilde a_t)e_t \geq \notag \\
& u_{s_t}^* - v^*,  \forall t \in [T] \notag \\ 
\quad & y, z_t, e_t \geq 0, \forall t \in [T].  \notag
\end{align}

%
%
At the margin, 
$y\delta$ corresponds to how the value of the optimal solution to the primal changes if we were to change the budget $B$ by $\delta$,
$z_t$ is the marginal value for revealing information at time step $t$, and $e_t$ is the minimum value that we need to increase $o_t$ to satisfy Constraint~\eqref{eq:learning-constraint}. 
Note that in the above dual problem,
we have a separate constraint for each $z_t$ and $e_t$.
Within this dual structure, the online arrival of constraints corresponds to the sequential arrival of decision variables, enabling the design of online approximation algorithms.
%

Let $u_{\max} = \max_{s\in S} u_{s}^* $ and 
$u_{\min} = \min_{s\in S} u_{s}^* $. Let $\pi_{\min}$ be the smallest positive difference between $u_{s_t}^* $ and $v^*$, i.e., $\pi_{\min} = \min_{s\in S} \max(u_{s}^* - v^*, 0)$.
Let $\pi_{\max}:= \max_{s\in S} |u_{s}^* - v^*|$.
We assume that an \emph{upper bound} on $u_{\max}$  and a \emph{lower bound} on $u_{\min}$ are known to the algorithm by domain knowledge.    
{Without loss of generality}, we assume that $0 \leq u_{s_t}^* - v^* \leq 1 $ for all $s_t\in \mathcal{S}$. Otherwise, we could rescale $u_{s_t}^* - v^*$ by $u_{\max}$ and $u_{\min}$ for all $s_t\in \mathcal{S}$. 
We outline the online primal-dual algorithm in Algorithm \ref{alg:primal-dual-step-constraint}. 

\begin{algorithm*}[!ht]
\caption{Online  Primal-Dual Algorithm Revealer with Learning Component}
\label{alg:primal-dual-step-constraint}
\begin{algorithmic}[1]
\State \textbf{Input:} $B, \{u^*_{s}\}_{s\in S},  
 v^*,
 c= (1+1/B)^B,
 \{\bar\phi(a)\}_{a\in\cal A}$. 

\State \textbf{Initialize:} $y \leftarrow 0, \{\beta_t\}_{t\in T}, e_j \leftarrow 0, \;\forall j\in J$.
\For {t = 1, ..., T}
\State The new context $s_t$ arrives, and we observe $\tilde a_t, \hat a_t.$
\If{$\tilde{a}_t \neq \hat{a}_t$ \textbf{and }{$\beta_t < \left\|\bar\phi(\tilde a_t) -\bar\phi(\hat a_t)\right\|_2$} \textbf{and } $u_{s_t}^* > {v}^*$}
\State $e_t = \frac{1}{\left\|\bar\phi(\tilde a_t) -\bar\phi(\hat a_t)\right\|_2}-\frac{\beta_t}{\left\|\bar\phi(\tilde a_t) -\bar\phi(\hat a_t)\right\|_2^2}$
\If{$u_{s_t}^* - v^* - y 
\leq 0$}

\State{$\beta_t=\left\|\bar\phi(\tilde a_t) -\bar\phi(\hat a_t)\right\|_2$},
{$e_t=0$}.
\EndIf
\Else 
\State{$\beta_t=\left\|\bar\phi(\tilde a_t) -\bar\phi(\hat a_t)\right\|_2$}, {$e_t = 0$}.
\EndIf
\If{$y < 1$ and $u_{s_t}^* - v^* - y + \left\|\bar\phi(\tilde a_t) -\bar\phi(\hat a_t)\right\|_2\mathbbm{1}(\hat a_t\neq \tilde a_t)e_t>0$}
\State ${\beta_t} = \max\left( \beta_t, {\left\|\bar\phi(\tilde a_t) -\bar\phi(\hat a_t)\right\|_2} \left(1-B+\sum_{i=1}^{t-1}o_i\right)\right)$
\State $z_t = u_{s_t}^* - v^* - y + \left\|\bar\phi(\tilde a_t) -\bar\phi(\hat a_t)\right\|_2\mathbbm{1}(\hat a_t\neq \tilde a_t)e_t$
\State $o_t = \min(u_{s_t}^* - v^* +\left\|\bar\phi(\tilde a_t) -\bar\phi(\hat a_t)\right\|_2\mathbbm{1}(\hat a_t\neq \tilde a_t)e_t, B-\sum_{i=1}^{t-1}o_i, 1)$
\State $y \leftarrow y(1+{o_t}/{B}) + {o_t}/{\left((c-1)\cdot B\right)}$.
\Else
\State $\beta_t = \left\|\bar\phi(\tilde a_t) -\bar\phi(\hat a_t)\right\|_2$, $z_t = 0$, $o_t = 0$.
\EndIf
\EndFor
\end{algorithmic}
\end{algorithm*}

At each iteration, this algorithm takes the budget ($B$), the optimal solution value of the recommender ($\{u_{s_t}^*\}$ when $O_t=1$) and $v^*$ when $O_t=0$), and the feature mapping when $s_t$ is not observed ($\{\bar\phi(a)\}_{a\in \mathcal{A}}$) as \emph{input}. It then ``simulates'' the optimal solution of the recommender when $O_t=0$
if the recommender were and were not to have access to $\mathcal{H}^{rev}_t$ ($\tilde a_t, \hat a_t$) and calculates $\|\bar\phi(\tilde a_t) -\bar\phi(\hat a_t)\|_2$. Based on this value, the algorithm decides whether to increase  $o_t$ (when compared with $o_t$ in Algorithm~\ref{alg:primal-dual}).
Algorithm~\ref{alg:primal-dual-step-constraint}
provides a feasible solution to the auxiliary problem, and 
only depends on the history it has observed so far. With $\{o_t^\text{AUX}\}_{t\in[T]}$  chosen according to Algorithm~\ref{alg:primal-dual-step-constraint},
define $\mathbb{E}[V^{\text{AUX}}]$ to be
$$\max_{a\in\mathcal{A}}\sum_{t=1}^T (o_t^\text{AUX}  \langle 
 \theta_*,
    \phi(s_t, a) \rangle + (1-o_t^\text{AUX})\langle \theta_*,
\bar\phi(a) \rangle ). 
$$
Similarly, let  $\{o_t^\text{MCLV}\}_{t\in[T]}$ be the solution to \eqref{Eq: clairvoyant-Learning} in   Appendix~\ref{app:learning_primal}, then $\mathbb{E}[V^{\text{MCLV}}]$ is $$
\max_{a\in\mathcal{A}}\sum_{t=1}^T (o_t^\text{MCLV}  \langle 
 \theta_*,
    \phi(s_t, a) \rangle + (1-o_t^\text{MCLV})\langle \theta_*,
\bar\phi(a)\rangle ). 
$$
We first show the following result:


\begin{restatable}{prop}{primaldualconstraint}   
\label{prop:primal-dual-learning-constraint}
For any $u_{s_t}^*$, $v^*$, and context arrival sequence,
$\mathbb{E}[V^{\text{AUX}}] \geq \pi_{\min}(1 - 1/c)\mathbb{E}[V^{\text{MCLV}}]$.
\end{restatable}

The proof of Proposition~\ref{prop:primal-dual-learning-constraint} 
(Appendix~\ref{proof:prop-primal-dual-learning}) consists of proving
that  Algorithm~\ref{alg:primal-dual-step-constraint} is (1) both primal and dual feasible, and (2) in each iteration (day), the ratio between the change in the primal and dual objective functions is bounded by $\pi_{\min}(1 - 1/c)$. By weak duality,  this implies that Algorithm~\ref{alg:primal-dual-step-constraint} is $\pi_{\min}(1 - 1/c)$-competitive. 
Moreover, when the ratio $\tau = 1/B$ tends to zero, the competitive ratio of the algorithm tends to the best-possible competitive ratio of $(1-1/e)$ \citep{buchbinder2007online}.
In addition,  as $t$ increases, Algorithm~\ref{alg:primal-dual-step-constraint} increasingly favors revealing information when the expected difference in rewards, $u^*_{s_t} - v^*$ is large. Thus, even though $\pi_{\min}$ appears in the competitive ratio, the performance of our algorithm is much better in practice. We illustrate this in Section~\ref{sec:experiments}.

Algorithm~\ref{alg:primal-dual-step-constraint} contains two competing constraints for deciding 
$o_t$, and the key technical challenge
is to ensure primal feasibility.
Some key observations 
include: 1) to avoid negative competitive ratios, we increase $o_t$ only if $u_{s_t}^* - v^*$ is positive;
2) when running out of budget, i.e., $u_{s_t}^* - v^* - y \leq 0$, we increase $\beta_t$ such that the learning constraint is always satisfied;
3) when $u_{s_t}^*-v^*<y$ and $e_t$ is not high enough to make $o_t$ positive, we increase $\beta_t$ such that the second constraint is satisfied. 
%

To complete the regret decomposition, we show
the following corollary (proof in Appendix~\ref{proof:cor}) 
holds:
\begin{restatable}{cor}{corlearning}\label{cor:competitive_with_learning}
For any $u_{s_t}^*$, $v^*$, and any context arrival sequence, 
$\mathbb{E}[V^{\text{AUX}}]\geq $
$\pi_{\min}(1 - 1/c) \mathbb{E}[V^{\text{CLV}}]$.
\end{restatable}

\section{BOUNDING BANDIT LEARNING LOSS}\label{sec:bandit}
Recall from Section~\ref{sec:setup} that 
the reward
under action $A_t$ 
and
context $S_t$, 
$X_{t} (S_t, A_t) = \langle \theta_*, \phi(S_t, A_t)\rangle O_t +
\eta_t,$ 
where
$\theta_*$ is the unknown reward parameter, and the noise $\eta_t$ is conditional mean-zero $1$-sub-Gaussian. For the purpose of proofs, we assume there exists a finite $W$ and finite $L$ for which $\|\theta_*\|_2 \leq W$ with Q-probability one and $\max_{a\in \mathcal{A}, s\in S}\|\phi(s,a)\|_2\leq L$ and $\max_{a\in \mathcal{A}, s\in S}\langle\phi(s,a),\theta_*\rangle \leq 1$ with Q-probability one.  
Let $x_{t}$ be the observed reward at time $t$.

In this section, we 
learn the unknown parameters
$\theta_*$
while 
making \emph{a limited number of} information-revealing decisions. 
%
We propose Algorithm~\ref{alg:learning-reward},
an
online learning and optimization algorithm that
strikes a two-way balance between (i) the exploration-exploitation dilemma for learning the unknown reward, and (ii) hedging against adversarially chosen context arrival sequence.  
%
%
%

It
consists of (i) a contextual UCB mechanism for learning
the unknown
$\theta_*$ 
for making the 
treatment decisions, and (ii) the online primal-dual subroutine (Algorithm~\ref{alg:primal-dual-step-constraint}) for making contextual-revealing decisions. 
At each iteration $t$, 
we maintain two uncertainty sets 
$\tilde C_t$ and $\hat C_t$ using histories $\mathcal{H}_t^{\mathrm{rev}}$ and $\mathcal{H}_t^{\mathrm{rec}}$, respectively, for the unknown $\theta_*$, using the high-probability confidence bound that we derived in Proposition~\ref{prop:concentration}.
Upon observing 
a new context 
$s_t$ by the revealer, the algorithm finds \emph{optimistic} treatments and parameters $(\tilde A_t, \tilde\theta_t)$ and $(\hat A_t, \hat\theta_t)$ given
$\tilde C_t$ and $\hat C_t$, 
respectively,
and derives optimistic reward estimates $\Tilde{u}_{s_t}^t$ and $\Tilde{v}^t$. Given these values,
the revealer deploys the online primal-dual subroutine (Algorithm~\ref{alg:primal-dual-step-constraint}) to decide
the  probability $o_t$
of revealing 
$\mathcal{H}_{t}^{\mathrm{rev}}$ to the recommender. 
If $O_t = 1$,
the recommender updates its uncertainty set, 
i.e., $\hat C_t = \tilde C_t$,
and take the corresponding optimistic action $\hat A_t^*$. Otherwise, the recommender exploits its latest uncertainty set 
and chooses the latest treatment. 
At each iteration, we update
$\hat C_{t+1}$ after observing the new reward feedback $X_{t}$ and context $S_t$. Let $\phi(s_0, a_0^*)=0$. 
See detailed steps in Algorithm~\ref{alg:learning-reward}.

\begin{algorithm}[t]
\caption{Online Learning and Optimization Algorithm}
\label{alg:learning-reward}
\begin{algorithmic}[1]
\State \textbf{Input:} $B, \mathbf{p}^* , \{\bar\phi(a) = \sum_{k=1}^K \phi(k,a)\mathbf{p}^*_k\}_{a\in\cal A}$
\State \textbf{Initialize:}
$\hat C_1$ and $\tilde C_1$ be the confidence interval for $\theta_*$ for recommender and revealer, respectively.
\For{$t=1,...,T$}
\State On each iteration $t$,  \emph{revealer} observes the new context $s_t$ and calculates:
\State $(\tilde A_t, \tilde\theta_t) = \argmax_{(a, \theta)\in\mathcal{A}\times \tilde C_t} \langle \bar\phi(a), \theta\rangle$,
\State $(\hat A_t, \hat\theta_t) = \argmax_{(a, \theta)\in\mathcal{A}\times \hat C_t} \langle \bar\phi(a), \theta\rangle$,
\State $\Tilde{u}_{s_t}^t  = \max_{(a, \theta)\in\mathcal{A}\times \tilde C_t} \langle \phi(s_t, a), \theta\rangle$,
\State   
$\Tilde{v}^t = \left\langle
\bar\phi(\tilde A_t),\Tilde{\theta}_t
\right\rangle$.
\State Given $\Tilde{u}_{s_t}^t, \tilde v_t,$ $\tilde A_t$, $\hat A_t$, and $B$, revealer uses Algorithm~\ref{alg:primal-dual-step-constraint} to reveal the history $\mathcal{H}_{t}^{\mathrm{rev}}$ to recommender with probability $o_t$. Let $O_t \sim \mathrm{Bernoulli}(o_t)$.
\If{$O_t=1$}
\State  \emph{Recommender} set $\hat C_t  = \tilde C_t$ and calculates:
\State $(\hat A^*_t, \theta_t)  = \argmax_{(a, \theta)\in\mathcal{A}\times \tilde C_t} \langle \phi(s_t, a), \theta\rangle$.
\State  \emph{Recommender} takes action $\hat A_t^*$.
\Else 
\State \emph{Recommender} set $\hat C_{t+1} = \hat C_t$ and takes action $\hat A_t^*=\hat A_t$.
\EndIf
\State \emph{Revealer} observes reward $X_t$ and $S_t$, and update $\tilde C_{t+1}$ according to Proposition~\ref{prop:concentration}.
\EndFor
\end{algorithmic}
\end{algorithm}



 \begin{figure*}[!ht]
	\centering
\includegraphics[width=0.46\linewidth,keepaspectratio]{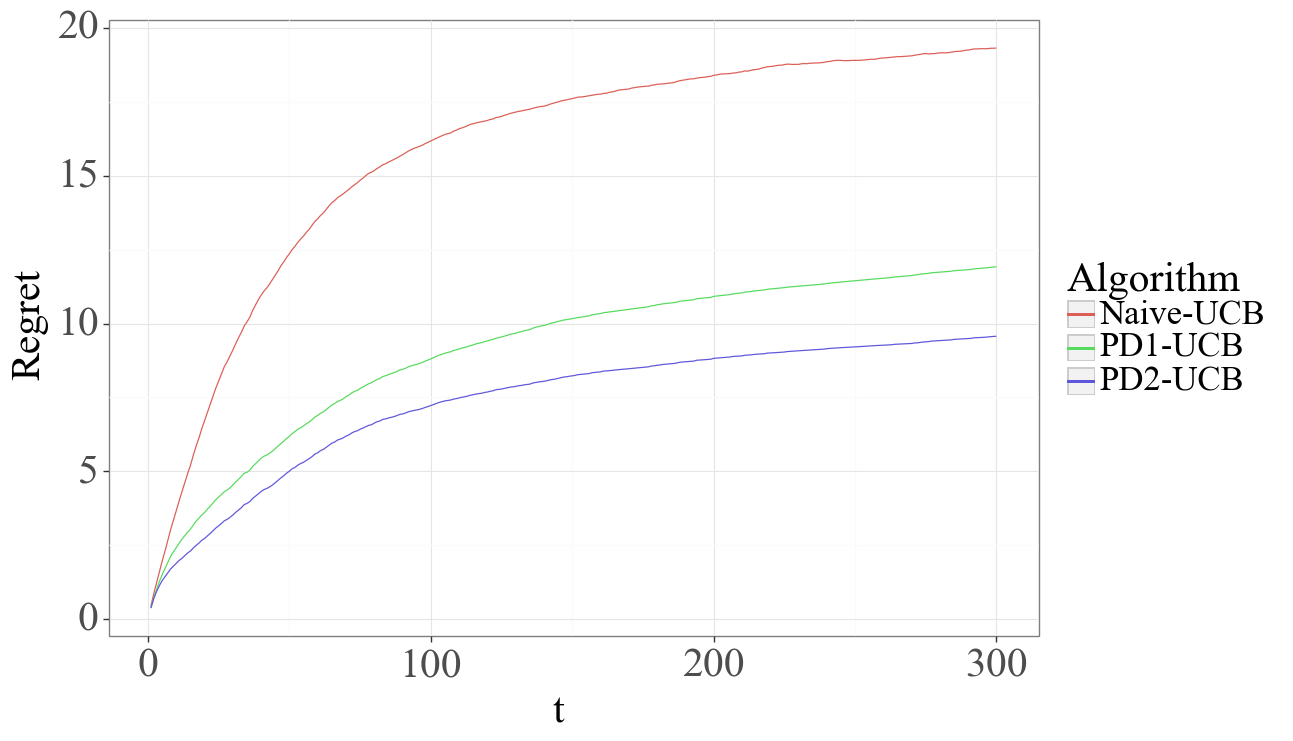}
\includegraphics[width=0.26\linewidth,keepaspectratio]{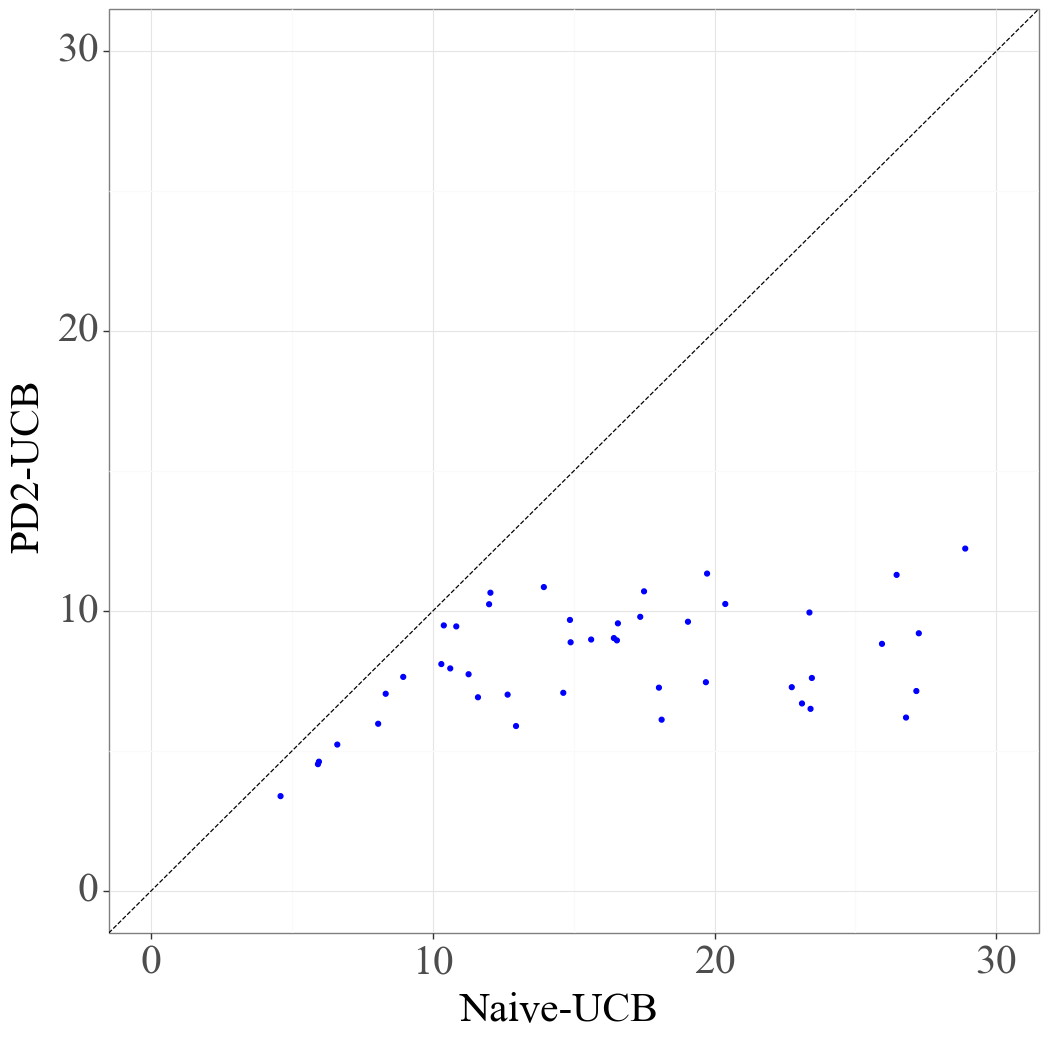}
\includegraphics[width=0.26\linewidth,keepaspectratio]{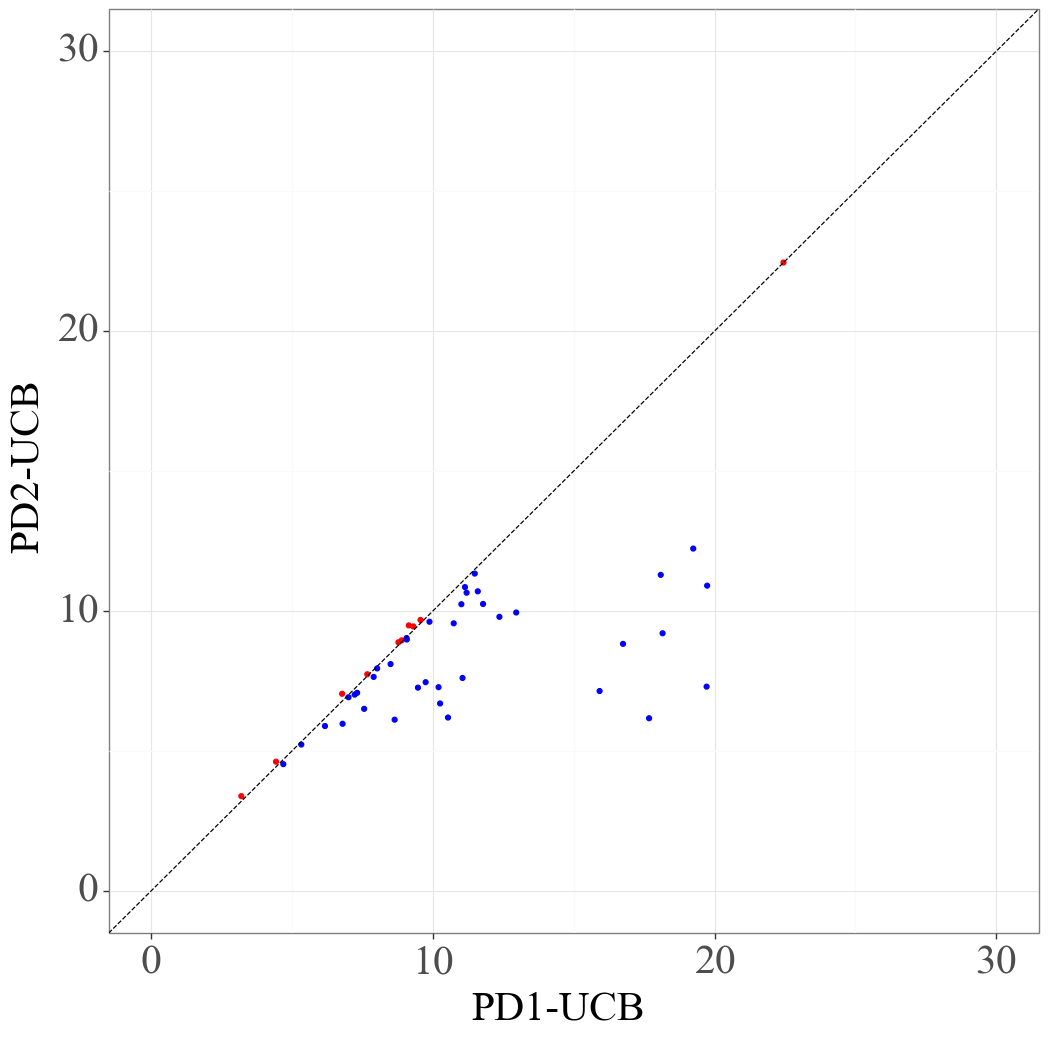}
\caption{Regret comparison under known $\mathbf{p}^*$ with $B=10$. Left: cumulative regret averaged over 50 instances (each with a unique $\theta_*$ value) each with 50 replications.
Middle and right: regret comparison between PD2-UCB, Naive-UCB, and PD1-UCB, at $T=300$;
each dot represents one instance averaged over 50 replications. 
}
\label{fig:regret_known_p}
\end{figure*}

\paragraph{Regret}
To analyze the regret of Algorithm~\ref{alg:learning-reward}, we 
first develop a high-probability confidence bound on the regularized least-square estimator of $\theta_*$ for the  revealer at time $t$ (Proposition~\ref{prop:concentration}).
We note that 
because of Constraint~\eqref{eq:learning-constraint}, the bandit learning loss relies \emph{only}  on the concentration of $\tilde C_t$.
We then bound the bandit learning loss ($\mathrm{BLL}_T$) associated with learning the unknown reward (Proposition~\ref{thm:bandit_learning}). Leveraging our bridging argument 
(Section \ref{sec:setup}), we 
prove our main result on bounding the regret by 
combining
the bandit learning loss (Proposition~\ref{thm:bandit_learning}) and the information reveal loss (Corollary~\ref{cor:competitive_with_learning}) together in Theorem~\ref{thm:mainresult}. Using the notation included in Appendix~\ref{app:proof} and with $\lambda=1/W^2$, we obtain Proposition~\ref{thm:bandit_learning} (proof in Appendix~\ref{append:proof_bandit_regret}).
Let $BLL_T=
\mathbb{E}[ V^{\text{AUX}} - V^{\text{ALG}} ] $, and also let
 $V^{\text{ALG}} = \sum_{t=1}^T (\langle
    \phi(S_t, \tilde A'_t) , \theta_* \rangle O^\text{ALG}_t+ \langle
     \bar\phi(\hat A_t) , \theta_* \rangle (1-O^\text{ALG}_t) )
$, where
$\{O^\text{ALG}_t\}_{t\in[T]}$ is chosen according to Algorithm~\ref{alg:learning-reward}, and $\tilde A'_t$ and $\hat A_t$ are the respective recommender's decision 
given the history is revealed or not. The bandit learning loss $BLL_T$ is defined as
\begin{align*}
&BLL_T:=\mathbb{E}\Big[\sum_{t=1}^T\langle\phi(S_t, A_t^*)O_t^\text{AUX} - \phi(S_t, \tilde A_t')O_t^\text{ALG}, \theta_*\rangle \\&\quad+\sum_{t=1}^T\langle\bar\phi(A^*)(1-O_t^\text{AUX}) - \bar\phi(\hat A_t)(1-O_t^\text{ALG}), \theta_*
\rangle\Big].
\end{align*}

Finally, we need the following assumption to bound the bandit learning loss of Algorithm~\ref{alg:learning-reward}:
\begin{assumption}\label{assumption:c_max}
Assume there exists a constant $c_{\max}$ such that $\|\bar\phi( \widetilde A_t)\|_{\widetilde V_{t}^{-1}(\lambda)} \leq c_{\max} \|\phi(S_t,  \widetilde A_t)\|_{\widetilde V_{t}^{-1}(\lambda)}$.
\end{assumption}

\begin{remark}[Existence of $c_{\max}$]\label{remark:c_max}
We observe that when the horizon is finite, the existence of such a constant $c_{\max}$ is implied by the fact that 
(i) $\widetilde V_{t}^{-1}(\lambda)$ 
is positive semidefinite, 
(ii) 
$\widetilde V_{t}^{-1}(\lambda)\preceq \widetilde V_{t-1}^{-1}(\lambda)$ for all time periods $t\in [T]$,
and (iii)  $\phi(k,a)$ is bounded for any $k\in K$ and $a\in\mathcal{A}$. 
When the horizon is infinite, 
Assumption~\ref{assumption:c_max} holds 
if the bandit algorithm does not converge to selecting a single action (we expect the reveal budget to grow with respect to the horizon length). 
\end{remark}





\begin{restatable}{prop}{banditregret}\label{thm:bandit_learning}
With probability $1-2\delta$, the bandit learning loss of the Algorithm~\ref{alg:learning-reward} is bounded by: 
\begin{align*}
\mathrm{BLL}_T 
&\leq
\max(c_{\max}, 1)\sqrt{8Td\gamma^2\log\left(\frac{d+TW^2L^2}{d}\right)}\\&+ W\sum_{t=1}^T \beta_t  
 + 2B\pi_{\max},
\end{align*} 
    where 
    $\gamma = 1+ \sqrt{2\log(\frac{1}{\delta}) + d\log(1+\frac{TW^2L^2}{d})}$, 
    $\beta_t = \mathcal{O}(1/(\sqrt{t}\log(B)))$, 
    $A^*_t =\arg\max_{a\in\mathcal{A}} \langle\theta_*, \phi(s_t, a)\rangle$,
 and $A^* = \arg\max_{a\in\mathcal{A}} \langle\theta_*, \sum_{k=1}^K \phi (k, a) \mathbf{p}_k\rangle$. 
\end{restatable}
For every budget level $B$ and horizon length $T$, there exists some constant $\alpha$
such that with $\beta_T = \alpha / (\sqrt{T}\log(B))$, Constraint~\eqref{eq:learning-constraint} 
satisfied at time $T$ without increasing $\beta_T$.
Thus, with  suitable choices of $\beta_t$'s, we achieve a sub-linear regret (i.e., $BLL_T\leq \mathcal{O}(d \sqrt{T} \log\left(T \right))$, see a more detailed discussion in Remark~\ref{remark:regret}).
{It is worth noting that the sublinearity of the regret is not the most crucial aspect of this problem. While it is possible to achieve sublinear regret by setting sufficiently large $\beta_t$'s,
this approach may result in a high constant in the regret. 
Therefore, in Section~\ref{sec:experiments}, we numerically
evaluate the performance of our algorithm in comparison to two benchmarks, 
demonstrating
the superior performance of our algorithm.}

Lastly, the regret of Algorithm~\ref{alg:learning-reward} (proof in Appendix~\ref{append:proof_main}) is bounded by combining the results of Corollary~\ref{cor:competitive_with_learning} and Proposition~\ref{thm:bandit_learning}. With $\gamma$ and $\beta_t$ defined above, and
 $\alpha = (1+\frac{1}{c-1})\frac{1}{\pi_{\min}}$, we have our main result:

\begin{restatable}{thm}{regret}\label{thm:mainresult}
With probability $1-2\delta$, the regret of Algorithm~\ref{alg:learning-reward} is bounded as follows: 
$\mathrm{Regret}_T
\leq 
\max(c_{\max}, 1)\sqrt{8Td\gamma^2\log\left(\frac{d+TW^2L^2}{d}\right)}+ W\sum_{t=1}^T \beta_t + 2B\pi_{\max} 
+ (1 - \alpha) 
\mathbb{E}[
V^{\text{CLV}}]$.
\end{restatable}

We note that 
$ \mathbb{E}[V^{CLV}] - \mathbb{E}[V^{AUX}]\leq ( 1 - \alpha )  
\mathbb{E}[
V^{\text{CLV}}]$ is in the order of $\mathcal{O}(B)$ and it does not depend on T. This is due to the fact that $\{O_t^\text{CLV}\}_{t=1}^T$ and $\{O_t^\text{AUX}\}_{t=1}^T$ differ in at most $B$ entries. Thus, $ \text{Regret}_T\leq \mathcal{O}(d \sqrt{T} \log(T )) + \mathcal{O}(B).$
\begin{remark}[Regret bound]\label{remark:regret}
In our regret upper bound, the last term $\left( 1 - \alpha \right) \mathbbm{E}[V^{\text{CLV}}]$ that comes from the online primal-dual subroutine is constant. The first two terms come from the bandit learning algorithm. We have to highlight how a sublinear regret for our algorithm can be achieved. Intuitively, from the definition of the bandit learning loss ($\mathrm{BLL}_T$), we observe that when $O_t = 1$ (i.e., classical regret of contextual bandits), we can readily obtain the sublinear regret. Also, when (1) $O_t = 0$ and (2) the optimal action of the recommender given the history $\mathcal{H}^{\mathrm{rec}}_T$ is the same as the one obtained using $\mathcal{H}^{\mathrm{rev}}_T$, then we can also achieve the sublinear regret of $\mathcal{O}(\sqrt{dT \log(T)})$. 
This implies that sublinearity can be achieved by 1) having a sufficiently high budget to explore all actions (i.e., we require $B>2|\mathcal{A}|$), and 2) revealing the $\mathcal{H}_t^{\mathrm{rev}}$ sufficiently late, i.e., when the optimal policy $A^*$ is still being learned.
We note that the latter can be achieved in Algorithm~\ref{alg:primal-dual} by 1) proper scaling of the revealing probability (through $u_{\max}$ so that we do not run out of budget before the horizon ends) and 2) the nature of the algorithm: when $t$ increases, the marginal value of a context ($u_{s_t}^* - v^* - y$) required to reveal information also increases. While in Algorithm~\ref{alg:primal-dual-step-constraint}, the latter can be achieved with proper initialization of the parameters $\beta_t$'s. 
\end{remark}

\section{EXPERIMENTS}
\label{sec:experiments}

We conduct experiments on both synthetic  and real-world datasets to demonstrate the effectiveness of the proposed algorithm in minimizing regret. To show the benefit of adding the novel learning constraint, we compare two variants of the proposed algorithm: 1) PD1-UCB (UCB with primal-dual without the learning constraint, i.e., replacing the subroutine in Algorithm~\ref{alg:learning-reward}  with Algorithm~\ref{alg:primal-dual}) and 2) PD2-UCB (Algorithm~\ref{alg:learning-reward}). We benchmark our algorithms with a naive UCB approach that reveals contexts with a fixed probability of $B/T$ (naive-UCB). The experiments are repeated $50$ times, and the cumulative regret, revealing probability, and competitive ratio are averaged and presented.

\paragraph{Synthetic Experiments Setup}
We consider a linear contextual bandit setting with $10$ discrete one-dimensional contexts, i.e., $|\mathcal{S}|=10$.
For each context $S_k$, 
we sample it
according to $\mathbf{p}_k$, which is drawn from a uniform distribution $U(0,1)$ and scaled by $\sum_k \mathbf{p}_k$,
For each \emph{instance}, every coordinate of the true reward parameter $\theta_* \in \mathbb{R}^d$ is sampled from $U(0,1)$.
The reward for a selected action $A_t$ in each instance is then generated by $X_t = \langle \theta_*, \phi(S_t, A_t)\rangle + \eta_t$, 
where $\phi(S_t, A_t)$ includes a one-hot vector (of length $|\mathcal{A}|-1$) 
denoting action $A_t$, a variable denoting context $S_t$, and 
$|\mathcal{A}|-1$
interaction terms. The noise $\eta_t$ is sampled from $N(0,\sigma^2)$ with $\sigma=0.1$. 
We set the number of actions to be $|\mathcal{A}|=5$, 
and the length of the time horizon to be $T=300$. At each step, $\tilde u_{s_t}^t$'s and $\tilde v^t$'s are normalized using $u_{\max}$ and $u_{\min}$. Throughout this section, we choose $\beta_t = 1.2\Delta_{\min}\log(10)\sqrt{10}/(\sqrt{t}\log(B))$, where $\Delta_{\min}:=\min_{k\in [K], a\in\mathcal{A}, a'\in\mathcal{A}\setminus \{a\}}\|\phi(k,a)-\phi(k,a')\|$.

\paragraph{Competitive Ratio} We first numerically inspect the empirical competitive ratio  of PD1-UCB and PD2-UCB under one instance
in Table~\ref{table:competitve ratio}, calculated by $\mathbbm{E}[V^{\text{AUX}}]/\mathbbm{E}[V^{\text{CLV}}]$. For PD1-UCB, we replace the sequence of $\{O_t\}_{t\in[T]}$ in $V^{\text{AUX}}$ by that chosen according to Algorithm~\ref{alg:primal-dual}.
The competitive ratios 
are for ground truth $\theta_*$
averaged over 
200 context arrival sequences.
We observe that both PD1-UCB and PD2-UCB have a competitive ratio that is higher than $1-1/e$ as stated in Corollary~\ref{cor:competitive_with_learning}. In addition, PD1-UCB has a slightly higher average competitive than PD2-UCB as expected, since at each step, PD2-UCB will like to increase $o_t$ to satisfy Constraint~\eqref{eq:learning-constraint}.

\paragraph{Regret  under Known $\mathbf{p}^*$}
We present the cumulative regret when $B=10$ (Figure~\ref{fig:regret_known_p}), $B=20$ (Figure~\ref{fig:scatter_20}), and $B=30$ (Figure~\ref{fig:scatter_30}).
We observe that PD2-UCB outperforms Naive-UCB and PD1-UCB 1) almost instance-wise (the dots above the 90 degree line in Figure~\ref{fig:regret_known_p} right is most likely due to noise), 2) by large margins on many instances ($\theta_*$ values).  
In addition, the benefit of our algorithm is greatest when the budget is low. 
We note that the regrets are in general increasing with respect to the budget
since
the optimal strategy for the clairvoyant is changing with respect to $B$.
We include additional experiments where $\mathbf{p}^*$ is unknown in Figures~\ref{fig:contextb10}, \ref{fig:contextb20}, and \ref{fig:contextb30}, and observe similar results.





\paragraph{Real-World Experiments}
%
Our problem is motivated by the \emph{Oralytics} mobile health application~\citep{trella2022designing}. 
For the purpose of illustrating the concept,
we use the ROBAS 2~\citep{shetty2020scalable} and ROBAS 3~\citep{trella2022designing} datasets to simulate a scenario involving $10$ users ($N = 10$) over 140 decision points ($T = 140$). This design follows the simulation environment presented in \cite{trella2022designing} but excludes the integration of delayed effects. The primary objective of this study is to assess the effectiveness of the proposed algorithm in terms of maximizing the cumulative brushing quality, a chosen reward metric. This total brushing quality is represented by $\sum_{t=1}^TQ_{i,t}$, where $Q_{i,t}$ denotes a non-negative measure of brushing quality observed following each decision point.
Due to the space limitation, we include the details of the simulation environment
in Appendix~\ref{app:robas3}.


To simulate a scenario with budgeted information disclosure, an additional budget of $B=30$ is introduced. At each decision instance $t$, the revealer has the choice to disclose the historical record up to that point. If the decision is against revealing, the recommender only has access to partial historical information, such as the time of day and weekend indication. We conducted $50$ simulation trials. The average reward across multiple time points $\frac{1}{N}\sum_{i=1}^N\frac{1}{t_0}\sum_{t=1}^{t_0}Q_{i,t}$, a metric proposed by \cite{trella2022designing},  is reported.

\begin{figure}[!t]
    \centering    \includegraphics[width=\linewidth,keepaspectratio]{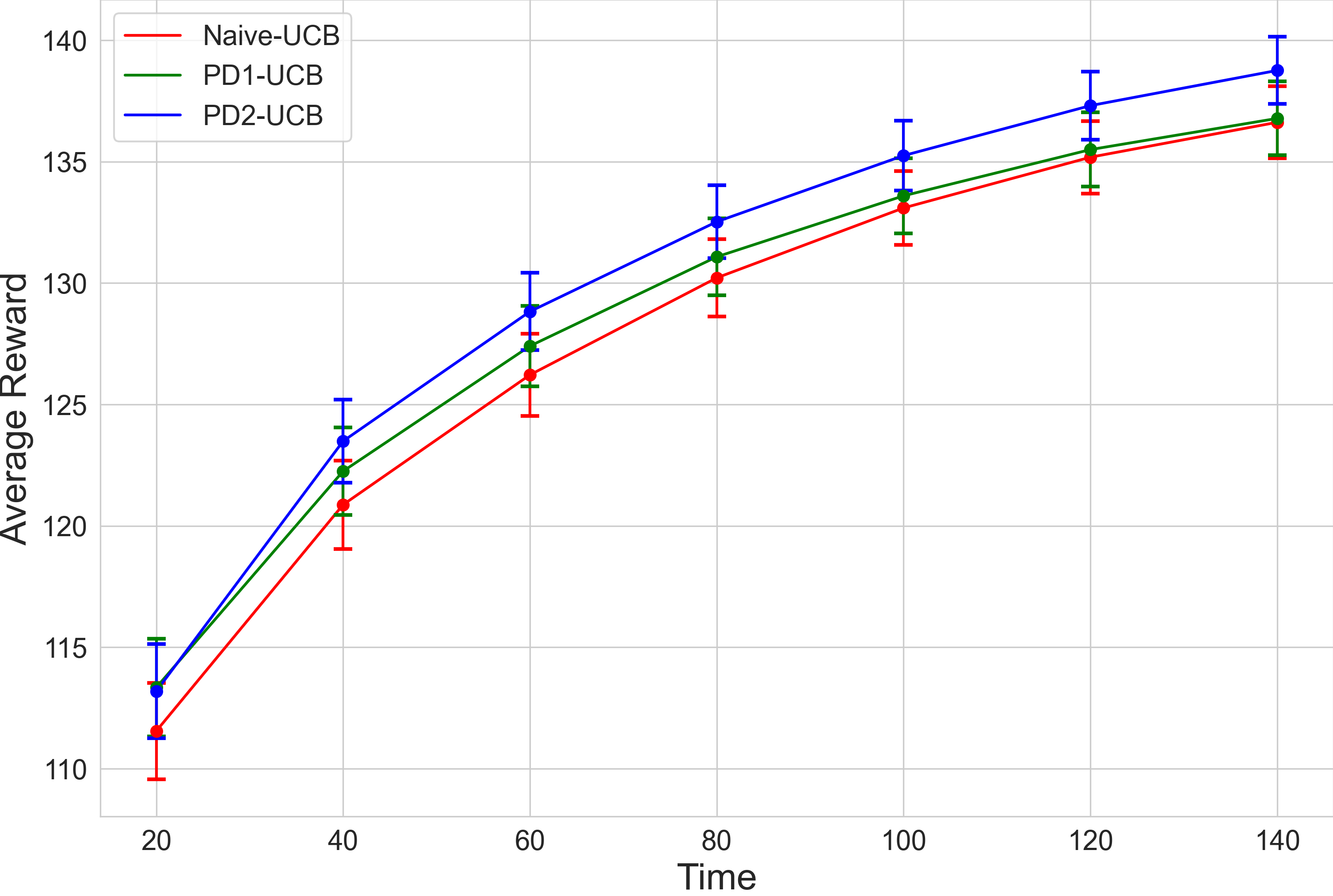}
    \caption{
    Average reward comparison on the ROBAS3 dataset. The y-axis is the mean and $\pm 1.96 \cdot$ standard error of the average user rewards $\left(\bar{R}=\frac{1}{10} \sum_{i=1}^{10} \frac{1}{t_0} \sum_{s=1}^{t_0} R_{i, s}\right)$ for decision times $t_0 \in[20,40,60,80,100,120,140]$ across $50$ experiments and 10 users. Standard error is $\frac{\sum_i^{10}\hat{\sigma}_i}{10\sqrt{50}}$ where $\hat{\sigma}_i$ is the user-specific standard error. 
    }
    \label{fig:reward}
\end{figure}

In Figure~\ref{fig:reward}, we observe that PD2-UCB outperforms PD1-UCB and Naive-UCB in terms of mean performance. However, due to 1) limited trial numbers and 2) the limited number of actions, the confidence intervals of the three methods overlap. We  defer larger-scale experiments to future work.

\paragraph{Conclusion and Future Works}
We develop a novel learning and optimization 
algorithm
for the problem of jointly optimizing the timing of the pro-treatment actions and personalized treatments.
This work sheds light on a new direction in online learning and optimization theory that holds promise for advancing digital health interventions.
Potential future extensions include:
a rigorous extension of our algorithm to other reinforcement learning algorithms,  
and incorporating a context predictor or noisy context.
\section*{Acknowledgement}
This work is supported by  NIH P41EB028242 and NIH P50DA054039.
\bibliographystyle{plainnat}
\bibliography{citation}

\newpage
\clearpage
\section*{Checklist}
 \begin{enumerate}

 \item For all models and algorithms presented, check if you include:
 \begin{enumerate}
   \item A clear description of the mathematical setting, assumptions, algorithm, and/or model. [Yes], please see Section 2 for the mathematical setting and assumptions, and Sections 3 and 4 for algorithms and model. 
   \item An analysis of the properties and complexity (time, space, sample size) of any algorithm. [Yes], we conducted the regret analysis of our algorithms; see Propositions 1 and 2, and Theorem 1. 
   \item (Optional) Anonymized source code, with specification of all dependencies, including external libraries. [Yes], specifications of all dependencies, including external libraries, have been set in our code. 
 \end{enumerate}

 \item For any theoretical claim, check if you include:
 \begin{enumerate}
   \item Statements of the full set of assumptions of all theoretical results. [Yes], we explained our assumptions in Sections 2, 3 and 4. 
   \item Complete proofs of all theoretical results. [Yes], all our proofs are provided in Sections C, D, E, and F of the Appendix. 
   \item Clear explanations of any assumptions. [Yes], we discussed our assumptions in Sections 2, 3 and 4. 
 \end{enumerate}

 \item For all figures and tables that present empirical results, check if you include:
 \begin{enumerate}
   \item The code, data, and instructions needed to reproduce the main experimental results (either in the supplemental material or as a URL). [Yes], our codes are included in the supplemental material. 
   \item All the training details (e.g., data splits, hyperparameters, how they were chosen). [Yes], these  details have been included in our codes. 
         \item A clear definition of the specific measure or statistics and error bars (e.g., with respect to the random seed after running experiments multiple times). [Yes], the definition of the error bar is included for Figure 3.
         \item A description of the computing infrastructure used. (e.g., type of GPUs, internal cluster, or cloud provider). [Yes], the detailed information is included in Appendix G.
 \end{enumerate}

 \item If you are using existing assets (e.g., code, data, models) or curating/releasing new assets, check if you include:
 \begin{enumerate}
   \item Citations of the creator If your work uses existing assets. [Yes], our real-world experiments uses an existing public simulation test bed, and we cited it in the main body.
   \item The license information of the assets, if applicable. [Not Applicable].
   \item New assets either in the supplemental material or as a URL, if applicable. [Not Applicable].
   \item Information about consent from data providers/curators. [Not Applicable].
   \item Discussion of sensible content if applicable, e.g., personally identifiable information or offensive content. [Not Applicable].
 \end{enumerate}

 \item If you used crowdsourcing or conducted research with human subjects, check if you include:
 \begin{enumerate}
   \item The full text of instructions given to participants and screenshots. [Not Applicable].
   \item Descriptions of potential participant risks, with links to Institutional Review Board (IRB) approvals if applicable. [Not Applicable].
   \item The estimated hourly wage paid to participants and the total amount spent on participant compensation. [Not Applicable].
 \end{enumerate}

 \end{enumerate}

\newpage
\renewcommand*{\theHprop}{\theprop}
\setcounter{remark}{0}
\renewcommand{\theremark}{\thesection.\arabic{remark}}
\setcounter{figure}{0}
\renewcommand{\thefigure}{\thesection.\arabic{figure}}
\setcounter{lem}{0}
\renewcommand{\thelem}{\thesection.\arabic{lem}}
\setcounter{algorithm}{0}
\renewcommand{\thealgorithm}{\thesection.\arabic{algorithm}}
\setcounter{prop}{0}
\renewcommand{\theprop}{\thesection.\arabic{prop}}
\setcounter{table}{0}
\renewcommand{\thetable}{\thesection.\arabic{table}}
\setcounter{remark}{0}
\renewcommand{\theremark}{\thesection.\arabic{remark}}
\setcounter{assumption}{0}
\renewcommand{\theassumption}{\thesection.\arabic{assumption}}
\appendix
\onecolumn
\section*{Appendix}
\section{Additional Motivating Examples: Addiction Treatment}\label{app:motivating_examples}
The objective of this section is to show an additional setting where our problem formulation can be useful. In the example of addiction treatments, the recommender could be the text message that the patient receives on their phone (produced by an RL agent sitting in the cloud). A revealer could be a staff member who observes all the information collected by the wearables.
Here, we assume that
completing the self-reports regularly is a mechanism that researchers use to ensure that the patient opens the App.
Similar to the commercial sensor example, we consider the case where the patient does not complete the required self-reports (pro-treatment action), and a staff member could call the patient or their family to reveal all their recent health status.  In this example, the context of the patient is partially observed. The sensor data is always observed by the revealer, but the survey data can only be obtained unless the staff reveals it. Once the patient completes the self-report (and thus opens the App), the recommender obtains both the sensor and survey data.
We note that the staff could use the sensor data to decide whether to reach out to the patient. 
In summary, when the patient does not complete the self-report, the recommender observes both the sensor data and the self-reports up to the last reveal. The staff uses the sensor data to decide when to reach out to the patient to reveal. Once revealed, the recommender observes all sensor and self-report data. 

\section{Extension to Partially Observed Context}\label{append:partial_context}
In many digital health applications, patients need to both (1) allow, passively, sensor  data through a data-collecting device to be collected, and (2) complete regular self-reports. 

Our problem setting can be extended to the setting, where the revealer only observes a part of the context. Namely, we can partition each state into two components: $S_t = [S_t^1, S_t^2]$. WLOG, let $S_t^1$ be the part of the state that is \emph{always} observed by the revealer at each time $t$, and let $S_t^2$ be the part of the state that can \emph{only} be observed when the revealer takes the action $O_t=1$.
For example, $S_t^1$ could correspond to the data collected by sensors, and $S_t^2$ could correspond to the self-reports in the above example.

Let $\ell(t)$ be the time of the last reveal. At each decision time $t$, the revealer observes the history $\mathcal{H}_t^{\mathrm{rev}} = \{ A_1,...,A_{t-1}, X_1, ..., X_{t-1}, O_1, ...,O_{t-1}, S_1, ...,  S_{\ell(t)}, S_{\ell(t)+1}^1,  ..., S_t^1\}$. Then, the reward that we observe at each time step can be decomposed as $X_t = \langle \phi(S^1_t, S^2_t, A_t), \theta_*\rangle +\eta_t$, where $\theta_*\in \mathbb{R}^d$ is an \emph{unknown} true reward parameter,  $\phi: \mathcal{S}\times \mathcal{A}\mapsto \mathbb{R}^d$ is a \emph{known} feature mapping, and the noise $\eta_t$ is conditional mean-zero $1$-sub-Gaussian. 

At each time, the revealer decides whether to reveal $\mathcal{H}_t^{\mathrm{rev}}$ to the recommender. If the revealer decides to take an action, i.e., $O_t=1$, then the
revealer additionally observes $\{S_{\ell(t)+1}^2, ..., S_t^2\}$, and the
recommender observes $ \mathcal{H}_t^{\mathrm{rev}} = \{ A_1,...,A_{t-1}, X_1, ..., X_{t-1}, O_1, ...,O_{t-1}, S_1, ...,  ..., S_t\}$. Otherwise, the recommender observes $\mathcal{H}_t^{\mathrm{rec}} = \{ A_1,...,A_{\ell(t)-1}, X_1, ..., X_{\ell(t)-1}, O_1, ...,O_{\ell(t)-1}, S_1, ...,  S_{\ell(t)}\}$. 

Let $M$ be the cardinality of  the set $S_t^2$, i.e., $S_t^2\in[M]$. 
Moreover, let $Q$ be the cardinality of the set $S_t^1$, i.e., $S_t^1\in[Q]$.
Assume that the
state $S_t^1$ is i.i.d. drawn from distribution $\mathbf{p}'$, and the
state $S_t^2$ is drawn i.i.d. from the conditional distribution  $\mathbf{p}''(S_t^1)=\mathbb{P}(S_t^2|S_t^1)$. 
In \eqref{pblm: clairvoyant}, we then have that $u^*_{s_t^1} =\argmax_{a\in\mathcal{A}}\langle\sum_{m=1}^M\phi(s_t^1, m, a)\mathbf{p}''{(s_t^1)}, \theta_*\rangle$, and $v^* = \argmax_{a\in\mathcal{A}}\langle\sum_{q=1}^Q\sum_{m=1}^M \phi(q, m, a)\mathbf{p}'_q{\mathbf{p}''({q})}_m,\theta_*
\rangle,$ and the rest of problem follows. In \eqref{Eq: clairvoyant-Learning}, the added constraint remains the same since the recommender in this problem setting does not get additional information than the current setting in the main body. Since observing partial information only affects the decision of the revealer, the learning part of the problem remains the same.




\section{Online Primal-Dual Algorithm without Learning Constraint}
\label{append:subsec_without_learning}
In this section, we lay out the road map for deriving the learning constraint and proving Proposition~\ref{prop:primal-dual-learning-constraint}.
In Algorithm \ref{alg:primal-dual}, we 
only take the budget into account 
and 
derive its competitive ratio
in Proposition~\ref{prop:primal-dual}. 
We 
first write
down the dual problem of \eqref{pblm: clairvoyant} as:
\begin{align}
\label{pblm: dual-clairvoyant}
\min_{y,z_t} \;\; 
By&+ \sum_{t=1}^T z_t
\notag \\ 
s.t. \quad &      y + z_t \geq u_{s_t}^* - v^*,  \; \forall t \in [T]
    \notag\\ 
    & y, z_t \geq 0, \; \forall t \in [T].
\tag{\textrm{\emph{Clairvoyant Dual}}}
\end{align}

At the margin, $y$ is the marginal value of the budget ($y\delta$ corresponds to how the value of the optimal solution to the primal change if we were to change the budget $B$ by $\delta$), and $z_t$ is the marginal value for revealing the context at time step $t$. We note that in \eqref{pblm: dual-clairvoyant}, we have a separate constraint for each $z_t$. Thus, when we do not know the context arrival sequence ahead of time (as in the clairvyant problem), the constraints in \eqref{pblm: dual-clairvoyant} are arriving in an online fashion (one-by-one). This provides a nice framework to analyze the online context arrival sequence.  

Let $u_{\max} = \max_{s\in S} u_{s_t}^* $ and $\pi_{\min} = \min_{s\in S} \max(u_{s_t}^* - v^*, 0)$, where $\pi_{\min}$ is the smallest positive difference between $u_{s_t}^* $ and $ v^*$.
We assume that an \emph{upper bound} on $u_{\max}$ is known to the algorithm by applying domain knowledge.    
\emph{Without loss of generality} (WLOG), we assume that $u_{s_t}^* - v^* \leq 1 $ for all $s_t\in \mathcal{S}$, since otherwise we could scale $u_{s_t}^* - v^*$ by $u_{\max}$ for all $s_t\in \mathcal{S}$. 
%
%
We outline the online primal-dual algorithm in Algorithm \ref{alg:primal-dual}. This algorithm 
provides a feasible solution to both the primal and dual problems, and 
only depends on the history it has observed so far. We show:

\begin{restatable}{prop}{primaldual}
\label{prop:primal-dual}
    For any $u_{s_t}^*$, $v^*$, and context arrival sequence, the value of the objective function of Algorithm~\ref{alg:primal-dual} is at least $\pi_{\min}(1 - 1/c)$ times that of ~\eqref{pblm: clairvoyant}.
\end{restatable}
The proof of Proposition~\ref{prop:primal-dual} (\S~\ref{app:proof-primal-dual}) consists of showing
that  Algorithm~\ref{alg:primal-dual} is (1) both primal and dual feasible, and (2) in each iteration (day), the ratio between the change in the primal and dual objective functions is bounded by $\pi_{\min}(1 - 1/c)$. By weak duality,  this implies that Algorithm~\ref{alg:primal-dual} is $\pi_{\min}(1 - 1/c)$-competitive. Moreover, note that when the ratio $\tau = 1/B$ tends to zero, the competitive ratio of the algorithm approaches the best-possible competitive ratio of $(1-1/e)$~\citep{buchbinder2007online}.
In addition, we note that as $t$ increases, Algorithm~\ref{alg:primal-dual} increasingly favors revealing the context information when the expected difference in rewards, $u^*_{s_t} - v^*$ is large. Thus, even though $\pi_{\min}$ appears in our competitive ratio, the performance of our algorithm is much better in practice. We illustrate this in \S~\ref{sec:experiments}.

We can use Algorithm~\ref{alg:primal-dual} as a subroutine in our learning algorithm (Alg.~\ref{alg:learning-reward}) to decide the revealing probability,
where we plug in the empirical estimates for $\{u_{s}^*\}_{s\in \mathcal{S}}$ and $v^*$ obtained by the \emph{revealer}\footnote{If we plug in the estimates obtained by the \emph{recommender} instead, i.e, $\{\hat{u}_{s}^0\}_{s\in \mathcal{S}}$ and $\hat{v}^0$, then our algorithm will not be robust with respect to parameter initialization.
Take one extreme example, in which 
the confidence intervals of 
$\{\hat{u}_{s}^0\}_{s\in \mathcal{S}}$ and $\hat{v}^0$ are initialized to be the same, then the current algorithm will \emph{never} choose to reveal the context. Thus, the recommender will never learn the optimal interventions.} using $\mathcal{H}_t^{\mathrm{rev}}$ at time $t$, denoting them $\{\tilde{u}_{s}^t\}_{s\in \mathcal{S}}$ and  $\tilde{v}^t$, respectively. 
Since the recommender has \emph{no} access to $\{\tilde{u}_{s}^t\}_{s\in \mathcal{S}}$ and $\tilde{v}^t$, as a subroutine, \eqref{pblm: clairvoyant} lacks a mechanism to connect
the quality of the estimates that the \emph{recommender} has at time $t$ to
the revealing decision $o_t$. Ideally, we would like $o_t$ to increase as the time since the last reveal increases.
This leads to the Constraint
\eqref{eq:learning-constraint} in \S~\ref{sec:auxiliary}.
\begin{algorithm}[t]
\caption{Online Primal-Dual Algorithm: From Clairvoyant to Auxiliary Revealer}
\label{alg:primal-dual}
\begin{algorithmic}
\State \textbf{Input:} $B, \{u_{s}^*\}_{s\in \mathcal{S}}, v^*, c= (1+1/B)^B$
\State \textbf{Initialize:} $y \leftarrow 0$
\State \text{On each day, a new context  $s_t$  arrives,} \\
\text{and the $t$-th constraint in the dual problem arrives}
\If{$y < 1$ and $u_{s_t}^* -v^* - y >0$}
\State $z_t = u_{s_t}^* - v^* - y$
\State $o_t = \min(u_{s_t}^* - v^*, B-\sum_{i=1}^{t-1}o_i)$
\State $y \leftarrow y(1+{o_t}/{B}) + {o_t}/{\left((c-1)\cdot B\right)}$
\Else
\State $z_t = 0$
\State $o_t = 0$
\EndIf
\end{algorithmic}
\end{algorithm}

\subsection{Proof of Proposition~\ref{prop:primal-dual}}
\label{app:proof-primal-dual}
\primaldual*
\begin{proof}
First, we set $z_t = u_{s_t}^* - v^* - y$ whenever $y<1$ and $u_{s_t}^* - v^* - y > 0$, this implies that the solution is dual feasible. To show primal feasibility, we need to show $\sum_{t=1}^T o_t \leq B$. We prove this by showing that $y>1$ for at most $B$ updates. Let $\mathcal{I}$ be the set of time indices that we reveal the context. We want to show that when  $\sum_{i\in \mathcal{I}}o_t \geq B$, then $y\geq 1$. We use $y^{(i)}$ to denote the value of $y$ in the $i$-th iteration, where $i\in\mathcal{I}$. 

We will prove the following bound: 
\begin{equation}
\label{eq:induction}
    y^{(i)}\geq \frac{1}{c-1}\left(c^\frac{{\sum_{t=1}^i o_t}}{B}-1\right).
\end{equation}
Thus, whenever $\sum_{i\in \mathcal{I}}o_t \geq B$, we have $y\geq 1$. We will prove by induction. 

Base case $i=0$: $y^{(0)}\geq 0$. Equation~\eqref{eq:induction} is trivially true.

Induction step: Assume that Equation (\ref{eq:induction}) is true for step $i-1$, then show for step $1$. We have:
\begin{align*}
    y^{(i)} &= y^{(i-1)}\left(1+
    \frac{u_{s_i}^* -v^*}{B}\right) + \frac{u_{s_i}^* - v^*}{(c-1) B} \\
    &\geq \frac{1}{c-1}\left[c^{\frac{\sum_{t=1}^{i-1}o_t}{B}}-1\right]\left(1+
    \frac{u_{s_i}^* -v^*}{B}\right) + \frac{u_{s_i}^* - v^*}{(c-1) B} \\
    &= \frac{1}{c-1}\Big[c^{\frac{\sum_{t=1}^{i-1}o_t}{B}}\left(1+
    \frac{u_{s_i}^* -v^*}{B}\right)-1 -
    \frac{u_{s_i}^* -v^*}{B}+ \frac{u_{s_i}^* - v^*}{ B}\Big]\\
    &\geq 
    \frac{1}{c-1}\left[c^{\frac{\sum_{t=1}^{i-1}o_t}{B}}c^{\frac{u_{s_i}^* -v^*}{B}}-1 \right] \\
    &= \frac{1}{c-1}\left[c^{\frac{\sum_{t=1}^{i}o_t}{B}}-1 \right],
\end{align*}
where the second inequality follows from the induction hypothesis, and the last inequality follows from the fact that $\ln{(1+m)}/m \geq \ln{(1+n)}/n$ for all $0\leq m\leq n \leq 1$, where $m=\frac{u_{s_i}^* -v^*}{B}$, and $n=\frac{1}{B}$. Thus, we have $c = (1+{1}/{B})^{B}$. 

Next, we will show that the ratio between the change in the dual and primal objective functions is bounded by $(1+1/(c-1))\pi_{\min}^{-1}$. If $y<1$ and $u_{s_t}^* - v^* - y > 0$, then the primal objective function increases by $(u_{s_t}^* - v^*)^2$, and the increase in the dual objective function is $B\Delta y + z_t = (u_{s_t}^* - v^*)y + (u_{s_t}^* - v^*)/(c-1) + u_{s_t}^* - v^* - y = (u_{s_t}^* - v^*)/(c-1) + (u_{s_t}^* - v^*-1)y$. Thus, the ratio between the change in the dual and primal is 

\begin{align*}
&\frac{\Delta \mathrm{Clairvoyant \;Dual}}{\Delta\mathrm{Clairvoyant}} =
\frac{(u_{s_t}^* - v^*)/(c-1) + u_{s_t}^* - v^* + (u_{s_t}^* - v^*-1)y}{(u_{s_t}^* - v^*)^2}\leq \left(1+\frac{1}{c-1}\right)\frac{1}{\pi_{\min}}.
\end{align*}

\end{proof}

\section{Online Primal-Dual with Learning Constraint}
\subsection{Clairvoyant with Learning (Primal)}\label{app:learning_primal}
\begin{align}
\label{Eq: clairvoyant-Learning}
\max_{o_t} \quad & \sum_{t=1}^T o_t\cdot u_{s_t}^* -o_t\cdot v^* \notag\\
s.t. \quad &  \sum_{t=1}^T o_t \leq B,  \tag{\textrm{\emph{Modified CLV}}} \\ 
\quad & - \left\|\bar\phi(\tilde a_t) -\bar\phi(\hat a_t)\right\|_2\mathbbm{1}(\hat a_t\neq \tilde a_t)o_t \notag\\
&\leq \beta_t - \left\|\bar\phi(\tilde a_t) -\bar\phi(\hat a_t)\right\|_2\mathbbm{1}(\hat a_t\neq \tilde a_t),\; \forall t \in [T] \notag\\
\quad & o_t \in [0,1], \; \forall t \in [T]. \notag
\end{align}

\subsection{Proof of Proposition~\ref{prop:primal-dual-learning-constraint}}\label{proof:prop-primal-dual-learning}
\primaldualconstraint*
\begin{proof}
First, we observe that $z_t = u_{s_t}^* - v^* - y + \left\|\bar\phi(\tilde a_t) -\bar\phi(\hat a_t)\right\|_2\mathbbm{1}(\hat a_t\neq \tilde a_t) e_t$ whenever $y<1$ and $u_{s_t}^* - v^* - y + \left\|\bar\phi(\tilde a_t) -\bar\phi(\hat a_t)\right\|_2\mathbbm{1}(\hat a_t\neq \tilde a_t) e_t> 0$, this implies that
$z_t\geq 0$. By construction, $e_t\geq 0 \; \forall t\in[T]$. Thus,
the solution is dual feasible.

To show primal feasibility, we first need to show $\sum_{t=1}^T o_t \leq B$.
We prove this by showing that when  $\sum_{i=1}^{t}o_i \geq B$, then $y\geq 1$. We use $y^{(i)}$ to denote the value of $y$ in the $i$-th iteration, where $i\leq t$. 

We will prove the following bound:
\begin{equation}
\label{eq:induction-robustness}
    y^{(i)}\geq \frac{1}{c-1}\left(c^\frac{{\sum_{j=1}^i o_j}}{B}-1\right).
\end{equation}
Thus, whenever $\sum_{i=1}^t o_i \geq B$ for some $t$, we have $y\geq 1$. We will prove this by induction. 

Base case $i=0$: $y^{(0)}\geq 0$. Equation~\ref{eq:induction-robustness} is trivially true.

Induction step: assume Equation (\ref{eq:induction-robustness}) is true for step $i-1$, show for step $1$. We have
\begin{align*}
    y^{(i)} &= y^{(i-1)}\left(1+
    \frac{o_i}{B}\right) + \frac{o_i}{(c-1) B} \\
    &\geq \frac{1}{c-1}\left[c^{\frac{\sum_{j=1}^{i-1}o_j}{B}}-1\right]\left(1+
    \frac{o_i}{B}\right) + \frac{o_i}{(c-1) B} \\
    &= \frac{1}{c-1}\left[c^{\frac{\sum_{j=1}^{i-1}o_j}{B}}\left(1+
    \frac{o_i}{B}\right)-1-
    \frac{o_i}{B}+ \frac{o_i}{ B}\right]\\
    &\geq 
    \frac{1}{c-1}\left[c^{\frac{\sum_{j=1}^{i-1}o_j}{B}}c^{\frac{o_i}{B}}-1 \right] = \frac{1}{c-1}\left[c^{\frac{\sum_{j=1}^{i}o_j}{B}}-1 \right],
\end{align*}
where the second inequality follows from the induction hypothesis, and the last inequality follows from the fact that $\ln{(1+m)}/m \geq \ln{(1+n)}/n$ for all $0\leq m\leq n \leq 1$, where $m=\frac{o_i}{B}$, and $n=\frac{1}{B}$. Thus, we have $c = (1+{1}/{B})^{B}$. 


Next, we need to show that $\left\|\bar\phi(\tilde a_t) -\bar\phi(\hat a_t)\right\|_2\mathbbm{1}(\hat a_t\neq \tilde a_t)(1-o_t)\leq \beta_t\; \forall t\in [T]$. We first observe that the constraint $u_{s_t}^*>{v}^*$ is critical for the last step of the primal-dual algorithm where we need to bound the ratio between the change in the primal and dual objective functions. Case 1a): $\hat a_t = \tilde a_t$ or $\beta_t \geq \left\|\bar\phi(\tilde a_t) -\bar\phi(\hat a_t)\right\|_2$.  In these cases, this constraint is automatically satisfied and thus we set
$e_t=0$. Case 1b): $u_{s_t}^*\leq {v}^*$. In this case, our algorithm updates $\beta_t $ to $\left\|\bar\phi(\tilde a_t) -\bar\phi(\hat a_t)\right\|_2$, and the constraint is satisfied.

Case 2: $\hat a_t \neq \tilde a_t$, $\beta_t < \left\|\bar\phi(\tilde a_t) -\bar\phi(\hat a_t)\right\|_2$, and $u_{s_t}^*> {v}^*$. In this case, $e_t = 
\frac{1}{\left\|\bar\phi(\tilde a_t) -\bar\phi(\hat a_t)\right\|_2}-\frac{\beta_t}{\left\|\bar\phi(\tilde a_t) -\bar\phi(\hat a_t)\right\|_2^2}$. 
Below, we will show that Equation~(\ref{eq:learning-constraint}) holds for all 3 cases.  

 
 Case 2a):  $1\leq \frac{\beta_t}{\left\|\bar\phi(\tilde a_t) -\bar\phi(\hat a_t)\right\|_2}$. This case will never happen because $\beta_t < \left\|\bar\phi(\tilde a_t) -\bar\phi(\hat a_t)\right\|_2$.
 
 Case 2b): $1 > \frac{\beta_t}{\left\|\bar\phi(\tilde a_t) -\bar\phi(\hat a_t)\right\|_2}$ 
 and $u_{s_t}^* - v^* -y + 1-\frac{\beta_t}{\left\|\bar\phi(\tilde a_t) -\bar\phi(\hat a_t)\right\|_2}>0$. In this case, $e_t = \frac{1}{\left\|\bar\phi(\tilde a_t) -\bar\phi(\hat a_t)\right\|_2}-\frac{\beta_t}{\left\|\bar\phi(\tilde a_t) -\bar\phi(\hat a_t)\right\|_2^2}$ and $o_t = \min (u_{s_t}^* -  v^* + 1 - \frac{\beta_t}{\left\|\bar\phi(\tilde a_t) -\bar\phi(\hat a_t)\right\|_2}, B-\sum_{i=1}^{t-1}o_i, 1)$. We first show that Equation (\ref{eq:learning-constraint}) holds for $o_t=u_{s_t}^* -  v^* + 1 - \frac{\beta_t}{\left\|\bar\phi(\tilde a_t) -\bar\phi(\hat a_t)\right\|_2}$: $1-o_t = ( v^*- u_{s_t}^* )  + \frac{\beta_t}{\left\|\bar\phi(\tilde a_t) -\bar\phi(\hat a_t)\right\|_2}\leq \frac{\beta_t}{\left\|\bar\phi(\tilde a_t) -\bar\phi(\hat a_t)\right\|_2}$, where the last inequality is due to the fact that $ u_{s_t}^* > v^*$. 
Since Equation (\ref{eq:learning-constraint}) holds trivially when $o_t=1$, next we show that Equation (\ref{eq:learning-constraint}) holds for $o_t=B-\sum_{i=1}^{t-1}o_i$. In other words, $1-o_t= 1-B+\sum_{i=1}^{t-1}o_i$. To ensure that $1-o_t \leq \frac{\beta_t}{\left\|\bar\phi(\tilde a_t) -\bar\phi(\hat a_t)\right\|_2}$, we need to impose ${\beta_t} \geq {\left\|\bar\phi(\tilde a_t) -\bar\phi(\hat a_t)\right\|_2} \left(1-B+\sum_{i=1}^{t-1}o_i\right)$. 
 
 Case 2c): $1 > \frac{\beta_t}{\left\|\bar\phi(\tilde a_t) -\bar\phi(\hat a_t)\right\|_2}$ 
 and $u_{s_t}^* -  v^*-y + 1-\frac{\beta_t}{\left\|\bar\phi(\tilde a_t) -\bar\phi(\hat a_t)\right\|_2}\leq 0$. In this case, $e_t = \frac{1}{\left\|\bar\phi(\tilde a_t) -\bar\phi(\hat a_t)\right\|_2}-\frac{\beta_t}{\left\|\bar\phi(\tilde a_t) -\bar\phi(\hat a_t)\right\|_2^2}$ and $o_t = 0$. Intuitively, under this case, $u_{s_t}^* -  v^*$ is less than $y$ and $e_t$ is not high enough to have a positive $o_t$. 
 Since $u_{s_t}^* -  v^* > 0$, then the setup of Case 2c implies that $1>\frac{\beta_t}{\left\|\bar\phi(\tilde a_t) -\bar\phi(\hat a_t)\right\|_2}\geq 1-y+u_{s_t}^* -  v^*$. There are two ways that we can avoid this from happening: by setting $\beta_t >\left\|\bar\phi(\tilde a_t) -\bar\phi(\hat a_t)\right\|_2$ or by setting $\beta_t\leq (1-y)\left\|\bar\phi(\tilde a_t) -\bar\phi(\hat a_t)\right\|_2$.
 Due to the nature of budget constraint, we will set $\beta_t >\left\|\bar\phi(\tilde a_t) -\bar\phi(\hat a_t)\right\|_2$.

Next, we will show that the ratio between the change in the dual and primal objective functions is bounded by $(1+1/(c-1))\pi_{\min}^{-1}$. If $y<1$ and $u_{s_t}^* - v^* - y + \left\|\bar\phi(\tilde a_t) -\bar\phi(\hat a_t)\right\|_2\mathbbm{1}(\hat a_t \neq \tilde a_t)e_t > 0$, then the primal objective function increases by $o_t(u_{s_t}^*- v^*)$, and the increase in the dual objective function is $B\Delta y + z_t + (\beta_t -\left\|\bar\phi(\tilde a_t) -\bar\phi(\hat a_t)\right\|_2\mathbbm{1}(\hat a_t \neq \tilde a_t)e_t \leq o_ty + o_t/(c-1) + o_t - y = o_t/(c-1) +o_t+ (o_t-1)y$, where the first inequality is due to the fact that $e_t$ is positive only when the coefficient $\beta_t -\left\|\bar\phi(\tilde a_t) -\bar\phi(\hat a_t)\right\|_2\mathbbm{1}(\hat a_t \neq \tilde a_t)$ is negative. Thus, the ratio between the change in the dual and primal is 
\begin{align*}
&\frac{\Delta \mathrm{Clairvoyant\; Dual \;with \;Learning}}{\Delta\mathrm{Clairvoyant \;with \;Learning}} = \frac{o_t/(c-1) + o_t + (o_t-1)y}{o_t(u_{s_t}^*- v^*)}\leq \left(1+\frac{1}{c-1}\right)\frac{1}{\pi_{\min}}.    
\end{align*}

Note that the last inequality is because $o_t\leq 1$, and $\frac{1}{u_{s_t}^*- v^*}\leq \frac{1}{\pi_{\min}}$ when $o_t$ is positive.
\end{proof}

\subsection{Proof of Corollary~\ref{cor:competitive_with_learning}}
\label{proof:cor}
\corlearning*
\begin{proof}[Proof of Corollary~\ref{cor:competitive_with_learning}]
    The proof of Proposition~\ref{prop:primal-dual-learning-constraint} shows that the solution provided by Algorithm~\ref{alg:primal-dual-step-constraint} is feasible to \eqref{Eq: clairvoyant-Learning}. Since the objective function in \eqref{Eq: clairvoyant-Learning} and \eqref{pblm: clairvoyant} are the same, and the feasible set in \eqref{Eq: clairvoyant-Learning} is a subset of that of \eqref{pblm: clairvoyant},
Algorithm~\ref{alg:primal-dual-step-constraint} also provides a feasible to \eqref{pblm: clairvoyant}. Thus, the last step in the proof of Proposition~\ref{prop:primal-dual-learning-constraint} holds for Corollary~\ref{cor:competitive_with_learning}, completing the proof.
\end{proof}

\section{Bandit Learning Loss under Known Context Distributions}\label{app:proof}


\paragraph{Notation}
Let $x\in \mathbb{R}^d$ be a $d$-dimensional vector, then $\|x\|_p$ is the $p$-norm of vector $x$. Let $A\in \mathbb{R}^{d\times d}$ be a positive definite matrix, then the weighted $2$-norm of $x$ is $\|x\|_A := \sqrt{x^TAx}$.

\begin{restatable}{prop}{concentration}\label{prop:concentration} 
Let $\bar\theta_t$ be the $L^2$-regularised least-square estimator for $\theta_*$ using $\mathcal{H}_t^{\mathrm{rec}}$, and 
let $\delta\in(0,1)$.
Then, with probability at least $1-\delta$, the following inequality holds for any $t\in [T]$ and
$\lambda>0$: 
\begin{align*}
 &\left\|\theta_* - \bar\theta_t\right\|_{\tilde V_t(\lambda)} \leq \sqrt{2\log\left(\frac{1}{\delta}\right) + \log\left(\frac{\det(\tilde V_t(\lambda))}{\lambda^d}\right)} + \sqrt{\lambda}\norm{\theta_*}_2,   
\end{align*}
where $\bar V_t(\lambda) = \lambda I + \sum_{i=1}^{t-1}\phi(s_i,a_i) \phi(s_i, a_t)^T$.
\end{restatable}
Note that 
the recommender constructs $\hat C_t$ by using $\mathcal{H}_t^{\mathrm{rec}}$ instead in the above 
high-probability confidence bound.
However, because of Constraint~\eqref{eq:learning-constraint}, the bandit learning loss relies \emph{only}  on the concentration of $\tilde C_t$ as stated in
Proposition~\ref{prop:concentration}. 
\subsection{Proof of Proposition~\ref{prop:concentration}}
\label{append:prop_concentration}

\begin{proof}[Proof of Proposition~\ref{prop:concentration}]
Recall that
we assumed $X_{t} (S_t, A_t)$ as the uncertain reward feedback of patient $t$ with context $S_t$ assigned to action $A_t$:
$$ X_{t} (S_t, A_t) = \langle \theta_*, \phi(s_t, a_t)\rangle +
\eta_t. 
$$
If we were to observe the realized reward up to time $t$ (noninclusive) at each step, 
the $L^2$-regularised least-square estimator of $\theta_*$ at time $t$, $\tilde \theta_t$, can be obtained by solving the following optimization problem:
\begin{align*}
L_t(\theta) &= \lambda \|\theta\|^2 + \sum_{i=1}^{t-1}
\left( X_i - \left\langle \theta, \phi(s_i, a_i)
\right\rangle  \right)^{2}, 
\end{align*}
where $\lambda > 0$ is the regularization parameter. We note that since the \emph{revealer} observed the context information at each step, so this update step is well-defined.

Minimizing the above term ($\nabla_\theta {L}_{t}(\theta) = 0$) yields the following estimator for $\theta_*$:
\begin{align*}
\bar{\theta}_t &= 
\left (  
\sum_{i=1}^{t-1} \phi(s_i,a_i)\phi(s_i, a_i)^T 
+ \lambda I
\right )^{-1}  
\left (  
\sum_{i=1}^{t-1} \phi(s_i,a_i)X_i
\right ).
\end{align*}

Let 
$$\tilde V_t(\lambda) = \lambda I + \sum_{i=1}^{t-1}\phi(s_i,a_i) \phi(s_i,a_i)^T,$$ and let $\tilde V_t = V_t(0)$, then we can rewrite $$\bar \theta_t = {\tilde V_t(\lambda)}^{-1}\left ( 
\tilde V_t \theta_* +
\sum_{i=1}^{t-1} 
 \phi(s_i,a_i)\eta_i
\right ).$$

Following the proof of Theorem 20.5 in 
\cite{lattimore2020bandit}, we can show that the following holds for any $t$:
\begin{align*}
\left\|\theta_* - \bar\theta_t\right\|_{\tilde V_t(\lambda)} 
&
=\left\|{\tilde V_t(\lambda)}^{-1} 
\left(
\sum_{i=1}^{t-1} \phi(s_i, a_i)\eta_i
\right) +
\left(\tilde V_t(\lambda)^{-1}\tilde V_t-I\right)\theta_*
\right\|_{\tilde V_t(\lambda)}\\
&\leq \left\|{\tilde V_t(\lambda)}^{-1} 
\left(
\sum_{i=1}^{t-1} \phi(s_i,a_i)\eta_i
\right) 
\right\|_{\tilde V_t(\lambda)} 
+
\left\|
\left(\tilde V_t(\lambda)^{-1}\tilde V_t-I\right)\theta_*
\right\|_{\tilde V_t(\lambda)}\\
&= \left\|\sum_{i=1}^{t-1} 
\phi(s_i,a_i)\eta_i
\right\|_{\tilde V_t(\lambda)^{-1}} 
+ 
\sqrt{\theta_*^\top (\tilde V_t(\lambda)^{-1}\tilde V_t-I)\tilde V_t(\lambda)(\tilde V_t(\lambda)^{-1}\tilde V_t-I)\theta_*
}\\& 
 \leq \left\|\sum_{i=1}^{t-1} 
\phi(s_i,a_i)\eta_i
\right\|_{\tilde V_t(\lambda)^{-1}} + \sqrt{\lambda}\norm{\theta_*}_2 
\end{align*}
where $\tilde V_t$ is the design matrix defined above and note that $\tilde V_t = \tilde V_t(\lambda) - \lambda I$.

Let $d$ be the dimension of $\phi(s_t,a_t)$, and since under $\bar \theta_t$ we would have observed the history of the rewards up to time $t-1$ and the history of states up to time $t$, we consider the $\sigma$-algebra $\mathcal{F}_t :=  \sigma\left(X_1, ..., X_{t-1}, \phi(S_1, A_1) , ..., \phi(S_t,A_{t})\right)$. 


Recall that the noises $\eta_t$'s are conditionally 1-subgaussian: for all $\alpha \in \mathbb{R}$ and $t\geq 1$, $\mathbbm{E}\left[\exp(\alpha \eta_t)|\mathcal{F}_{t-1}\right]\leq \exp \left(\frac{\alpha^2}{2}\right) a.s.$ 
Then, following Lemmas 20.2 and 20.3, and Theorem 20.4 in \cite{lattimore2020bandit}, we have that for all $\lambda >0$ and $\delta \in (0,1)$:
$$\mathbbm{P}\left(\exists t\in \mathbbm{N}: \left\|\sum_{i=1}^{t-1} 
\phi_i(a_i)\eta_i
\right\|_{\tilde V_t(\lambda)^{-1}}^2\geq 2\log(1/\delta) + \log(\det(\tilde V_t(\lambda))/\lambda^d)\right)\leq \delta.$$ 
Together, we have that
with probability $1-\delta$,
$$
\left\|\theta_* - \bar \theta_t\right\|_{\tilde V_t(\lambda)} \leq 
\sqrt{2\log\left(\frac{1}{\delta}\right) + \log\left(\frac{\det(\tilde V_t(\lambda))}{\lambda^d}\right)} + \sqrt{\lambda}\norm{\theta_*}_2.
$$

\end{proof}

\subsection{Proof of Proposition~\ref{thm:bandit_learning}}
\label{append:proof_bandit_regret}
\banditregret*

\begin{proof}[Proof of Proposition~\ref{thm:bandit_learning}] 
At each time $t$, there are 4 quantities that the information revealer and treatment recommender calculate in our algorithm:
\begin{align*}
\text{For } O^\text{ALG}_t=0 :\; 
  (\tilde A_t, \tilde\theta_t) &= \argmax_{(a, \theta)\in\mathcal{A}\times \tilde C_t} \langle \bar\phi(a), \theta\rangle,\\
(\hat A_t, \hat\theta_t) &= \argmax_{(a, \theta)\in\mathcal{A}\times \hat C_t} \langle \bar\phi(a), \theta\rangle,\\ 
\text{For } O^\text{ALG}_t=1:\;
  (\tilde A'_t, \tilde\theta'_t) &= \argmax_{(a, \theta)\in\mathcal{A}\times \tilde C_t} \langle \phi(S_t, a), \theta\rangle,\\
(\hat A'_t, \hat\theta'_t) &= \argmax_{(a, \theta)\in\mathcal{A}\times \hat C_t} \langle \phi(S_t, a), \theta\rangle.
\end{align*}
We first note that when $O^\text{ALG}_t=1$, $\hat C_{t}= \tilde C_t$, so $\tilde A'_t=\hat A'_t$ and $\tilde\theta'_t=\hat\theta'_t$. Thus, when $O^\text{ALG}_t=0$, the treatment recommender takes the action $\hat A_t$, and when $O^\text{ALG}_t=1$, the treatment recommender takes the action $\tilde A'_t$.

Since when $O^\text{ALG}_t=0$, the recommender does not observe the context $S_t$, the best action that it can take is $A^* =\argmax_{a\in \mathcal{A}}\langle\bar\phi(a),\theta_*\rangle$. 
Let $A^*_t = \argmax_{a\in \mathcal{A}}\langle\phi(S_t,a),\theta_*\rangle$.
Thus,
we will compare the performance of our algorithm a ``weaker''  benchmark (who have no access to $S_t$ when $O_t^\text{ALG}=0$) by decomposing the bandit learning loss as follows:
\begin{align*}
\mathrm{BLL}_T &= \mathbb{E}\left[\sum_{t=1}^T\langle\phi(S_t, A_t^*)O_t^\text{AUX} - \phi(S_t, \tilde A_t')O_t^\text{ALG}, \theta_*\rangle
+\sum_{t=1}^T\langle\bar\phi(A^*)(1-O_t^\text{AUX}) - \bar\phi(\hat A_t)(1-O_t^\text{ALG}), \theta_*
\rangle\right]\\
&\leq \underbrace{\mathbb{E}\left[\sum_{t=1}^T\langle\phi(S_t, A_t^*) - \phi(S_t, \tilde A_t'), \theta_*\rangle O_t^\text{ALG}
+\sum_{t=1}^T\langle\bar\phi(A^*) - \bar\phi(\hat A_t), \theta_*
\rangle (1-O_t^\text{ALG})\right]}_{\text{Term I}}\\
&+\underbrace{\mathbb{E}\left[\sum_{t=1}^T 
\langle\phi(S_t, A_t^*)-\bar\phi(A^*) , \theta_*\rangle (O_t^\text{AUX}-O_t^\text{ALG})
\right]}_{\text{Term II}}
\end{align*}
Note that in our above derivation, we replace the auxiliary revealing decision of $O_t^\text{AUX}$ with $O_t^\text{AUX} - O_t^\text{ALG} + O_t^\text{ALG}$. 

We then prove how the first expectation (Term I) in above can be bounded as follows:

\begin{align*}
\text{Term I} =  
&\; \mathbb{E}\left[\sum_{t=1}^T \left\langle
    \phi(S_t, A^*_t) - \phi(S_t, \tilde A'_t) , \theta_* \right\rangle O^\text{ALG}_t+\sum_{t=1}^T \left\langle
    \bar\phi(A^*) - \bar\phi(\hat A_t) , \theta_* \right\rangle (1-O^\text{ALG}_t)\right]\\
        = & \; \mathbb{E}\left[\sum_{t=1}^T \left\langle
    \phi(S_t, A^*_t) - \phi(S_t, \tilde A'_t) , \theta_* \right\rangle O^\text{ALG}_t + \sum_{t=1}^T \left\langle
    \bar\phi(A^*) - \bar\phi(\tilde A_t), \theta_* \right\rangle (1-O^\text{ALG}_t) \right] 
    \\  + &\;
\mathbb{E}\left[\sum_{t=1}^T \left\langle
     \bar\phi(\tilde A_t) - \bar\phi(\hat A_t) , \theta_* \right\rangle (1-O^\text{ALG}_t)
    \mathbbm{1}_{\tilde A_t \neq \hat A_t}
    \right].
    \end{align*}
In above, the first expectation is the standard bandit loss where we observe the entire history up to time $t$. The second term is obtained due to the fact that  when $O^\text{ALG}_t=1$, $\tilde\theta(t) = \hat\theta(t)$, {implying that}  $\tilde A_t = \hat A_t$, and {$\bar\phi(\tilde A_t)= \bar\phi(\hat A_t)$.} 


Consider the following historical information:
\begin{align*}
\mathcal{F}_t :=  \sigma&(X_1, ..., X_{t-1}, A_1, ..., A_{t-1}, O_1, ..., O_{t-1}, S_1, ..., S_{t}),
\end{align*} 
where $A_1, ..., A_{t-1}$ are the random variables indicating the actual treatment taken from time $1$ to $t-1$.
Since our constraint indicates that $(1-o^\text{ALG}_t)\mathbbm{1}_{\tilde a_t \neq \hat a_t}\big\|
\bar\phi(\tilde a_t) - \bar\phi(\hat a_t)\big\|_2\leq \beta_t$ given $\mathcal{F}_{t}$, then using Cauchy–Schwarz inequality, the second term above is bounded by:
\begin{align*}
\mathbb{E} \left[ \left\langle
    \bar\phi(\tilde A_t) - \bar\phi(\hat A_t) , \theta_* \right\rangle (1-O^\text{ALG}_t)\mathbbm{1}_{\tilde A_t \neq \hat A_t}\right]
      & \leq 
    \mathbb{E} \left[\left| \left\langle
    \bar\phi(\tilde A_t) - \bar\phi(\hat A_t) , \theta_* \right\rangle (1-O^\text{ALG}_t)\mathbbm{1}_{\tilde A_t \neq \hat A_t}\right|\right]\\
    & =\mathbb{E}{_{\mathcal{F}_{t}}}\left[\mathbb{E}_{O^\text{ALG}_t}\left[\left| \left\langle
    \bar\phi(\tilde A_t) - \bar\phi(\hat A_t) , \theta_* \right\rangle (1-O^\text{ALG}_t)\mathbbm{1}_{\tilde A_t \neq \hat A_t}\right||\mathcal{F}_{t}\right]\right]
    \\
    & \leq (1-o^\text{ALG}_t)\mathbbm{1}_{\tilde a_t \neq \hat a_t}
    \left\|
\bar\phi(\tilde a_t) - \bar\phi(\hat a_t)\right\|_2 \|\theta_*\|_2
\leq W \beta_t.
\end{align*}

{We note that in the second line above, when conditioned on the $\sigma$-algebra $\mathcal{F}_{t}$, the only uncertainty coming from the inner expectation is from observation $O^\text{ALG}_t$.}

{
By the construction of UCB and the fact that $\theta_*, \tilde\theta'_t, \tilde\theta_t\in \tilde C_t$, we have
that: 
\begin{align*}
\left\langle 
\phi(S_t, A^*_t), \theta_* 
\right\rangle 
& \leq 
\left\langle 
\phi(S_t, \tilde A'_t), \tilde\theta'_t 
\right\rangle, \text{ and}\\
\left\langle 
\bar\phi(A^*), \theta_* 
\right\rangle 
 &\leq 
\left\langle 
\bar\phi(\tilde A_t), \tilde\theta_t 
\right\rangle. 
\end{align*}
}
Thus, by Cauchy-Schwarz and the above facts, we have the following: 
\begin{align*}
\text{Term I} &\leq
\mathbb{E}\left[\sum_{t=1}^T \left\langle
    \phi(S_t, A^*_t) - \phi(S_t, \tilde A'_t) , \theta_* \right\rangle O^\text{ALG}_t + \sum_{t=1}^T \left\langle
    \bar\phi(A^*) - \bar\phi(\tilde A_t),\theta_* \right\rangle (1-O^\text{ALG}_t)\right] +W\sum_{t=1}^T \beta_t
    \\&
    \leq 
\mathbb{E}\left[\sum_{t=1}^T \left\langle
    \phi(S_t, \tilde A'_t) , \tilde\theta'_t -\theta_*  \right\rangle O^\text{ALG}_t + \sum_{t=1}^T \left\langle
    \bar\phi(\tilde A_t), \tilde\theta_t - \theta_* \right\rangle (1-O^\text{ALG}_t)\right] +W\sum_{t=1}^T \beta_t
    \\&
\leq  \mathbb{E}\left[\sum_{t=1}^T \|\phi(S_t, \tilde A'_t)\|_{\tilde V_{t}^{-1}(\lambda)}\|\tilde\theta'_t -\theta_*\|_{\tilde V_{t}(\lambda)}O^\text{ALG}_t 
+ \|\bar\phi( \tilde A_t)\|_{\tilde V_{t}^{-1}(\lambda)}\|\tilde\theta_t -\theta_*\|_{\tilde V_{t}(\lambda)}(1-O^\text{ALG}_t)
\right]
+W\sum_{t=1}^T \beta_t,
\end{align*}
where $\tilde V_t(\lambda) = V_0 + \sum_{i=1}^{t-1}\phi(s_i, a_i)\phi(s_i, a_i)^T$, and $V_0=\lambda I$. 


By Assumption~\ref{assumption:c_max}, we know there exists a constant $c_{\max}$ such that $\|\bar\phi( \widetilde A_t)\|_{\widetilde V_{t}^{-1}(\lambda)} \leq c_{\max} \|\phi(S_t,  \widetilde A_t)\|_{\widetilde V_{t}^{-1}(\lambda)}$.


In addition, we note that since $\widetilde V_t^{-1}(\lambda)$ does not depend on the action taken at  time period $t \in [T]$, we can bound $\|\phi(S_t, \widetilde A_t)\|_{\widetilde V_t^{-1}(\lambda)}$ using Lemma 19.4 of~\cite{lattimore2020bandit}.

In addition, recall that we obtained a high probability confidence set for the unknown parameter $\theta_*$ in Proposition \ref{prop:concentration} as the confidence bound $\left\|\theta_* - \bar \theta_t(\lambda)\right\|_{\widetilde V_t(\lambda)} \leq 
\sqrt{2\log(1/\delta) + \log(\det(\widetilde V_t(\lambda)) / \lambda^d )} + \sqrt{\lambda}\norm{\theta_*}_2$. 
Plugging in the value $\lambda = 1/W^2$, and also utilizing the inequality that $\frac{\det{V_t(\lambda)}}{\lambda^d}\leq 
\left(\mathrm{trace}\left(\frac{V_t(\lambda)}{\lambda d}\right)\right)^d\leq \left(1+\frac{TL^2}{\lambda d}\right)^d$, we can get that the width of the confidence interval is bounded by $\gamma =
1+ \sqrt{2\log\left(\frac{1}{\delta}\right) + d\log(1+\frac{TW^2L^2}{d})}$.
Thus, we have that $\left\|\theta_* - \bar\theta_t(\lambda)\right\|_{\widetilde V_t(\lambda)} \leq \gamma$. Furthermore, since both parameters $\widetilde \theta_t$ and $\theta_*$ belong to this confidence bound, i.e., $\widetilde \theta_t, \theta_* \in \widetilde \Theta_t = \left\{\theta \in \mathbb{R}^d: \|\theta - \bar\theta_t(\lambda) \|_{\widetilde V_t(\lambda)}\leq \gamma \right\},$ we have:
$\left\| \widetilde \theta_t - \theta_*\right\|_{\widetilde V_t(\lambda)} \leq \|\widetilde\theta_t -\bar\theta_t(\lambda)\|_{\widetilde V_{t}(\lambda)}+ \| \theta_* - \bar\theta_t(\lambda)\|_{\widetilde V_{t}(\lambda)} \leq 2\gamma.
$

Considering the above two explanations together, we can now continue bounding Term (I) with a probability of at least $1-2\delta$ as follows: 

\begin{align*}
    \text{Term I} 
     &\leq
   2\gamma \mathbbm{E}\left[  \sqrt{T\sum_{t=1}^T \min\left(1, \|\phi(S_t, \tilde A'_t)\|_{\tilde V_{t}^{-1}}^2\right)(O^\text{ALG}_t)^2} \right]
   \\&+ 2\gamma\mathbbm{E}\left[\sqrt{T\sum_{t=1}^T\min\left(1, \|\bar\phi( \tilde A_t)\|_{\tilde V_{t}^{-1}}^2)\right)(1-O^\text{ALG}_t)^2}\right] + W\sum_{t=1}^T \beta_t\\&
\leq 2\gamma \max(c_{\max}, 1)
\sqrt{2Td\log\left(\frac{\mathrm{trace}(V_0)+ TL^2}{d\det(V_0)^{1/d}}\right)}+ W\sum_{t=1}^T \beta_t
\\&
\leq \max(c_{\max}, 1) \sqrt{8Td\gamma^2\log\left(\frac{d+TW^2L^2}{d}\right)}+ W\sum_{t=1}^T \beta_t,
\end{align*}
where the first inequality follows from the fact that $O^\text{ALG}_t$ is binary and the last inequality follows from Lemma 19.4 of \cite{lattimore2020bandit}.

We next prove how the second expectation (Term II) in our regret decomposition can be bounded. Let $\pi_{\max}:= \max_{s\in S} |u_{s}^* - v^*|$,
where $u_{s_t}^*= \langle\phi(s_t, A_t^*), \theta_*\rangle$, and $v^* = \langle\bar\phi(A^*), \theta_*\rangle$.
Thus, Term II is bounded by: 
\begin{align}
\label{eq:extra-term}
\text{Term II} = \mathbb{E}\left[
\sum_{t=1}^T 
\langle\phi(S_t, A_t^*)-\bar\phi(A^*) , \theta_*\rangle (O_t^\text{AUX}-O_t^\text{ALG})
\right]  
\leq 
\pi_{\max}\mathbb{E}\left[ \sum_{t=1}^T
\left|(O_t^\text{AUX}-O_t^\text{ALG})\right|
\right]\leq 2B \pi_{\max}. 
\end{align}

We note that by Assumption~\ref{assum:budget}, Equation~\ref{eq:extra-term} is bounded by $\mathcal{O}(\sqrt{T})$.

Putting the bounds for Terms I and II, we can bound the bandit learning loss as:
\begin{align*}
\mathrm{BLL}_T 
\leq
\max(c_{\max}, 1) \sqrt{8Td\gamma^2\log\left(\frac{d+TW^2L^2}{d}\right)}+ W\sum_{t=1}^T \beta_t + 2B\pi_{\max}.
\end{align*}
 
%
%
%
\end{proof}

\subsection{Proof of Theorem~\ref{thm:mainresult}}\label{append:proof_main}
\regret*
\begin{proof}[Proof of Theorem~\ref{thm:bandit_learning}] 
Recall we defined three models and then establish the following bridging argument for the regret decomposition:
\begin{align*}
\mathrm{Regret}_T & \leq \mathbb{E}\left[ V^{\text{CLV}} \right] - \mathbb{E}\left[ V^{\text{ALG}} \right] 
= \underbrace{ \mathbb{E}\left[ V^{\text{AUX}} - V^{\text{ALG}} \right] }_\text{Bandit Learning Loss} + \underbrace{ \mathbb{E}\left[ V^{\text{CLV}} - V^{\text{AUX}} \right] }_\text{Information Reveal Loss}.
\end{align*}

From our online primal-dual analysis, we derive the following competitive ratio:
$$
\mathbbm{E}\left[V^{\text{AUX}}\right] / \mathbbm{E}\left[V^{\text{CLV}}\right]
\geq \left(1+\frac{1}{c-1}\right)\frac{1}{\pi_{\min}},
$$
which implies the following by a simple algebra and letting $\alpha = \left(1+\frac{1}{c-1}\right)\frac{1}{\pi_{\min}}$:
$$
\mathbbm{E}[ V^{\text{CLV}}] - \mathbbm{E}\left[V^{\text{AUX}}  \right]\leq
 \left( 1 - \alpha \right) \mathbbm{E}[V^{\text{CLV}}].
$$
Also, from our Proposition \ref{thm:bandit_learning}, we have the following:
\begin{align*}
\mathrm{BLL}_T & = \mathbb{E}\left[ V^{\text{AUX}} - V^{\text{ALG}} \right]\\
& = \mathbb{E}\left[\sum_{t=1}^T\langle\phi(S_t, A_t^*)O_t^\text{AUX} - \phi(S_t, \tilde A_t')O_t^\text{ALG}, \theta_*\rangle
+\sum_{t=1}^T\langle\bar\phi(A^*)(1-O_t^\text{AUX}) - \bar\phi(\hat A_t)(1-O_t^\text{ALG}), \theta_*
\rangle\right]\\
&\leq
\max(c_{\max}, 1)\sqrt{8Td\gamma^2\log\left(\frac{d+TW^2L^2}{d}\right)}+ W\sum_{t=1}^T \beta_t + 2B\pi_{\max}.
\end{align*}

So, summing the upper bounds established above for each term, we have the following:
\begin{align*}
\mathrm{Regret}_T &\leq
\max(c_{\max}, 1)\sqrt{8Td\gamma^2\log\left(\frac{d+TW^2L^2}{d}\right)}+ W\sum_{t=1}^T \beta_t  + 2B\pi_{\max}
+ \left( 1 - \alpha \right)  
\mathbb{E}\left[
V^{\text{CLV}}
\right].
\end{align*}
\end{proof}

\section{Extension to Unknown Context Distribution}\label{subsec:learn_context}
We next  
extend our online learning setting
to the case, where we also learn the unknown context distribution, $\mathbf{p}^*$, in 
Algorithm~\ref{alg:learning-reward-context}.
To derive the regret,
we leverage the \emph{empirical Berstein's inequality} to build a high-probability confidence bound $\tilde P_t$ for the latent context distribution (see Lemma~\ref{lemma:Cont-distr}). Let $\tilde{\mathbf{p}}^t_k$ be the empirical average estimate of $\mathbf{p}^*_k$. 
With $\gamma$ and $\beta_t$ defined above, and $\zeta_t= \sqrt{\frac{2 \; \tilde{\mathbf{p}}_k^t  
\left(  1 -  \tilde{\mathbf{p}}_k^t \right)
\log(\frac{2KT}{\delta}) }{\max\{ m(k, t), 1 \}}}
+ \frac{7\log(\frac{2KT}{\delta})}{3 \left( \max\{ m(k, t)-1, 1 \} \right)}$, where $m(k, t)$ is the number of times that the context $k$ has been observed up to time $t$, the bandit learning loss in Algorithm~\ref{alg:learning-reward-context} is defined as $BLL_T:=\mathbb{E}[\sum_{t=1}^T \langle
    \phi(S_t, A^*_t) O_t^\text{AUX} - \phi(S_t, \tilde A'_t) O_t^\text{ALG}, \theta_* \rangle ]
+\mathbb{E}[\sum_{t=1}^T\sum_{k=1}^K \langle
    \phi(k,A^*) (1-O_t^\text{AUX}) - \phi(k,\hat A_t) (1-O_t^\text{ALG}), \theta_*  \rangle \mathbf{p}^*_k ] $.

\begin{restatable}{prop}{banditcontextregret}
\label{thm:bandit_context_learning}
With probability $1-4\delta$, the bandit learning loss in    Algorithm~\ref{alg:learning-reward-context} is bounded by: \begin{align*}
\mathrm{BLL}_T &=
\leq 
\max(c_{\max}, 1)\sqrt{8Td\gamma^2\log\left(\frac{d+TW^2L^2}{d}\right)} + W\sum_{t=1}^T \beta_t
+ 2B\pi_{\max}
+ (WL+1)\sum_{t=1}^T\sum_{k=1}^K \zeta_t 
 \\& = \mathcal{O}\left(d \sqrt{T} \log\left(T \right) + K \sqrt{T}\right), \end{align*} 
where 
$\tilde A'_t$ and $\hat A_t$ are the respective Algorithm~\ref{alg:learning-reward-context}'s treatments given the history is revealed or not to the recommender.
\end{restatable}
The proof of Proposition~\ref{thm:bandit_context_learning} is included in Appendix
~\ref{append:proof_unknown_context_distribution}.
Algorithm~\ref{alg:learning-reward-context} is computationally expensive since, at each step,
it involves optimizing over 
two convex uncertainty sets when calculating the optimal action. While this step can be solved using an existing bilinear optimization solver, the objective function in our problem is neither convex nor concave, making the problem NP-hard.
Instead,
we plugin the empirical mean estimate of $\mathbf{p}^*$ 
in Algorithm~\ref{alg:learning-reward-context},
and numerically evaluate its performance in
Appendix~\ref{app:figures}.

\begin{algorithm}[h]
\caption{Online $\theta^*$ and $\mathbf{p}^*$ Learning \& Optimization Algorithm }
\label{alg:learning-reward-context}
\begin{algorithmic}
\State \textbf{Input:} $B$
\State \textbf{Initialize:}
$\hat C_1$ and $\tilde C_1$ be the confidence interval of $\theta_*$ of recommender and revealer, respectively; $\tilde P_1,\hat P_1$ be the confidence interval of $\mathbf{p}^*$ of recommender and revealer, respectively.
\For{$t=1,...,T$}
\State On each iteration $t$,  \emph{revealer} observes the new context $s_t$ and calculates:
\State $(\tilde A_t, \tilde\theta_t, \tilde{ \mathbf{p}}^t) = \argmax_{(a, \theta, \mathbf{p})\in\mathcal{A}\times \tilde C_t\times \tilde P_t} \langle \sum_{k=1}^K\phi(k,a)\mathbf{p}_k, \theta\rangle$,
\State $(\hat A_t, \hat\theta_t, \hat{\mathbf{p}}^t) = \argmax_{(a, \theta,\mathbf{p})\in\mathcal{A}\times \hat C_t\times \hat{{P}}_t} \langle \sum_{k=1}^K\phi(k,a)\mathbf{p}_k, \theta\rangle$,
\State $\Tilde{u}_{s_t}^t  = \max_{(a, \theta)\in\mathcal{A}\times \tilde C_t\times } \langle \phi(s_t, a), \theta\rangle$,
\State   
$\Tilde{v}^t = \left\langle
\sum_{k=1}^K\phi(k,\tilde A_t)\tilde{\mathbf{p}}_k^t,\Tilde{\theta}_t
\right\rangle$.
\State Given $\Tilde{u}_{s_t}^t, \tilde v_t,$ $\tilde A_t$, $\hat A_t$, and $B$,  revealer uses Algorithm~\ref{alg:primal-dual-step-constraint} to reveal the history $\mathcal{H}_{t}^{\mathrm{rev}}$ to recommender with probability: $O_t = \mathrm{Bernoulli}(o_t)$.
\If{$O_t=1$}
\State  \emph{Recommender} set $\hat C_t  = \tilde C_t$ and $\hat P_t = \tilde P_t$ and calculates:
\State $(\hat A^*_t, \theta_t)  = \argmax_{(a, \theta)\in\mathcal{A}\times \tilde C_t} \langle \phi(s_t, a), \theta\rangle$.
\State  \emph{Recommender} takes action $\hat A_t^*$.
\Else 
\State \emph{Recommender} set $\hat C_{t+1} = \hat C_t$, $\hat P_{t+1} = \hat P_t$, and takes action $\hat A_t^*=\hat A_t$.
\EndIf
\State \emph{Revealer} observes reward $X_t$ and $S_t$, and update $\tilde C_{t+1}$ and $\tilde P_{t+1}$.
\EndFor
\end{algorithmic}
\end{algorithm}

\subsection{Proof}\label{append:proof_unknown_context_distribution}
\begin{restatable}{lem}{berstein}\label{lemma:berstein}
(Empirical Berstein's Inequality \cite{maurer2009empirical}) Let $X = (X_1, X_2, ..., X_n)$ be i.i.d. random vector with values in $[0, 1]^n$, and let $\delta \in (0, 1)$. Then, we have the following bound holds with probability at least $1-\delta$:
\begin{align*}
\mathbb{E}\left[ X \right] - \frac{1}{n} \sum_{i=1}^n X_i \; \leq \; \sqrt{\frac{2 V_n(X) \log(\frac{2}{\delta}) }{n}}
+ \frac{7\log(\frac{2}{\delta})}{3(n-1)},
\end{align*}
where $V_n(X)$ is the sample variance. 
\end{restatable}

\begin{restatable}{lem}{confidenceberstein}\label{lemma:Cont-distr}
(Confidence Bound on Context Distribution) If we use the empirical average estimate $\widehat{\mathbf{p}}_k(t)$  for 
 estimating the latent context distribution $\mathbf{p}^*_k = \mathbb{P}(S_t = k)$) for each context $k \in [K]$ at iteration $t \in [T]$, then the following bound holds with probability at least $1-2\delta$:
\begin{align}
\left|
\mathbf{p}^*_k - \widehat{\mathbf{p}}_k(t) 
\right| \; \leq 
\sqrt{\frac{2 \; \widehat{\mathbf{p}}_k(t)  
\left(  1 -  \widehat{\mathbf{p}}_k(t) \right)
\log(\frac{2KT}{\delta}) }{\max\{ m(k, t), 1 \}}}
+ \frac{7\log(\frac{2KT}{\delta})}{3 \left( \max\{ m(k, t)-1, 1 \} \right)},
\end{align}
where $m(k, t)$ is the number of time that the context $k$ has been observed up to time $t$. 
\end{restatable}

\begin{proof}[Proof of Lemma~\ref{lemma:Cont-distr}] 
First, we establish the following for estimating the latent context distribution (i.e., $\mathbf{p}_k = \mathbb{P}(S_t = k)$) for each context $k$:
\begin{align*}
\widehat{\mathbf{p}}_k(t) = \frac{\sum_{u = 1}^{t-1} \mathbbm{1}(S_u = k)}{\sum_{u = 1}^{t-1}\sum_{m = 1}^{K} \mathbbm{1}(S_u = m)}, 
\end{align*}
Using the empirical Berstein's inequality in Lemma~\ref{lemma:berstein} and making a union-bound argument, the resulting bound is obtained with probability at least $1-2\delta$.
\end{proof}



\banditcontextregret*
\begin{proof}[Proof of Proposition
~\ref{thm:bandit_context_learning}
]
In this proof, we take the learning of context distribution into account. Let $\tilde P_t$ be the uncertainty set for the context distribution that contains the ground truth context distribution $\mathbf{p}^*$ at time $t$.

At each time $t$, there are 4 quantities that the information revealer and  treatment recommender calculate:
\begin{align}
\text{For } O^\text{ALG}_t=0 :\; 
  (\tilde A_t, \tilde\theta_t, \tilde{\mathbf{p}}^t) &= \argmax_{(a, \theta, \mathbf{p})\in\mathcal{A}\times \tilde C_t\times \tilde P_t} \left\langle \sum_{k=1}^K \phi(k,a)\mathbf{p}, \theta\right\rangle,\label{eq:tilde_A_learn_p}\\
(\hat A_t, \hat\theta_t, \hat{\mathbf{p}}^t) &= \argmax_{(a, \theta, \mathbf{p})\in\mathcal{A}\times \hat C_t\times \hat P_t} \left\langle \sum_{k=1}^K \phi(k,a)\mathbf{p}, \theta\right\rangle,\label{eq:hat_a_learn_p}\\ 
\text{For } O^\text{ALG}_t=1:\;
  (\tilde A'_t, \tilde\theta'_t) &= \argmax_{(a, \theta)\in\mathcal{A}\times \tilde C_t} \langle \phi(S_t, a), \theta\rangle,\\
(\hat A'_t, \hat\theta'_t) &= \argmax_{(a, \theta)\in\mathcal{A}\times \hat C_t} \langle \phi(S_t, a), \theta\rangle.
\end{align}
We first note that when $O^\text{ALG}_t=1$, $\hat C_{t}= \tilde C_t$, so $\tilde A'_t=\hat A'_t$ and $\tilde\theta'_t=\hat\theta'_t$. Thus, when $O^\text{ALG}_t=0$, the treatment recommender takes the action $\hat A_t$, and when $O^\text{ALG}_t=1$, the treatment recommender takes the action $\tilde A'_t$.

Since when $O^\text{ALG}_t=0$, the recommender does not observe the context $S_t$, the best action that it can take is $A^* =\argmax_{a\in \mathcal{A}}\langle\bar\phi(a),\theta_*\rangle$. 
Let $A^*_t = \argmax_{a\in \mathcal{A}}\langle\phi(S_t,a),\theta_*\rangle$.
Thus,
we will compare the performance of our algorithm to a ``weaker''  benchmark (who has no access to $S_t$ when $O^\text{ALG}_t=0$). The bandit regret at time $T$ is the following:


\begin{align*}
\mathrm{BLL}_T &=
\mathbb{E}\left[\sum_{t=1}^T \left\langle
    \phi(S_t, A^*_t) O_t^\text{AUX} - \phi(S_t, \tilde A'_t) O_t^\text{ALG}, \theta_* \right\rangle \right] \\&+\mathbb{E}\left[\sum_{t=1}^T\sum_{k=1}^K \left\langle
    \phi(k,A^*) (1-O_t^\text{AUX}) - \phi(k,\hat A_t) (1-O_t^\text{ALG}), \theta_*  \right\rangle \mathbf{p}^*_k \right]
    \\&= \mathbb{E}\left[\sum_{t=1}^T\langle\phi(S_t, A_t^*) - \phi(S_t, \tilde A_t'), \theta_*\rangle O_t^\text{ALG}
+\sum_{t=1}^T\sum_{k=1}^K\langle\phi(k,A^*) - \phi(k,\hat A_t), \theta_*
\rangle \mathbf{p}^*_k (1-O_t^\text{ALG})   \right]\\
&+\mathbb{E}\left[\sum_{t=1}^T\langle\phi(S_t, A_t^*) , \theta_*\rangle (O_t^\text{AUX}-O_t^\text{ALG})
+\sum_{t=1}^T\sum_{k=1}^K\langle\phi(k, A^*), \theta_*
\rangle \mathbf{p}^*_k \left(O_t^\text{ALG} -O_t^\text{AUX}\right)\right]
\\&=  \mathbb{E}\left[\sum_{t=1}^T\langle\phi(S_t, A_t^*) - \phi(S_t, \tilde A_t'), \theta_*\rangle O_t^\text{ALG} \right] +\mathbb{E}\left[\sum_{t=1}^T\sum_{k=1}^K\langle\phi(k,A^*) - \phi(k,\tilde A_t), \theta_* \rangle \mathbf{p}^*_k (1-O_t^\text{ALG})  \right]\\
&+\mathbb{E}\left[\sum_{t=1}^T\sum_{k=1}^K\langle \phi(k,\tilde A_t) - \phi(k,\hat A_t), \theta_*
\rangle \mathbf{p}^*_k  (1-O_t^\text{ALG}) \mathbbm{1}_{\tilde A_t \neq \hat A_t} \right] 
\\&
+\mathbb{E}\left[\sum_{t=1}^T\langle\phi(S_t, A_t^*) - \bar\phi(A^*), \theta_*\rangle (O_t^\text{AUX}-O_t^\text{ALG}) \right].
    \end{align*}

To obtain the above decomposition, we first replace the auxiliary revealing decision of $O_t^\text{AUX}$ with $O_t^\text{AUX} - O_t^\text{ALG} + O_t^\text{ALG}$. We then add and subtract the term $\phi(k,\tilde A_t)$, and use the fact that $\bar\phi(A^*) = \sum_{k=1}^K\phi(k, A^*)\mathbf{p}^*_k$.


The last expectation term above can be bounded using the similar procedure for bounding term II in Proposition~\ref{thm:bandit_learning}, which results in:
\begin{align*}
\mathbb{E}\left[\sum_{t=1}^T\langle\phi(S_t, A_t^*) - \bar\phi(A^*), \theta_*\rangle (O_t^\text{AUX}-O_t^\text{ALG}) \right] \leq 2B\pi_{\max}.
\end{align*}

We then take a look at bounding the third expectation term.
Consider the following  information history:
$$
\mathcal{F}_t :=  \sigma(\{X_i\}_{i\in[t-1]}, \{O^\text{ALG}_i\}_{i\in[t-1]}, 
\{\phi(k, A_i)\}_{k \in K}^{i\in[t-1]}, \{S_i\}_{i\in[t]}, \tilde{\mathbf{p}}^1, ..., \tilde{\mathbf{p}}^t , \hat{\mathbf{p}}^1,..., \hat{\mathbf{p}}^t),
$$
where the notation $\{X_i\}_{i\in[j]}$ represents the enumerations of $X_i$ from $i=1$ to $j$, and similarly for the other variables.


Our constraint indicates that $(1-o_t^{\text{ALG}})\mathbbm{1}_{\tilde a_t \neq \hat a_t}\left\|
\sum_{k=1}^K\phi(k,\tilde a_t)\tilde{\mathbf{p}}_k^t - \sum_{k=1}^K\phi(k,\hat a_t)\tilde{\mathbf{p}}_k^t\right\|_2\leq \beta_t$ given the information history 
 $\mathcal{F}_{t}$. 
%
Using this inequality along with the Cauchy–Schwarz inequality, the third expectation term above can be bounded as follows:
\begin{align*}
&\quad\mathbb{E} \left[ \sum_{t=1}^T\sum_{k=1}^K\left\langle
    \phi(k,\tilde A_t) - \phi(k,\hat A_t) , \theta_* \right\rangle \mathbf{p}^*_k(1-O_t^\text{ALG})\mathbbm{1}_{\tilde A_t \neq \hat A_t}\right]  
    \\& 
    \quad 
    \leq \sum_{t=1}^T
    \mathbb{E} \left[\left|\sum_{k=1}^K \left\langle
    \phi(k,\tilde A_t) - \phi(k,\hat A_t), \theta_* \right\rangle \left(\mathbf{p}^*_k -\tilde{\mathbf{p}}_k^t+\tilde{\mathbf{p}}_k^t\right)(1-O_t^\text{ALG})\mathbbm{1}_{\tilde A_t \neq \hat A_t}\right|\right]
    \\& 
    \quad
    = \sum_{t=1}^T \mathbb{E}{_{\mathcal{F}_{t}}}\left[\mathbb{E}_{O_t^\text{ALG}}\left[\left| \sum_{k=1}^K\left\langle
    \phi(k,\tilde A_t) - \phi(k,\hat A_t) , \theta_*  \right\rangle \left(\mathbf{p}^*_k -\tilde{\mathbf{p}}_k^t+\tilde{\mathbf{p}}_k^t\right) (1-O_t^\text{ALG})\mathbbm{1}_{\tilde A_t \neq \hat A_t}\right||\mathcal{F}_{t}\right]\right]
    \\&
    \quad 
    \leq \sum_{t=1}^T (1-o_t^\text{ALG})\mathbbm{1}_{\tilde a_t \neq \hat a_t}
   \left\|
\sum_{k=1}^K\left(\phi(k,\tilde a_t)- \phi(k,\hat a_t)\right)\tilde{\mathbf{p}}_k^t \right\|_2  
\|\theta_*\|_2 
\\ &\quad
+ \sum_{t=1}^T (1-o_t^\text{ALG})\mathbbm{1}_{\tilde a_t \neq \hat a_t}\left\|\sum_{k=1}^K\left(\phi(k,\tilde a_t)- \phi(k,\hat a_t)\right)(\mathbf{p}^*_k-\tilde{\mathbf{p}}_k^t)\right\|_2 \|\theta_*\|_2 
\\&\quad
\leq W \sum_{t=1}^T \beta_t +  WL \sum_{t=1}^T \left|\sum_{k=1}^K{ \left( \mathbf{p}^*_k -\tilde{\mathbf{p}}_k^t \right) }\right|.
\end{align*}
The last inequality is due to the fact that $\|\theta_*\|_2 \leq W$ and $\max_{a\in\mathcal{A}, s\in \mathcal{S}} \|\phi(s,a)\|_2\leq L$.
{We note that in the second line above, when conditioned on the $\sigma$-algebra $\mathcal{F}_{t}$, the only uncertainty coming from the inner expectation is from observation $O_t^{\text{ALG}}$.}

{
By the construction of UCB (optimism) and the fact that $\theta_*, \tilde\theta'_t, \tilde\theta_t\in \tilde C_t$, we have the inequalities: 
\begin{align*}
\left\langle 
\phi(S_t, A^*_t), \theta_* 
\right\rangle 
& \leq 
\left\langle 
\phi(S_t, \tilde A'_t), \tilde\theta'_t 
\right\rangle, \text{ and}\\
\left\langle 
\sum_{k=1}^K\phi(k,A^*)\mathbf{p}^*_k, \theta_* 
\right\rangle 
 &\leq 
\left\langle 
\sum_{k=1}^K\phi(k, \tilde A_t)\tilde{\mathbf{p}}^t_{k}, \tilde\theta_t 
\right\rangle. 
\end{align*}
}
By Cauchy-Schwarz and the above facts, we can establish the following arguments to bound the bandit regret: 
\begin{align*}
\mathrm{BLL}_T &\leq
\mathbb{E}\left[\sum_{t=1}^T \left\langle
    \phi(S_t, A^*_t) - \phi(S_t, \tilde A'_t) , \theta_* \right\rangle O_t^{\text{ALG}}\right] 
    +W\sum_{t=1}^T \beta_t + WL \sum_{t=1}^T\left|\sum_{k=1}^K\mathbf{p}^*_k - \tilde{\mathbf{p}}^t_k\right|\\
    &+ \mathbb{E}\left[\sum_{t=1}^T \left\langle
    \sum_{k=1}^K\phi(k, A^*)\mathbf{p}^*_k - \sum_{k=1}^K\phi(k, \tilde A_t)\mathbf{p}^*_k,\theta_* \right\rangle  (1-O_t^{\text{ALG}})\right] +2B\pi_{\max}
    \\&
    \leq 
\mathbb{E}\left[\sum_{t=1}^T \left\langle
    \phi(S_t, \tilde A'_t) , \tilde\theta'_t -\theta_*  \right\rangle O_t^{\text{ALG}}\right]
    +W\sum_{t=1}^T \beta_t
    +WL \sum_{t=1}^T\left|\sum_{k=1}^K\mathbf{p}^*_k - \tilde{\mathbf{p}}^t_k\right|
    \\
    & +\mathbb{E}\left[\sum_{t=1}^T \sum_{k=1}^K\left\langle
    \phi(k,\tilde A_t), \left(\tilde{\mathbf{p}}^t_k-\mathbf{p}^*_k+\mathbf{p}^*_k\right)\tilde\theta_t - \mathbf{p}^*_k\theta_* \right\rangle (1-O_t^{\text{ALG}})\right] +2B\pi_{\max}
    \\&\leq  \mathbb{E}\left[\sum_{t=1}^T \|\phi(S_t, \tilde A'_t)\|_{\tilde V_{t}^{-1}}\|\tilde\theta'_t -\theta_*\|_{\tilde V_{t}}O_t^{\text{ALG}} \right]
+W\sum_{t=1}^T \beta_t
    +WL \sum_{t=1}^T\left|\sum_{k=1}^K\mathbf{p}^*_k - \tilde{\mathbf{p}}^t_k\right|
\\& +\mathbb{E}\left[\sum_{t=1}^T
\left(\left\|\sum_{k=1}^K \phi(k, \tilde A_t)\mathbf{p}^*_k\right\|_{\tilde V_{t}(\lambda)^{-1}}\|\tilde\theta_t -\theta_*\|_{\tilde V_{t}(\lambda)} 
+\sum_{k=1}^K\left\langle
    \phi(k,\tilde A_t), \left(\tilde{\mathbf{p}}^t_k-\mathbf{p}^*_k\right)\tilde\theta_t  \right\rangle
\right)(1-O_t^{\text{ALG}})
\right] 
+ 2B\pi_{\max}.
\end{align*}
{Our assumption indicates that $\max_{a\in \mathcal{A}, s\in S} \langle\phi(s, a), \theta_* \rangle\leq 1$ with Q-probability one. Then, it is also reasonable to assume that for all $t$, we have $\max_{a\in \mathcal{A}, s\in S, \theta\in \tilde C_t} \langle\phi(s, a), \theta \rangle\leq 1$ with Q-probability one. Thus, the bandit regret above satisfies the following:
\begin{align*}
\mathrm{BLL}_T &\leq
\mathbb{E}\left[\sum_{t=1}^T \|\phi(S_t, \tilde A'_t)\|_{\tilde V_{t}^{-1}}\|\tilde\theta'_t -\theta_*\|_{\tilde V_{t}}O_t^{\text{ALG}} \right]
+W\sum_{t=1}^T \beta_t     
+WL \sum_{t=1}^T\left|\sum_{k=1}^K\mathbf{p}^*_k - \tilde{\mathbf{p}}^t_k\right|
\\&+\mathbb{E}\left[\sum_{t=1}^T
\left(\left\|\sum_{k=1}^K\phi(k, \tilde A_t)\mathbf{p}^*_k\right\|_{\tilde V_{t}(\lambda)^{-1}}\|\tilde\theta_t -\theta_*\|_{\tilde V_{t}(\lambda)}
+\sum_{k=1}^K\left(\tilde{\mathbf{p}}^t_k-\mathbf{p}^*_k\right)
\right)(1-O_t^{\text{ALG}})
\right]+2B\pi_{\max}.
\end{align*}
}

Furthermore, from the result of Proposition~\ref{prop:concentration}, if we plug $\lambda = 1/W^2$ in, and also employ the inequality that $\frac{\det{V_t(\lambda)}}{\lambda^d}\leq 
\left(\mathrm{trace}\left(\frac{V_t(\lambda)}{\lambda d}\right)\right)^d\leq \left(1+\frac{TL^2}{\lambda d}\right)^d$, we can get that the width of the confidence interval is bounded by $\gamma =
1+ \sqrt{2\log\left(\frac{1}{\delta}\right) + d\log(1+\frac{TW^2L^2}{d})}$. We assume that the reward $X_t$ that we get at each round is bounded by 1. Thus, we can establish the following to bound the bandit regret:
\begin{align*}
    \mathrm{BLL}_T
    &\leq
    2\gamma\mathbbm{E}\left[  \sqrt{T\sum_{t=1}^T \min\left(1, \|\phi(S_t, \tilde A'_t)\|_{\tilde V_{t}^{-1}}^2\right){(O_t^{\text{ALG}})}^2} \right] 
    +(WL+1) \sum_{t=1}^T\left|\sum_{k=1}^K\mathbf{p}^*_k - \tilde{\mathbf{p}}^t_k\right|
    \\&
    + 2\gamma\mathbbm{E}\left[\sqrt{T\sum_{t=1}^T\min\left(1, \left\|\sum_{k=1}^K\phi(k, \tilde A_t)\mathbf{p}^*_k\right\|_{\tilde V_{t}^{-1}(\lambda)}^2\right)(1-O_t^{\text{ALG}})^2}\right]   + W\sum_{t=1}^T \beta_t +2B\pi_{\max}
    \\&
\leq 2\gamma \max(c_{\max},1)\sqrt{2Td\log\left(\frac{\mathrm{trace}(V_0)+ TL^2}{d\det(V_0)^{1/d}}\right)} + W\sum_{t=1}^T \beta_t
    +(WL+1) \sum_{t=1}^T\left|\sum_{k=1}^K\mathbf{p}^*_k - \tilde{\mathbf{p}}^t_k\right| +2B\pi_{\max}
\\&
\leq \max(c_{\max},1)\sqrt{8Td\gamma^2\log\left(\frac{d+TW^2L^2}{d}\right)} + W\sum_{t=1}^T \beta_t
+(WL+1) \sum_{t=1}^T\sum_{k=1}^K\left|\mathbf{p}^*_k - \tilde{\mathbf{p}}^t_k\right|+2B\pi_{\max},
\end{align*}
where the first inequality is due to the fact that $O_t^{\text{ALG}}$ is binary, and the last inequality follows from Lemma 19.4 of \cite{lattimore2020bandit}. Finally, using the high-probability bound we established in Lemma~\ref{lemma:Cont-distr} for the latent context distribution, we complete the proof by establishing the following bound:
\begin{align*}
\mathrm{BLL}_T &\leq\max(c_{\max},1)\sqrt{8Td\gamma^2\log\left(\frac{d+TW^2L^2}{d}\right)} + W\sum_{t=1}^T \beta_t
+(WL+1) \sum_{t=1}^T\sum_{k=1}^K\zeta_t  +2B\pi_{\max},
\end{align*}
where $\zeta_t = \sqrt{\frac{2 \; \tilde{\mathbf{p}}_k^t  
\left(  1 -  \tilde{\mathbf{p}}_k^t \right)
\log(\frac{2KT}{\delta}) }{\max\{ m(k, t), 1 \}}}
+ \frac{7\log(\frac{2KT}{\delta})}{3 \left( \max\{ m(k, t)-1, 1 \} \right)}.$
\end{proof}

\section{Additional Synthetic Experimental Results}
\label{app:figures}
All experiments were run on a high performance computing cluster with a 12-core CPU.
\begin{table*}[h]
\centering
\begin{tabular}{ccccccc}
\toprule
$B$& $2$ & 
$4$ 
& $8$ & $16$ & $32$ & $64$ \\ \midrule
PD1-UCB  & $0.805$ & $0.832$ & $0.849$ & $0.861$ & $0.870$ & $0.879$ \\ 
PD2-UCB & $0.759$   & $0.807$ & $0.836$ & $0.854$ & $0.867$ & $0.878$\\ 
 \bottomrule
\end{tabular}
\caption{Empirical competitive ratio for one value of 
$\theta_*$ averaged over $200$ context arrival sequences. }
\label{table:competitve ratio}
\end{table*}

\begin{figure}
    \centering
\includegraphics[width=0.46\linewidth, keepaspectratio]{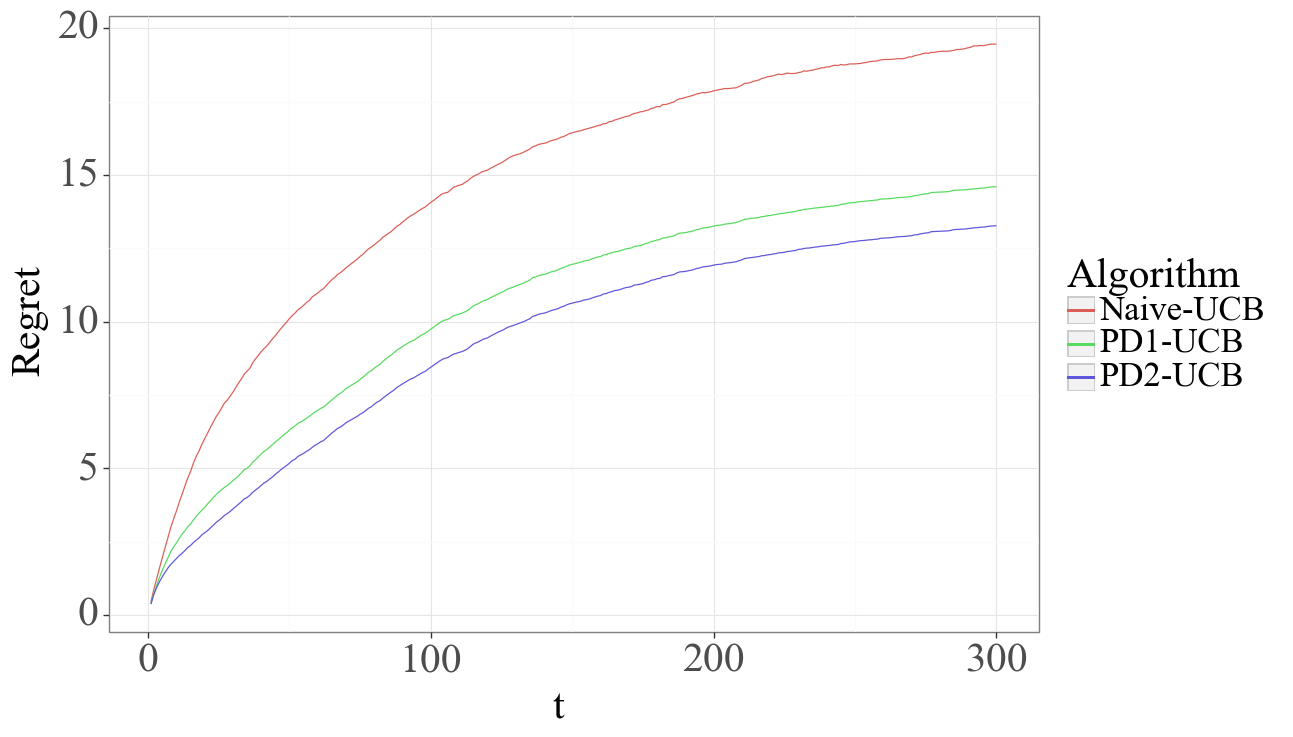}
\includegraphics[width=0.26\linewidth, keepaspectratio]{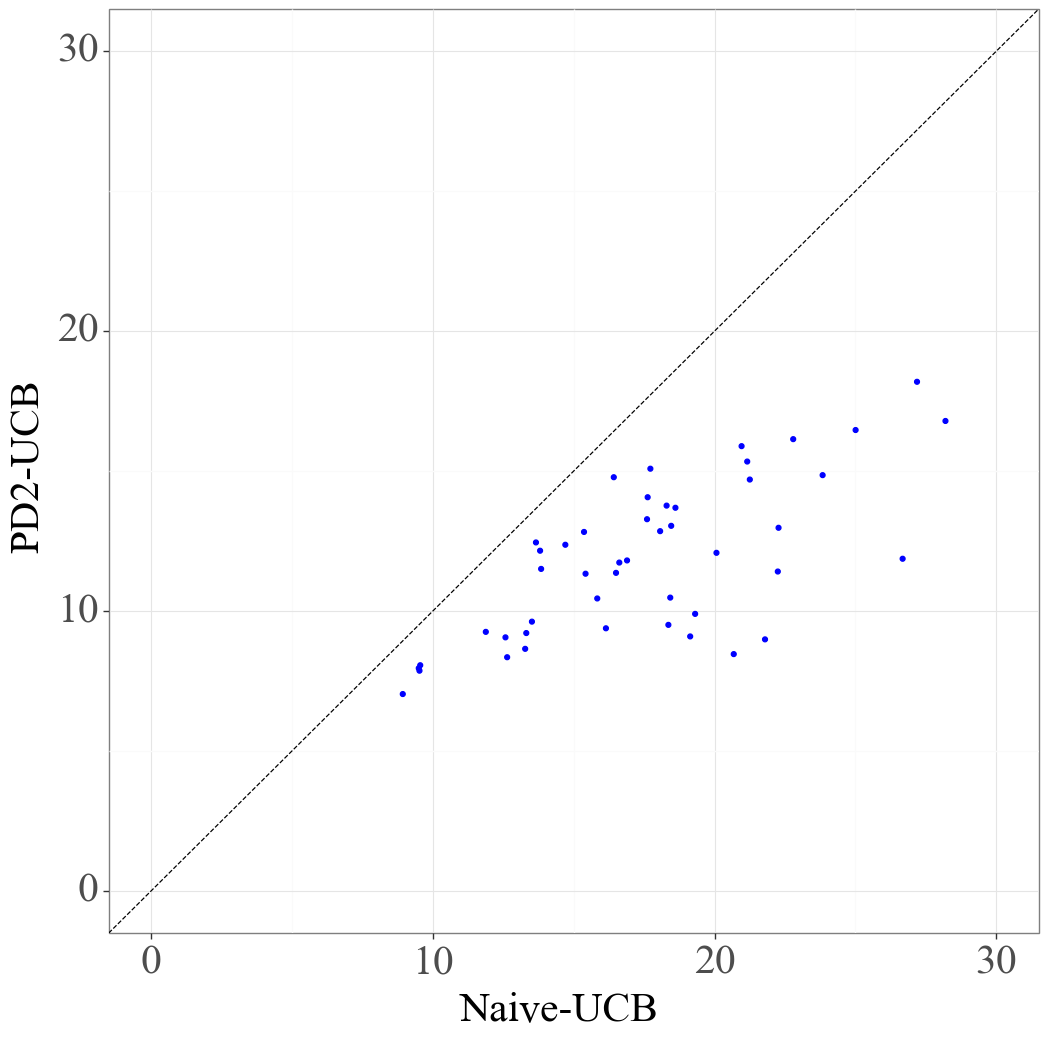}
\includegraphics[width=0.26\linewidth, keepaspectratio]{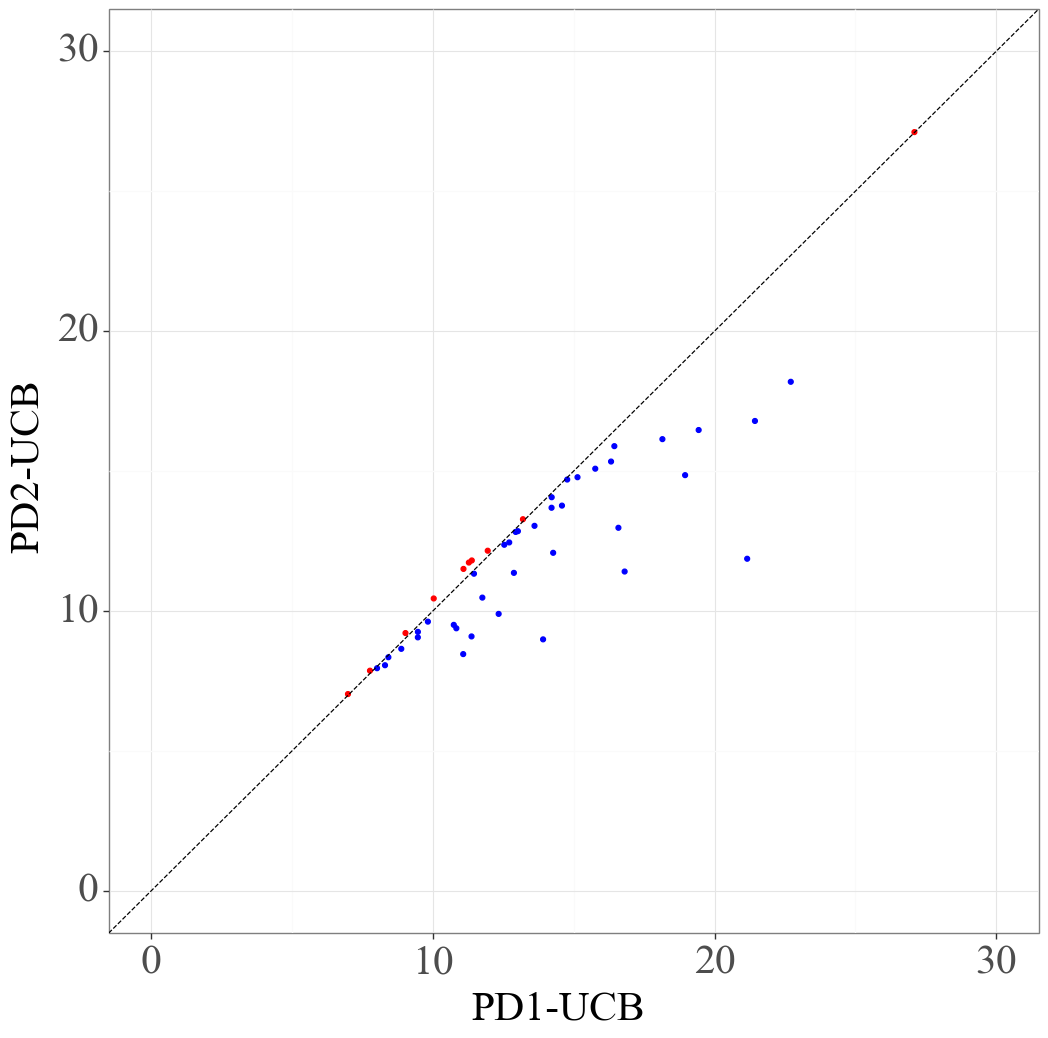}
    \caption{Average regret (left) and scatter plot for $B=20$ at $T=300$ under known $\mathbf{p}^*$. Each dot corresponds to one instance averaged over 50 replications.}
    \label{fig:scatter_20}
\end{figure}

\begin{figure}
    \centering
\includegraphics[width=0.46\linewidth, keepaspectratio]{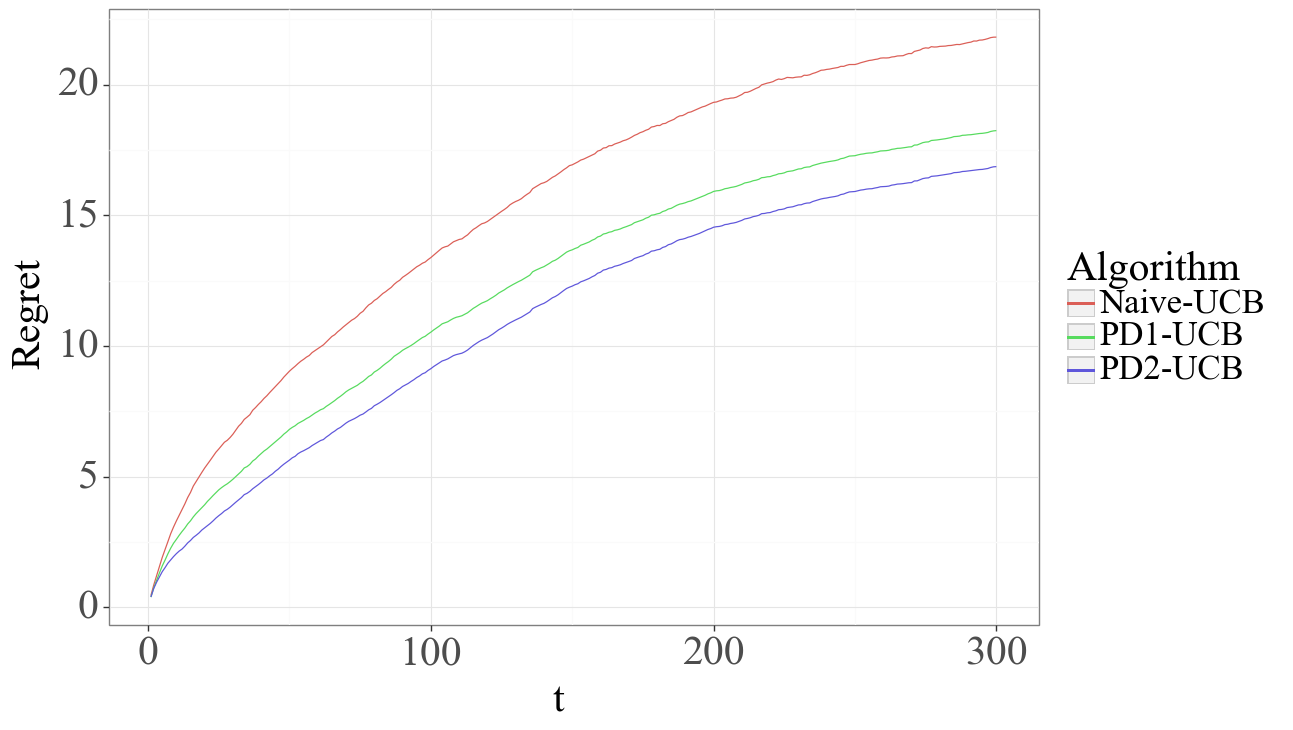}
\includegraphics[width=0.26\linewidth, keepaspectratio]{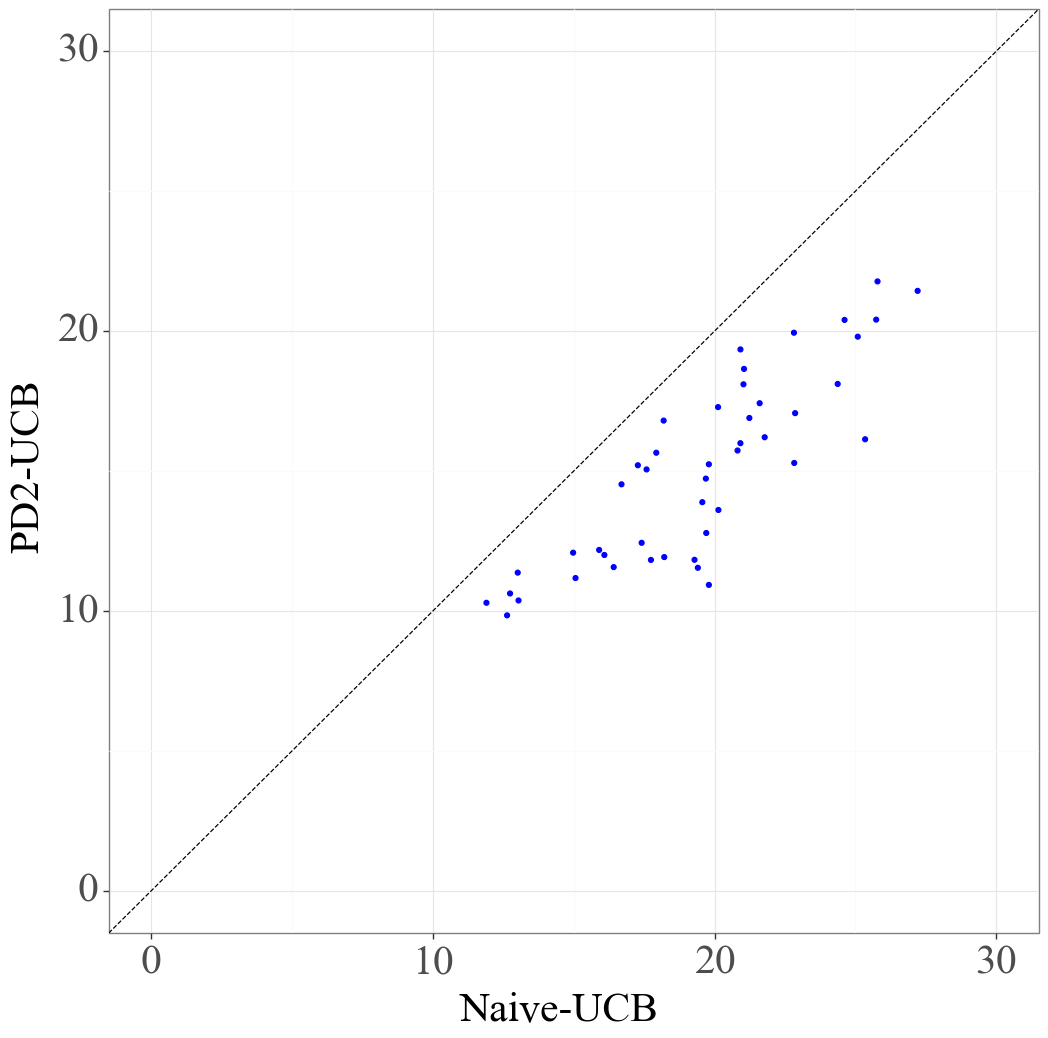}
\includegraphics[width=0.26\linewidth, keepaspectratio]{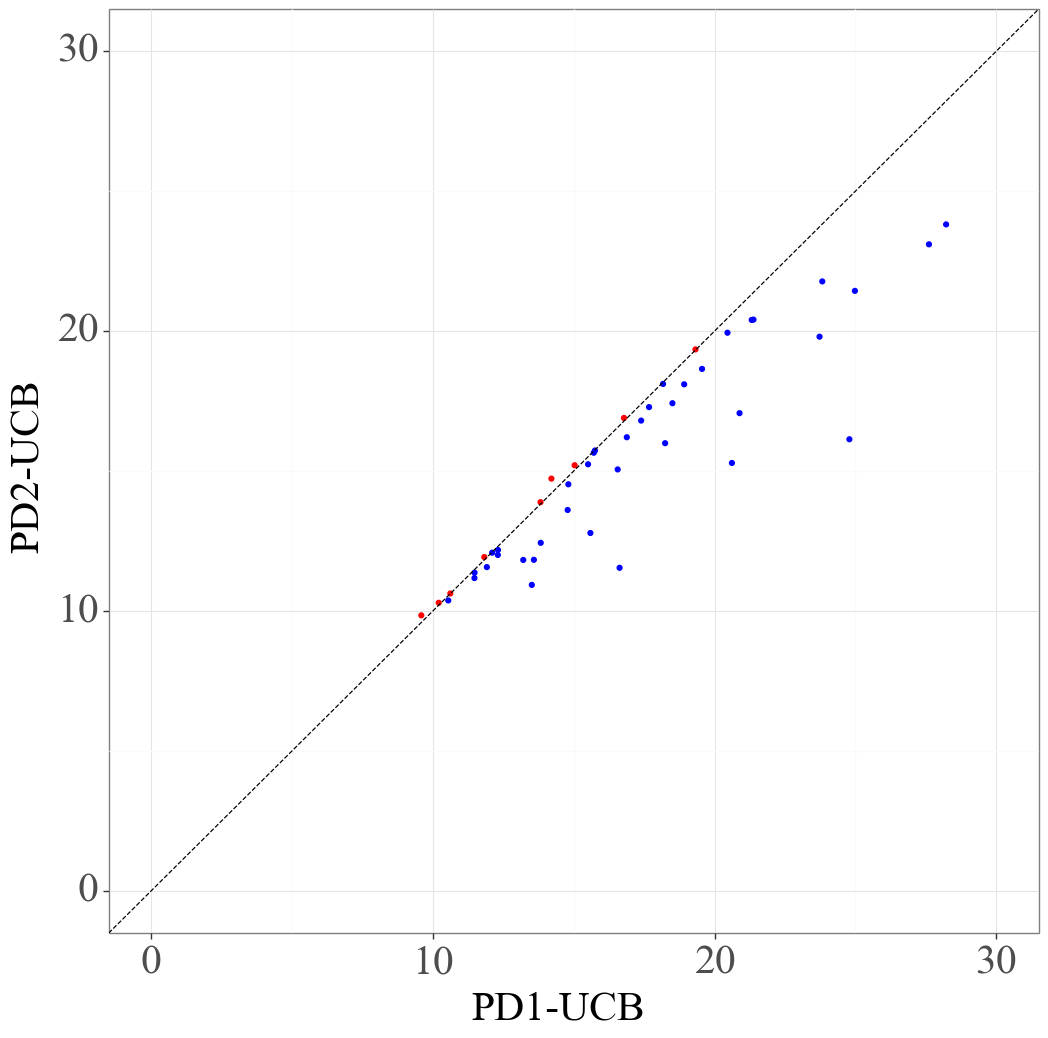}
    \caption{Average regret (left) and scatter plots for $B=30$ at $T=300$ under known $\mathbf{p}^*$. Each dot corresponds to one instance averaged over 50 replications.}
    \label{fig:scatter_30}
\end{figure}

\begin{figure}
    \centering
\includegraphics[width=0.46\linewidth, keepaspectratio]{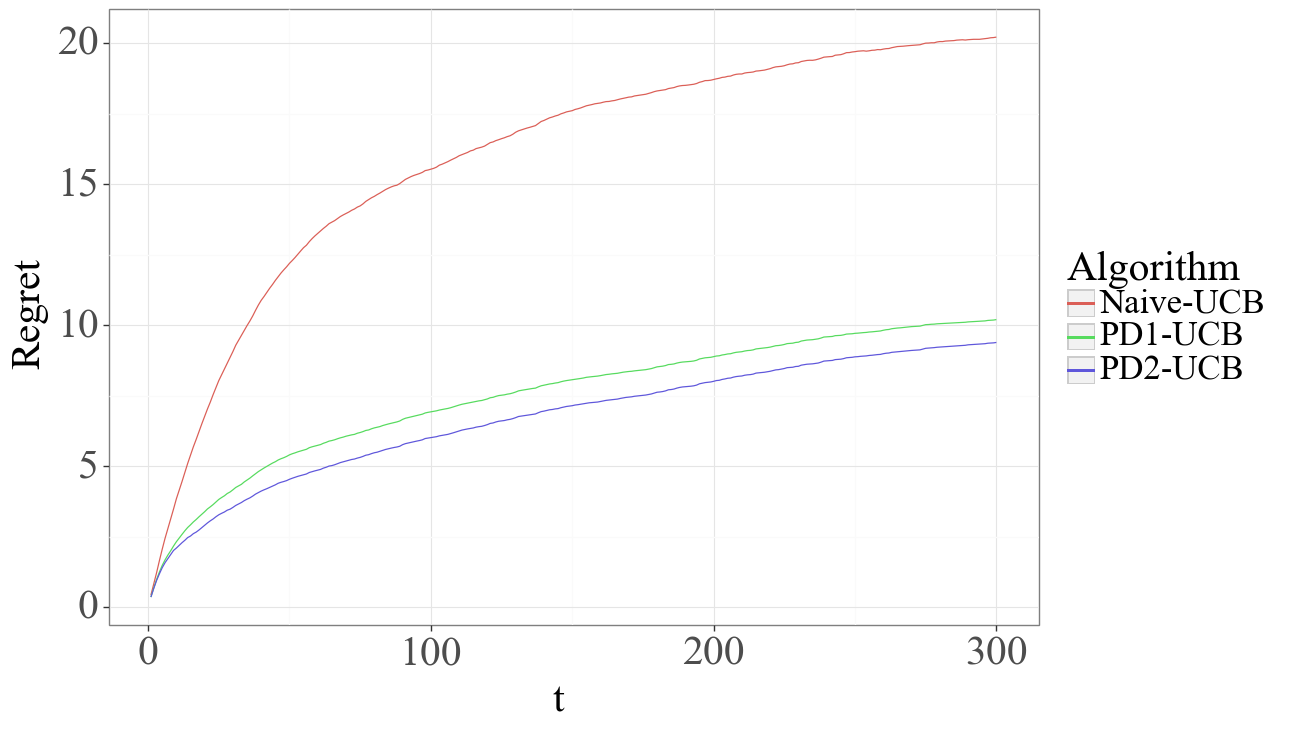}
\includegraphics[width=0.26\linewidth, keepaspectratio]{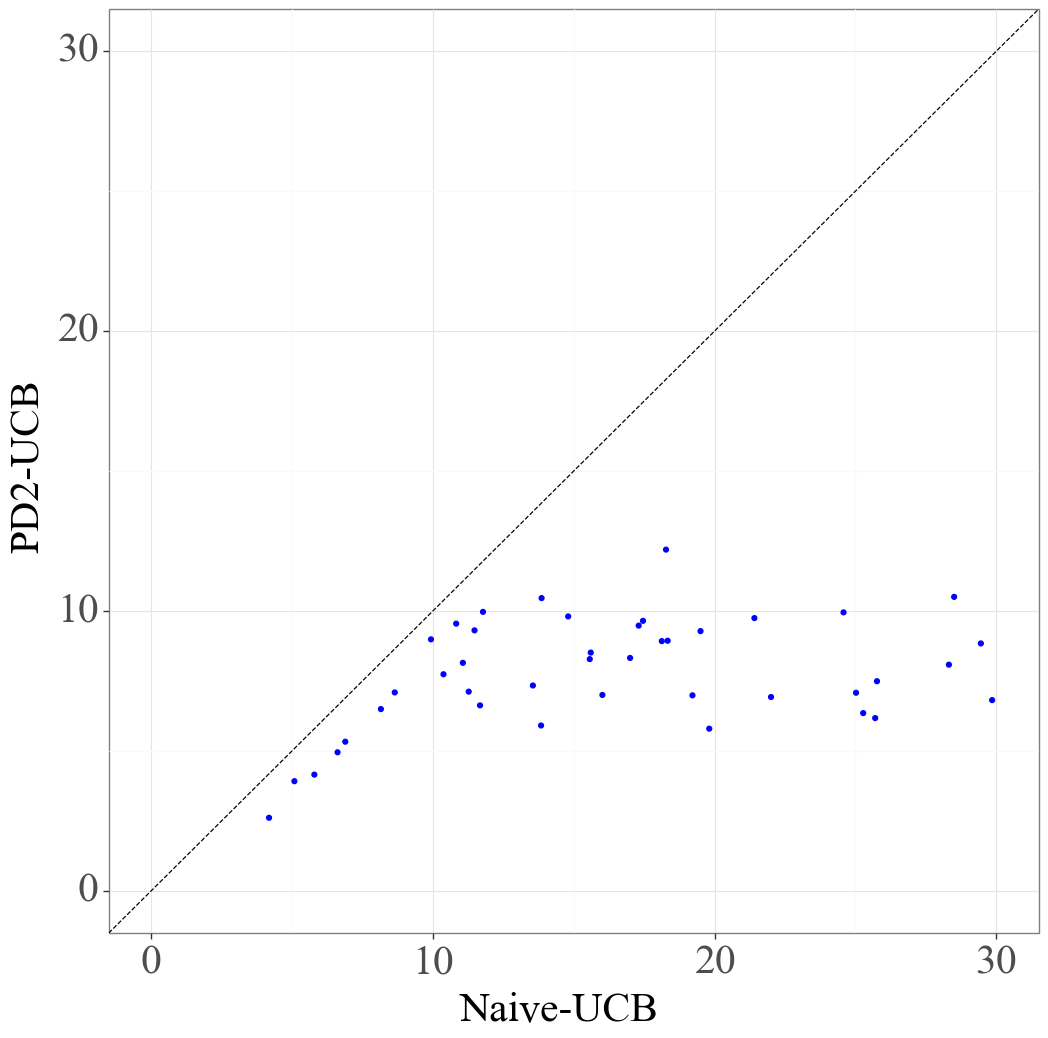}
\includegraphics[width=0.26\linewidth, keepaspectratio]{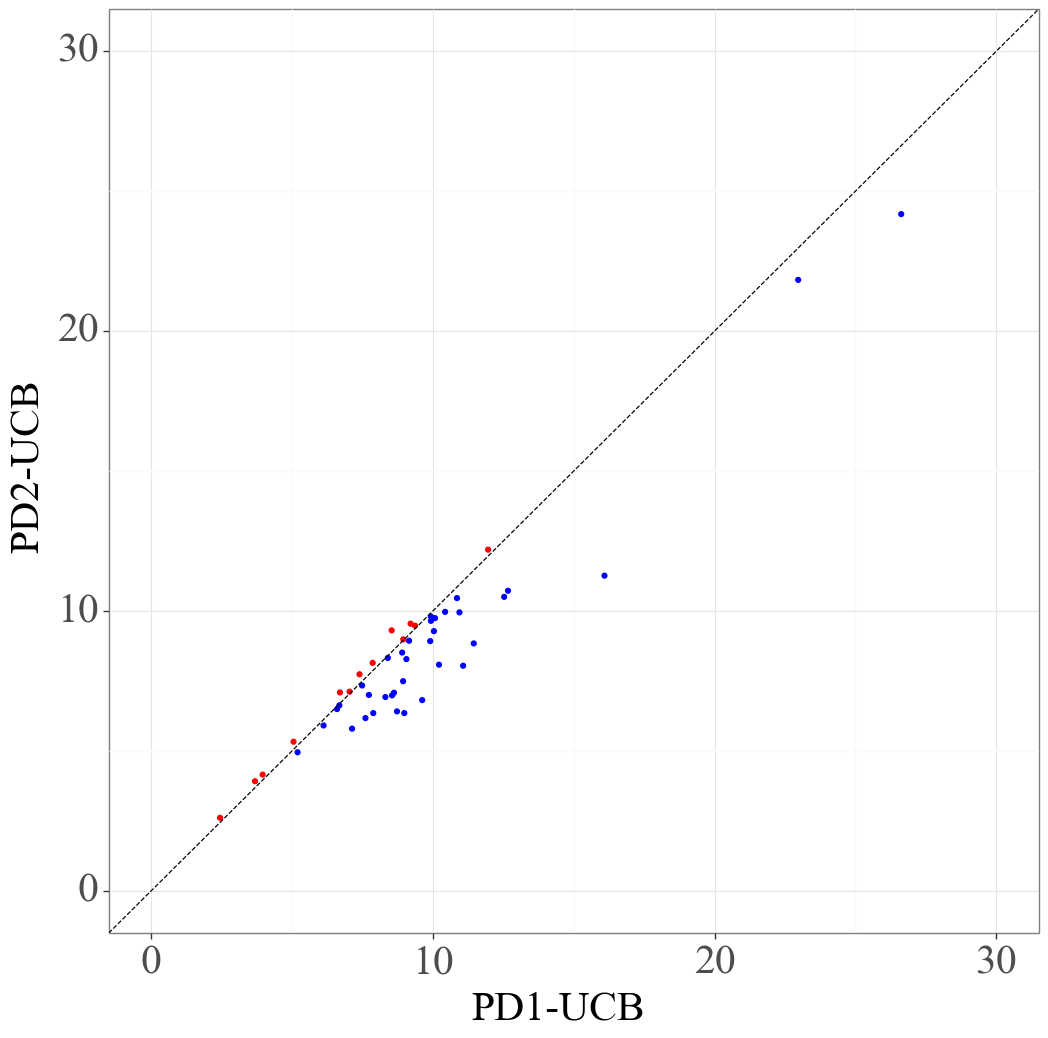}
    \caption{Average regret (left) and scatter plots for $B=10$ at $T=300$ under unknown $\mathbf{p}^*$. Each dot corresponds to one instance averaged over 50 replications.}
    \label{fig:contextb10}
\end{figure}

\begin{figure}
    \centering
\includegraphics[width=0.46\linewidth, keepaspectratio]{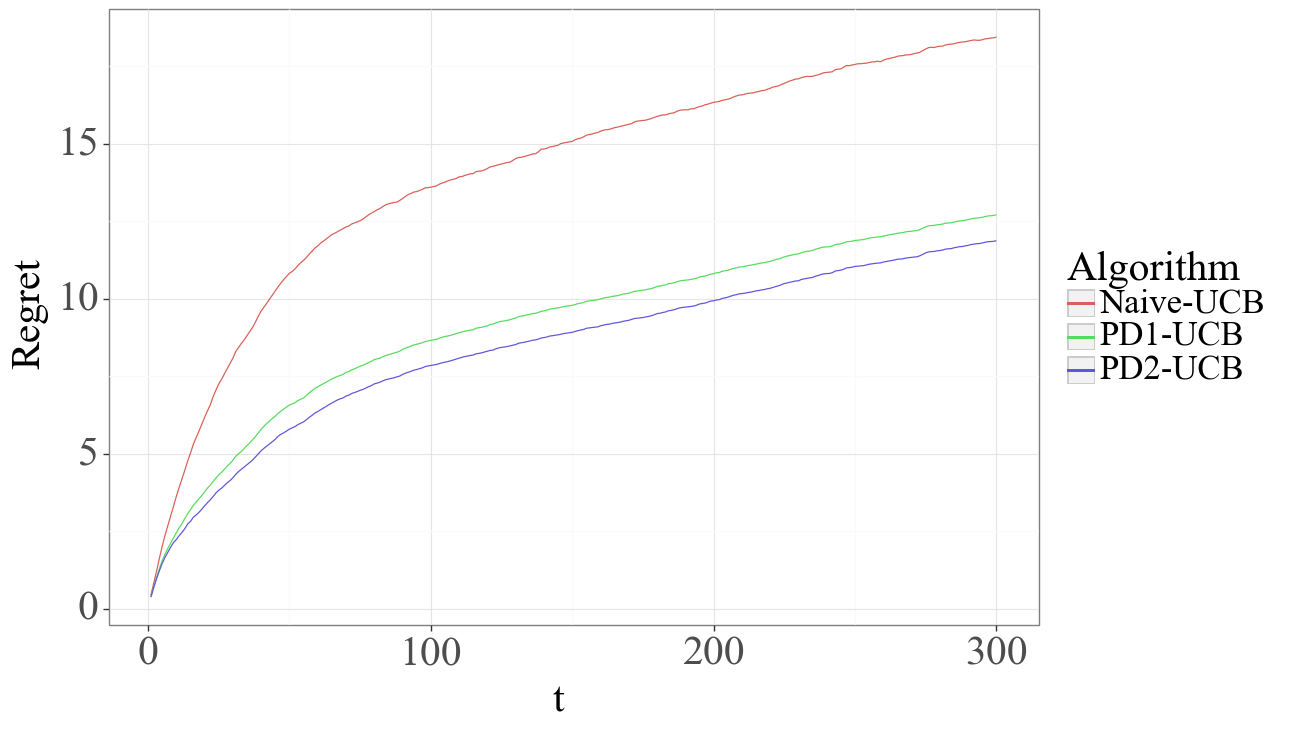}
\includegraphics[width=0.26\linewidth, keepaspectratio]{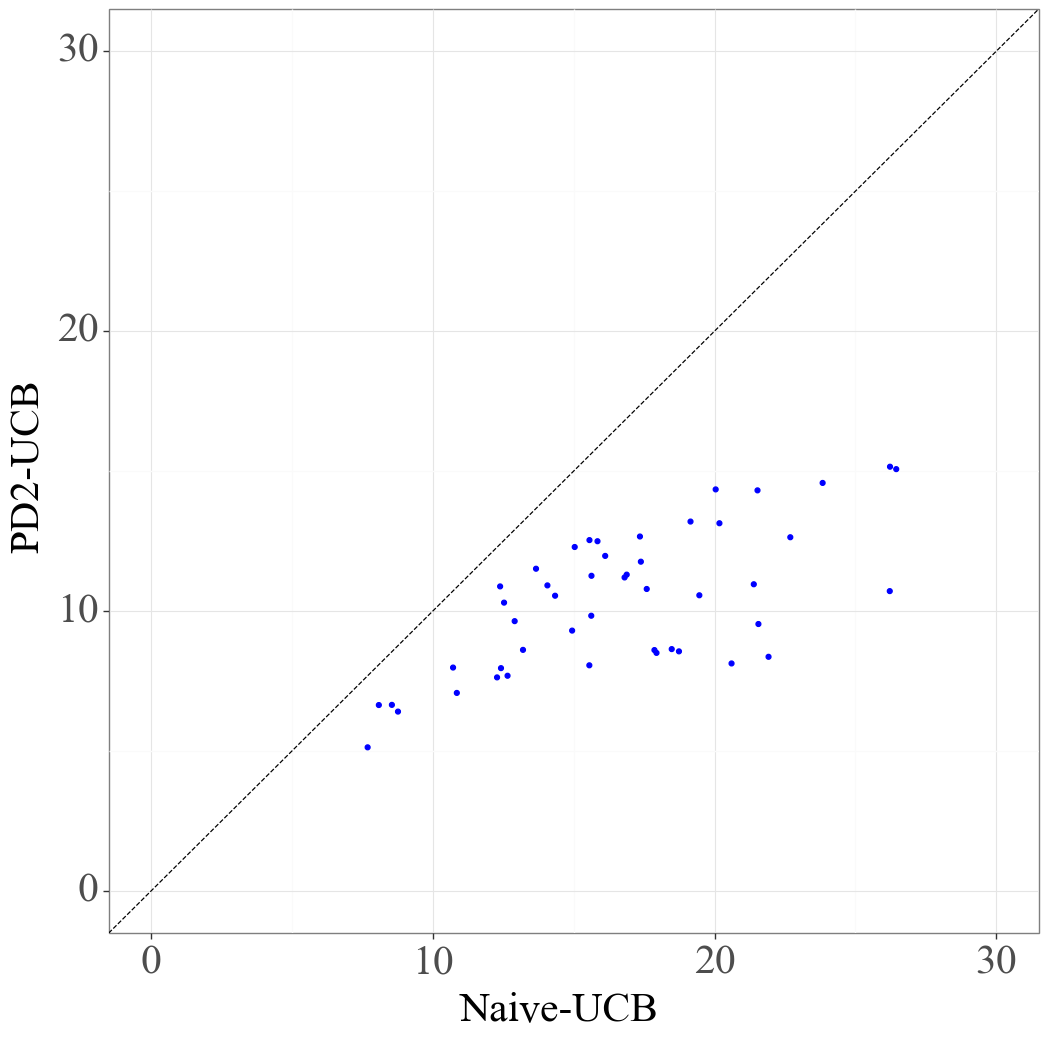}
\includegraphics[width=0.26\linewidth, keepaspectratio]{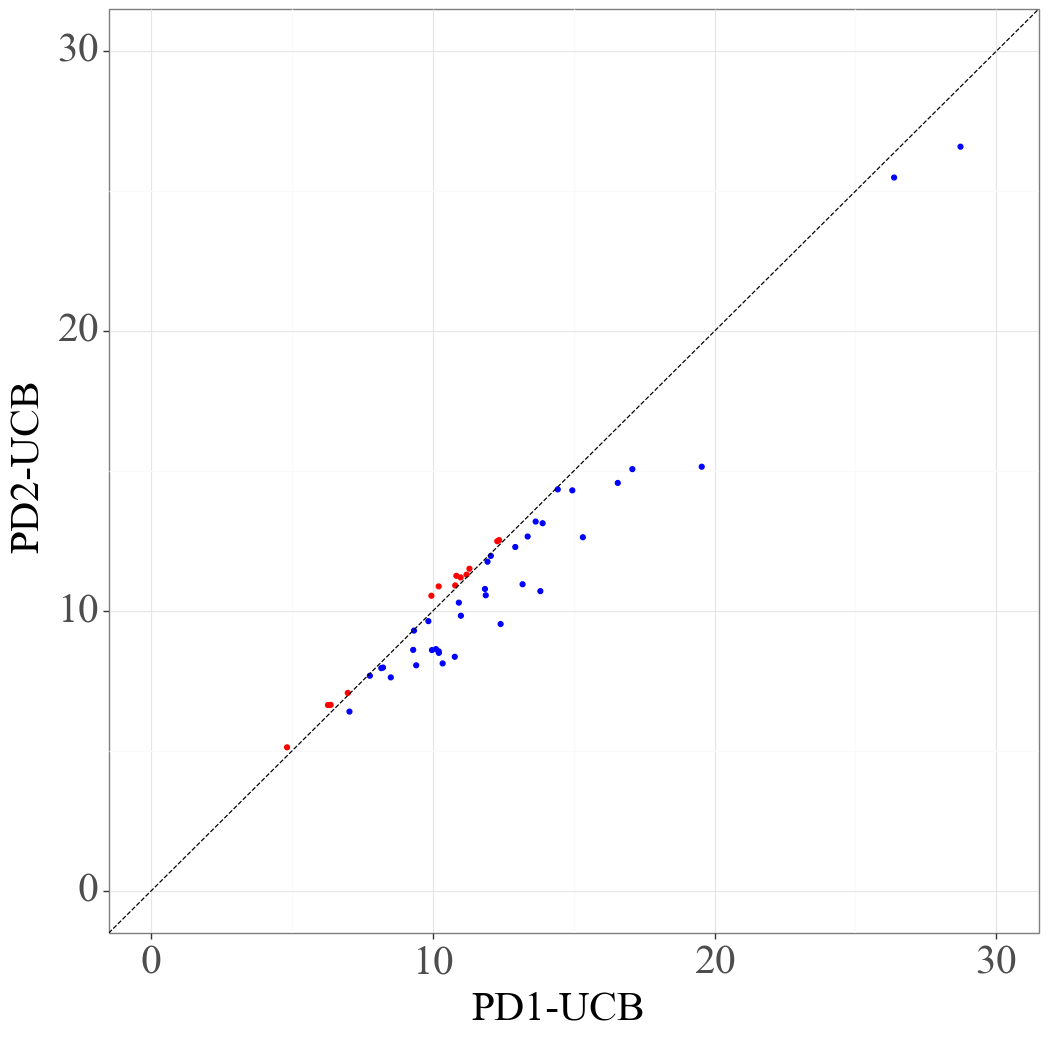}
    \caption{Average regret (left) and scatter plots for $B=20$ at $T=300$ under unknown $\mathbf{p}^*$. Each dot corresponds to one instance averaged over 50 replications.}
    \label{fig:contextb20}
\end{figure}

\begin{figure}
    \centering
\includegraphics[width=0.46\linewidth, keepaspectratio]{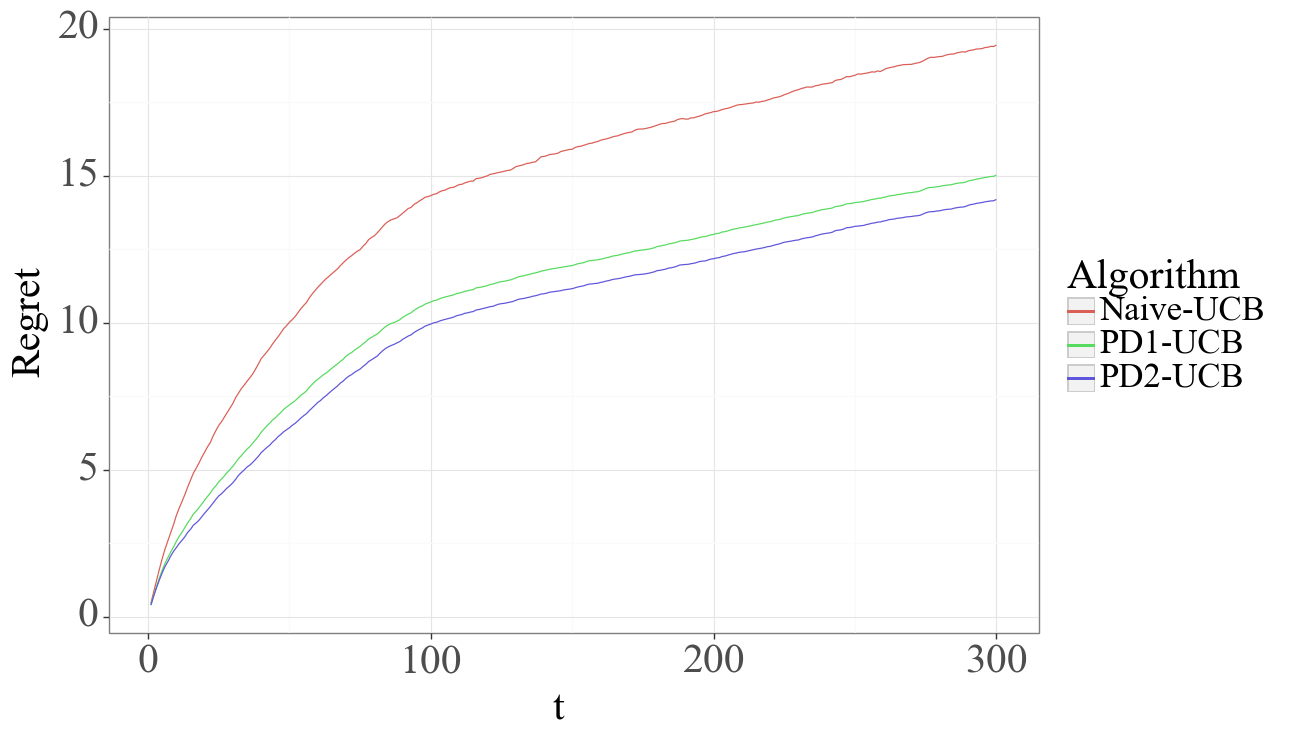}
\includegraphics[width=0.26\linewidth, keepaspectratio]{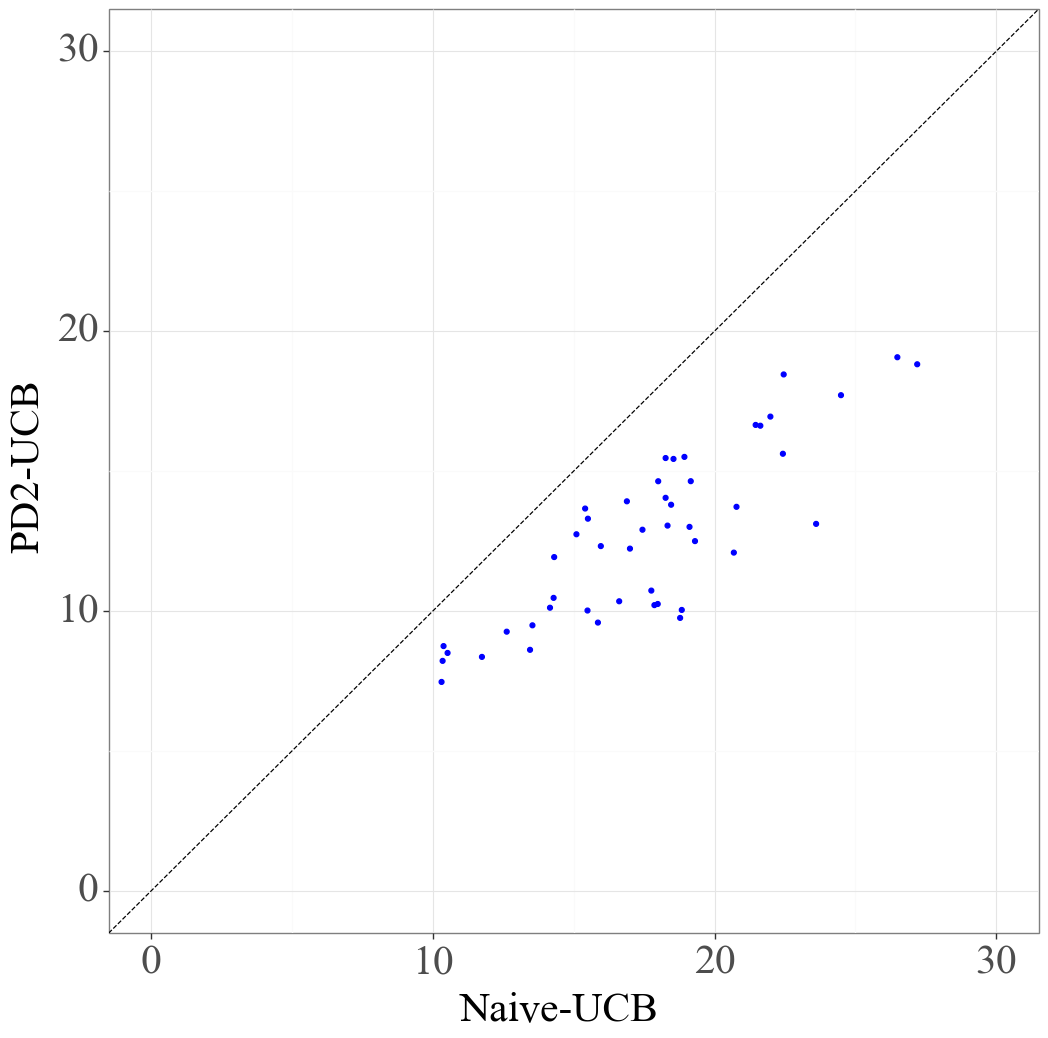}
\includegraphics[width=0.26\linewidth, keepaspectratio]{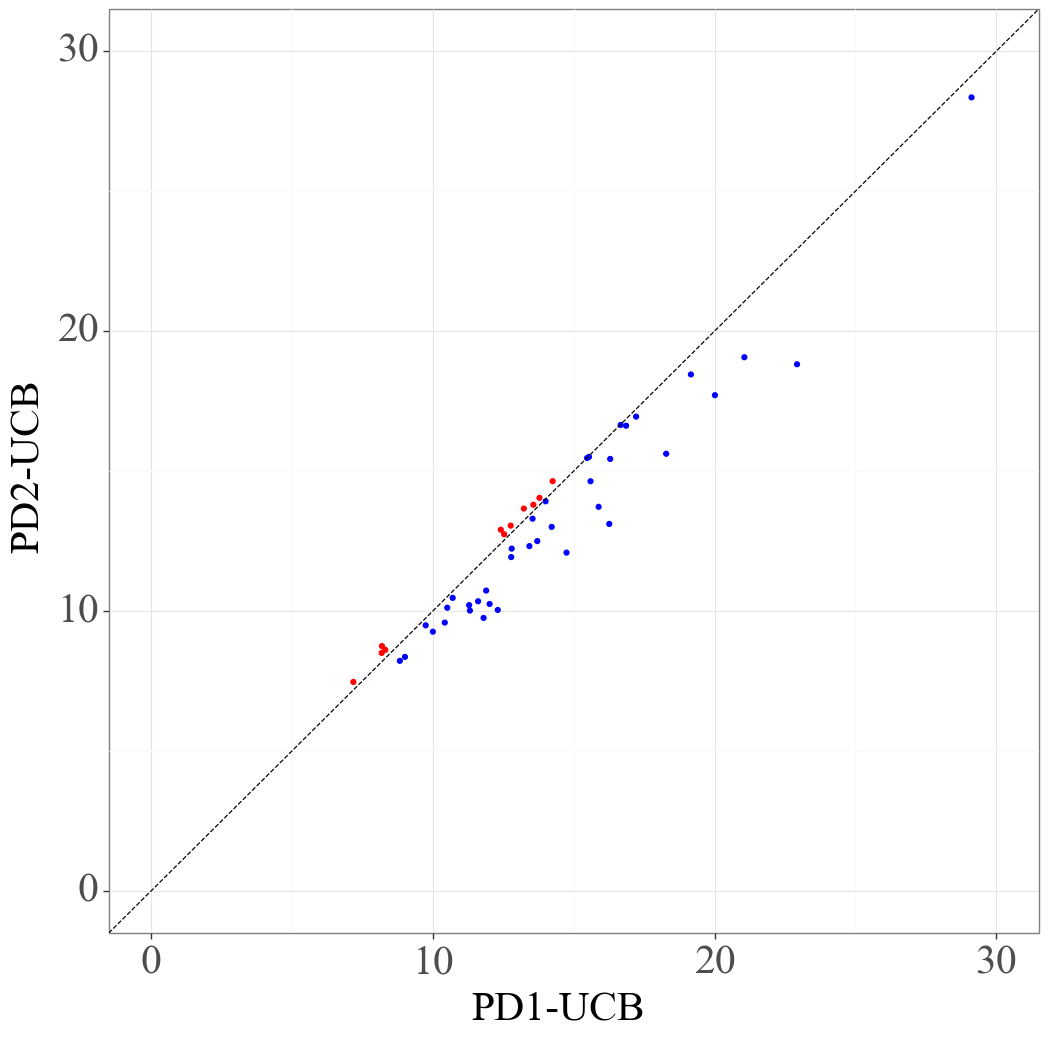}
    \caption{Average regret (left) and scatter plots for $B=30$ at $T=300$ under unknown $\mathbf{p}^*$. Each dot corresponds to one instance averaged over 50 replications.}
    \label{fig:contextb30}
\end{figure}

\clearpage
\section{Real-World Experiments on ROBAS 2 and 3}\label{app:robas3}

The simulation environment yields brushing quality $Q_{i,t}$ in response to action $A_{i,t}$ in state $S_{i,t}$, based on the following mathematical formulation:
$$
\begin{gathered}
Z_{i,t} \sim \operatorname{Bern}\left(1-\widetilde{p}_{ t}\right), \\
\widetilde{p}_{ t}=\operatorname{sigmoid}\left(g\left(S_{i,t}\right)^T w_{b}-A_{i,t} \max \left(h\left(S_{i,t}\right)^T \Delta_{ B}, 0\right)\right), \\
Y_{i,t} \sim \operatorname{Pois}\left(\lambda_{i,t}\right), \\
\lambda_{i,t}=\exp \left(g\left(S_{i,t}\right)^T w_{p}+A_{i,t} \max \left(h\left(S_{i,t}\right)^T \Delta_{N}, 0\right)\right), \\
Q_{i,t}=Z_{i, t} Y_{i,t}.
\end{gathered}
$$
Here, $g\left(S_{i,t}\right)$ is the baseline feature vector, and $h\left(S_{i,t}\right)$ represents the feature vector interacting with the effect size. Similar to \cite{trella2022designing}, we consider binary actions (sending a message versus not sending a message). For further details, refer to \cite{trella2022designing}. 




\end{document}